\documentclass[sn-mathphys]{sn-jnl}% Math and Physical Sciences Reference Style
\usepackage{subfig}
\usepackage{xcolor}
\usepackage{paralist}
\usepackage[normalem]{ulem}   % just here for del command, remove later
\usepackage{mathtools}
\usepackage{siunitx}
\usepackage{hyperref}
\usepackage{cleveref}

\usepackage{bm} % for bold symbol for greeks

\newcommand{\myparanovspace}[1]{\noindent \textbf{#1}\,\,}
\newcommand{\mypara}[1]{\vspace{4pt plus 2pt minus 2pt}\myparanovspace{#1}}
\newcommand{\myparastartsentence}[1]{\smallskip\noindent \textbf{#1}}
\providecommand{\abs}[1]{\lvert #1 \rvert}
\providecommand{\absbig}[1]{\Big\lvert #1 \Big\rvert}
\providecommand{\norm}[1]{\lVert #1 \rVert}
\providecommand{\normbig}[1]{\Big\lVert #1 \Big\rVert}
\newcommand{\diam}{\textrm{diam}}
\newcommand{\supp}{\textrm{supp}}

\newcommand{\todo}[1]{{\textcolor{red}{TODO: #1}}}
\newcommand{\Real}{\mathbb{R}}
\newcommand{\Exp}{\mathbb{E}}

\newcommand{\ppr}{n_r}  % points per ray

\newcommand{\Lewis}[1]{{\textcolor{magenta}{ \textbf{Lewis:} #1 }}}
\newcommand{\Louis}[1]{{\textcolor{brown}{ \textbf{Louis:} #1 }}}
\newcommand{\Richard}[1]{{\textcolor{blue}{ \textbf{Richard:} #1 }}}
\newcommand{\Colin}[1]{{\textcolor{orange}{ \textbf{cbm:} #1 }}}
\newcommand{\blue}[1]{{\textcolor{blue}{#1}}}

\newcommand{\del}[1]{{\textcolor{red}{ \sout{#1}}}}

\newcommand{\bbP}{\mathbb{P}}
\newcommand{\bbE}{\mathbb{E}}

\DeclareMathOperator*{\argmin}{arg\,min}

\renewcommand{\del}[1]{}
\renewcommand{\todo}[1]{}
\renewcommand{\Richard}[1]{}
\renewcommand{\Lewis}[1]{}
\renewcommand{\Louis}[1]{}
\renewcommand{\Colin}[1]{}
\renewcommand{\blue}[1]{#1}

\renewcommand{\S}{Sec.}

\renewcommand{\vec}[1]{\bm{#1}}

\jyear{2023}% %I changed this to 2022, hope it's correct

\theoremstyle{thmstyleone}%
\newtheorem{theorem}{Theorem}%  meant for continuous numbers
\newtheorem{proposition}[theorem]{Proposition}% 

\newtheorem{lemma}[theorem]{Lemma}

\theoremstyle{thmstyletwo}%
\newtheorem{remark}{Remark}%

\theoremstyle{thmstylethree}%
\newtheorem{definition}{Definition}%

\begin{document}

%%%%%%%%% TITLE
\title{Nearest Neighbor Sampling of Point Sets using Rays}

\author*[1]{\fnm{Liangchen} \sur{Liu}} \email{lcliu@utexas.edu} 
\equalcont{These authors contributed equally to this work.}
% \orcid{0000-0002-5102-8842}
% TODO: Louis affiliation?
\author[2]{\fnm{Louis} \sur{Ly}}\email{louisly@utexas.edu}

\author[3]{\fnm{Colin} \sur{B. Macdonald}}\email{cbm@math.ubc.ca}
\equalcont{These authors contributed equally to this work.}
% \orcid{ 0000-0002-4557-8637}
\author[1,2]{\fnm{Richard} \sur{Tsai}}\email{ytsai@math.utexas.edu}
\equalcont{These authors contributed equally to this work.}

\affil[1]{\orgdiv{Department of Mathematics}, \orgname{The University of Texas at Austin}, \orgaddress{\street{2515 Speedway}, \city{Austin}, \postcode{78712}, \state{TX}, \country{USA}}}

\affil[2]{\orgdiv{Oden Institute for Computational Engineering and Sciences}, \orgname{The University of Texas at Austin}, \orgaddress{\street{201 E 24th St}, \city{Austin}, \postcode{78712}, \state{TX}, \country{USA}}}

\affil[3]{\orgdiv{Department of Mathematics}, \orgname{University of British Columbia}, \orgaddress{\street{1984 Mathematics Rd}, \city{Vancouver}, \postcode{V6T 1Z2}, \state{BC}, \country{Canada}}}

\abstract{%
  We propose a new framework for the sampling, compression, and analysis of distributions of point sets and other geometric objects embedded in Euclidean spaces.
  Our approach involves constructing a tensor called the RaySense sketch, which captures nearest neighbors from the underlying geometry of points along a set of rays.
  We explore various operations that can be performed on the RaySense sketch, leading to different properties and potential applications.
  %For example,
  Statistical information about the data set can be extracted from the sketch,
  independent of the ray set.
  Line integrals on point sets can be efficiently computed using the sketch.
  We also present several examples illustrating applications of the proposed strategy in practical scenarios.
  %We present promising results from ``RayNN'', a neural network for the classification of point clouds based on RaySense signatures. \Lewis{maybe want to put down the emphasis on neural network?}
}

  %We show that the cost of the sampling approach is asymptotically independent of the ray sets.

% instructions said $4-6$ keywords, so we cut these:
% neural networks, importance sampling, integral transform

%\pacs[MSC Classification]{68T09,65D19}
% 68T09 Computational aspects of data analysis and big data
% 65D19 Computational issues in computer and robotic vision
% 68T07 Artificial neural networks and deep learning
% 65D40: High-dimensional functions; sparse grids?
% 65R10 Numerical methods for integral transforms: perhaps a stretch!

\maketitle

\section{Introduction}\label{sec:intro}

The comparison and analysis of objects in $d$-dimensional Euclidean space are fundamental problems in many areas of science and engineering, such as computer graphics, image processing, machine learning, and computational biology, to name a few.

When comparing objects in Euclidean space, one usually assumes that the underlying objects are solid or continuous. Typical examples include data manifolds and physical or probabilistic density representations.

One commonly used approach is distance-based comparison, which involves measuring the distance between two objects using metrics such as Euclidean distance, Manhattan distance, or Mahalanobis distance \cite{chandra1936generalised}. When the underlying object can be viewed as distributions, optimal transport \cite{villani2009optimal, peyre2019computational}, utilizing the Wasserstein distance, is also a popular choice.

However, in general, distance-based comparisons overlook helpful geometric information of the underlying objects, which turns out to be useful in many applications. This is because intrinsic geometric features such as curvature and volume are invariant to rotations and translations. Shape-based information is also suitable for comparing objects of different sizes or resolutions. Therefore, comparison techniques using geometric features are favorable choices in many scenarios, such as object recognition, classification, and segmentation. 

One simple example illustrating such an idea is the Monte--Carlo rejection sampling technique used to approximate the volume of an object. This technique considers the collision of $0$-dimensional objects (points) with the target object. This is a specific example of a collision detection algorithm \cite{lin1998collision}, which is popular in the computer graphics and computational geometry communities \cite{jimenez20013d}. The idea of collisions is also considered in the field of integral geometry \cite{santalo2004integral}, where one investigates the collision probability of certain affine subspaces with the target data manifold to deduce information about the manifold.

% \sout{Famous examples include Buffon's needle problem and the Crofton
% formula.}\Richard{I striked out the previous sentence because there is more detail description in Section 1.1.}
Furthermore, in integral geometry, one considers integral transforms on the underlying objects. Typical examples are the X-ray transform \cite{solmonXRay1976} and the Radon transform \cite{radon20051,helgason1980radon}. These transforms can provide a more compact and informative representation than the original data. For example, one can recover spectral information from the X-ray transform or reconstruct the original object through an inverse Radon transform.

However, the aforementioned techniques generally rely on the assumption that the underlying object is solid or continuous. With the prevalence of big data and advancements in sensing technology, such as LiDAR, the analysis and comparison of point cloud data (which consists of a set of points in some $d$-dimensional Euclidean space) have gained increasing attention, yet they pose challenges to classical approaches.

\begin{figure*}[htbp]
		\includegraphics[width=0.99\textwidth]{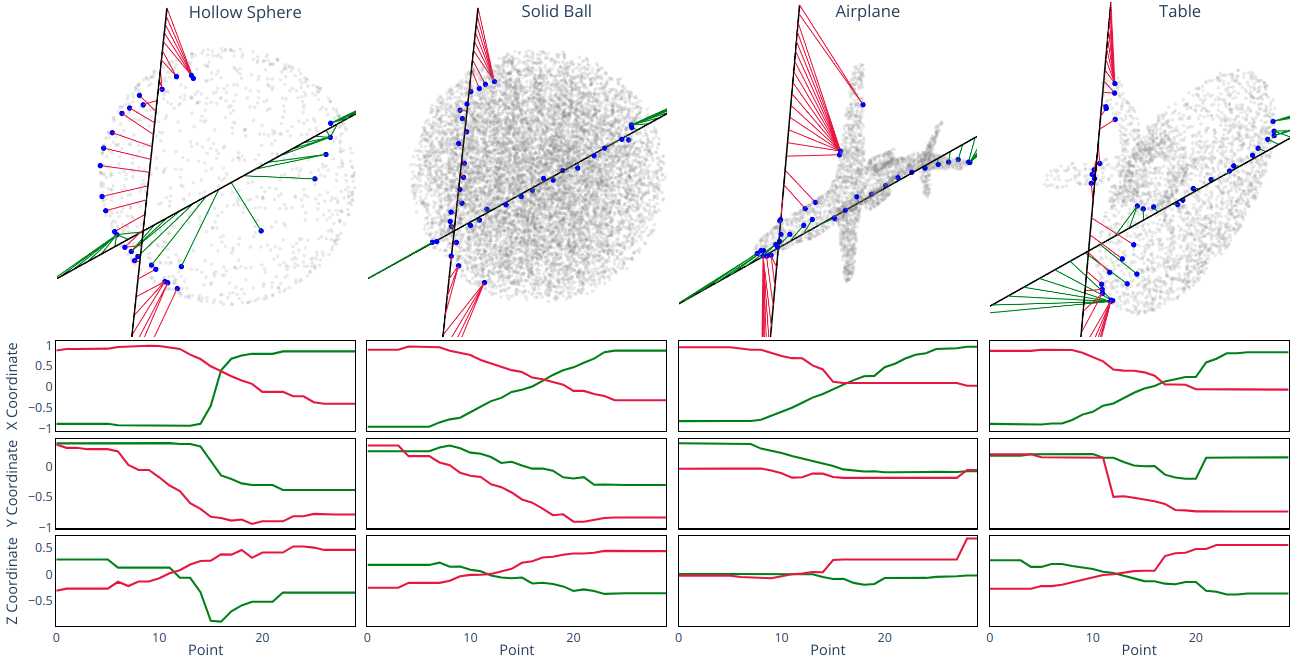}
	\caption{RaySense sketches using 30 sample points per ray.
	Row 1: visualization of two rays (black) through points sampled from various objects (gray).
	Closest point pairs are shown in green and red. Rows 2--4: the $x$, $\vec y$, and $z$ coordinates of the closest points to the ray.}
	\label{fig:grid}
\end{figure*}

We propose a novel method for sampling, comparing, and analyzing point clouds, or other geometric objects, embedded in high-dimensional space. We call our approach ``RaySense" because it ``senses" the structure of the object $\Gamma$ by sending line segments through the ambient space occupied by $\Gamma$ and recording some functions of the nearest neighbors in $\Gamma$ to points on the line segment. Motivated by the X-ray transform, we will refer to the oriented line segments as \emph{``rays"}. We can then work with this sampled data as a ``sketch" of the original object $\Gamma$, which can be a point cloud, triangulated surface, volumetric representation, etc.
A visualization of the proposed method applied to $3$D point clouds is
provided in Fig.~\ref{fig:grid}.

Our method incorporates several ideas mentioned above:
\begin{itemize}
\item In the context of integral geometry, we also consider using affine subspaces to detect the underlying geometry. To overcome the discontinuity from the object representation relative to the topology of the ambient space (as is the case with point clouds), we search for nearest neighbors within the representation. To prevent the computational cost of high-dimensional operations in practice, we work with a low-dimensional affine subspace. In this paper, we use $1$-dimensional lines.
\item In the context of inverse problems, our observed data consist of some points on $\Gamma$ obtained by the closest point projection of points on the rays. This is somewhat analogous to seismic imaging, where designated points on each ray correspond to geophones that record the first arrival time of waves from known sources.
\item By using a fixed number of rays and sampling a fixed number of points along a ray, the resulting tensor from our method, the RaySense sketch, is of fixed size, even for point clouds with different cardinalities. Then, metrics suitable for comparing tensors are available for use, or one could consider other distance-based approaches to compare different RaySense sketches.
\end{itemize}

In this work we will focus exclusively on point clouds since it is one of the most challenging data representations for many algorithms. When the object is a point cloud, finding the RaySense samples is straightforward via discrete nearest neighbor searches. 
There are computationally efficient algorithms for performing nearest-in-Euclidean-distance neighbor searches. Examples include tree-based algorithms~\cite{Bentley1975kdtree}, grid-based algorithms~\cite{tsai2002rapid}, and building an indexing structure \cite{johnson2019billion}. In very high dimensions, one may also consider randomized nearest neighbor search algorithms~such as \cite{indyk1998approximate, JonesOsipovRokhlin2011approxnn}, or certain dimension reduction approaches.

The remaining paper is organized as follows:
in \Cref{sec:signature}, we provide a detailed description of the RaySense method;
in \Cref{sec:analysis}, we present various properties of the RaySense method, along with theoretical analysis, including line integral approximations and salient point sampling;
finally, in \Cref{sec:application}, we demonstrate that the concept of RaySense can be applied to many different downstream tasks, including salient point/outlier detection, point cloud registration, and point cloud classification.
\subsection{Related work}

In this section, we provide a more detailed discussion of the existing literature to contextualize our approach.

\paragraph{Integral geometry}
In the field of Integral Geometry \cite{santalo2004integral}, one uses the probability of intersection of affine subspaces of different dimensions with the target data manifold to deduce information about the manifold itself.
For example, in the classic problem of Buffon's needle, one determines the length of a dropped needle by investigating the probability of intersections with strips on a floor.
  Similarly, Crofton's formula connects the length of a $2$D plane curve with the expected number of intersections with random lines. In these cases, the interaction information obtained from the ``sensing'' affine subspaces is binary: yes or no.
One thus has a counting problem: how frequently affine subspaces intersect with the data manifold.
From these probabilities, geometric information about the manifold can be extracted, relying on a duality between the dimensions of the ``sensing'' subspaces and the Hausdorff dimensions of the data set; see e.g.,  \cite{hadwiger1957vorlesungen, klain1997introduction}.
Nevertheless, such approaches may be inefficient in practical computations.
%This probability is proportional to the integral quantities that one wishes to measure from the data.
%This line of study leads to Hadwiger’s theorem \todo{cite}.

Our idea of using rays is perhaps most-closely related to the X-ray transform,
which coincides with the Radon Transform in 2D \cite{Natterer:computerized_tomography_siambook}.
In an X-ray transform, one integrates a given real-valued function
defined in $\mathbb{R}^d$ along lines, while in the Radon transform, one
integrates the given function restricted on hyperplanes of $\mathbb{R}^d$.

We advocate using lines (rays) to record information about the underlying data along each ray, instead of accumulating a scalar or binary ``yes/no'' information over each rays.
In this paper, we collect points in the data set closest to the points on the rays, along with the values of some function at those points.
With such data, we can compute approximate line integrals and relate our method to the X-ray transform.
% One can draw an analogy to seismic imaging, where designated points on each ray correspond to geophones that record the first arrival time of waves from known sources.
% One difference is that in seismic imaging, the sensor arrays typically lie on top of the domain of interest---in RaySense, the sensors are placed on rays that penetrate the ambient space.

\paragraph{Computer vision}

From the perspective of the computer vision community,
our method can be considered as a shape descriptor,
mapping from 3D point sets to a more informative feature space where point sets can be easily compared.
Generally, descriptors aim to capture statistics related to local and global features.
See \cite{kazmi2013survey} for a survey. More recently, researchers have combined shape descriptors with machine learning \cite{fang20153d,rostami2019survey,sinha2016deep,xie2016deepshape}. 
But all these works focus primarily for point sets in $3$D, RaySense applies more generally to data in arbitrary dimensions.

Some methods use machine learning directly on the point set to  learn features for specific tasks, such as classification
\cite{atzmon2018point,klokov2017escape,li2018pointcnn,qi2017pointnet, qi2017pointnet++, simonovsky2017dynamic, wang2017cnn,wang2019dynamic, zhou2018voxelnet}.
PointNet \cite{qi2017pointnet} pioneered deep learning on point clouds by applying independent operations on each point and aggregating features using a symmetric function. % to learn the global shape descriptor.
Building upon that, other architectures \cite{qi2017pointnet++,Shen_2018_CVPR} exploit neighboring information to extract local descriptors.
SO-Net \cite{li2018so} uses self-organizing maps to hierarchically group nodes while applying fully-connected layers for feature extraction.
PCNN \cite{atzmon2018point} defines an extension and pulling operator similar to the closest point method \cite{ruuth2008simple, macdonald2013simple} to facilitate the implementation of regular convolution. DGCNN \cite{wang2019dynamic} and PointCNN \cite{li2018pointcnn} generalize the convolution operator to point sets.

In contrast, our approach uses the RaySense sketch as input rather than applying machine learning directly to the point set.
In \S~\ref{sec:NN}, we present a deep learning model for $3$D point cloud classification based on this idea. Our experiments suggest that
the model is very efficient for classification, and
%our experiments suggest that by storing \sout{$\eta$}
%\uline{multiple} % \Lewis{the $\eta$ is not introduced}
%nearest neighbors to points on the rays, the resulting classifiers are robust against outliers.
the resulting classifiers can be robust against outliers by storing multiple nearest neighbors to points on the ray.

\paragraph{Unsupervised learning methods}

Our method, by recording the closest points from the underlying point cloud, can be thought as a sampling scheme.
However, the RaySense sampling is biased towards the ``salient'' points in the underlying geometry, as will be discussed in~\S~\ref{sec:salient}.
By further retaining only the most frequently sampled points, RaySense resembles key-point detectors \cite{krig2016interest} or compression algorithms.
The idea of understanding the overall structure of the underlying data through key points is closely related to archetypal analysis \cite{cutler1994archetypal},
which involves finding representative points (archetypes) in a dataset and using a convex combination of these archetypes to represent other points. 
%\del{However, Archetypal analysis may be sensitive to initialization and the number of archetypes used in the analysis.}
See also \cite{Osting:ArchetypalAnalysis} for a recent work on the consistency of archetypal analysis. 
% \del{
% To the best our knowledge, there is no well-defined metric for determining the quality of the representation, making it hard to compare the results of different analyses.
%  }
% \Colin{why speculate here?}

Incidentally, \cite{graham2017approximate} also employs the concept of rays in conjunction with spherical volume to approximate the convex hull. 
Our method can also capture vertices of the convex hull when the rays are sufficiently long, as it will effectively sample points on the portions of the boundary that have relatively large positive curvature.

Farthest Point Sampling (FPS) is a widely-used sampling method in computational geometry and machine learning for selecting a subset of points from a larger dataset with the goal of maximizing their spread; see e.g. \cite{eldar1997farthest}. The process begins by randomly picking a point from the dataset, followed by iterative selection of the point farthest from those already chosen, until a desired number of points are selected. This technique is useful for reducing the size of large datasets while preserving their overall structure. However, it can be computationally expensive and may not always yield the optimal solution.
In two and three dimensions, assuming that the data distribution is supported in a bounded convex set with a smooth boundary, FPS  tends to oversample areas with high curvature.

%Computational techniques used in ray-casting and
%ray-tracing~\cite{KrugereWestermann2003:gpuvolume, hadwiger2005:raycasting, Peddie2019RayTracingBook}
%may be useful for improving efficiency of RaySense implementations.

%\begin{figure}[tb]
%\centering
%\includegraphics[width=0.9\linewidth]{figures/ray_density.png}
%\caption{\label{fig:rays} Density of rays from method R1 (left) and R2 (right). Red circle indicates the $\ell^2$ ball.}
%\end{figure}

\section{Methods}\label{sec:signature}

The essential elements of the proposed sampling strategy include (1) the data set $\Gamma\subset\mathbb{R}^d$;
(2) the nearest neighbor (closest point) projection, $P_{\Gamma}$, to $\Gamma$; (3) a distribution of lines in $\mathbb{R}^d$.

The data set is given by a discrete set of $N$ points, each in $\Real^d$:
\begin{equation}
  \Gamma \subset \Real^d, \qquad \Gamma = \{\vec X_i\}_{i=1}^N,  \qquad \vec{X}_i\in \mathbb R^d,
\end{equation}
(later for certain results we will place more assumptions, for example that $\Gamma$ might be sampled from a density.)

Let $\mathcal{L}$ denote a distribution of lines,
 parameterized by $\vec{\theta}\in\mathbb{S}^{d-1}$ and $\vec b\in\mathbb{R}^d$,
let $\vec r(s)$ denote a line in $\mathcal{L}$ 
parameterized by its length
\begin{equation*}
 \vec r(s) = \vec b + s \vec{\theta}.
\end{equation*}
The parameterization gives an orientation to the line, and thereby we refer to $\vec r$ as a ray.
Along $\vec r(s)$, we sample from the data set $\Gamma$ using the nearest neighbor projection
\begin{equation}
  \mathcal{P}_{\Gamma} \vec r(s) \in \argmin_{\vec y\in\Gamma} \|\vec r(s)-\vec y\|_2.
  \label{eq:cp}
\end{equation}
In cases of non-uniqueness, we choose arbitrarily.

Using the nearest neighbors, we define various RaySense sampling operators denoted
with $\mathcal{S}$
which sample from the point cloud $\Gamma$ into some ``feature space'' $\mathbb{X} \subset \Real^c$.
The simplest choice is the feature space of closest points defined next.

\begin{definition}
  The closest point feature space is sampled by
  \begin{equation}
    \mathcal{S}[\Gamma] : \Real^d \to \Real^d,
    \qquad
    \mathcal{S}[\Gamma](\vec r(s))  = \mathcal{P}_{\Gamma}\vec r(s),
    \qquad
    \vec r \sim \mathcal{L}.
    \label{eq:closest_point_feature_space}
  \end{equation}
\end{definition}

One might also be interested in the value of a scalar or vector function % $g : \Gamma \to \Real^c$
(e.g., color or temperature data at each point in the point cloud). 
\begin{definition}
  The RaySense sampling operator of a function $g : \Gamma \to \Real^c$ is
  \begin{equation}
    \mathcal{S}[\Gamma, g] : \Real^d \to \Real^c,
    \qquad
    \mathcal{S}[\Gamma, g](\vec r(s))  = g\big(\mathcal{P}_{\Gamma}\vec r(s)\big) \blue{,}
    \qquad
    \vec r \sim \mathcal{L}.
    \label{eq:g_feature_space}
  \end{equation}
  Note for the identity function we have $\mathcal{S}[\Gamma, \mathrm{id}] = \mathcal{S}[\Gamma]$.
\end{definition}
We will further present examples involving the use of \emph{multiple} nearest neighbors,
where the $\eta$-th nearest neighbor is
\begin{align}\label{def:k-closest_point_old}
  \mathcal{P}_{\Gamma}^{\eta}(\vec r(s)) &:= \argmin_{\vec y \in \Gamma
    \setminus\{\mathcal{P}_{\Gamma}^1\vec r(s), \ldots, \mathcal{P}_{\Gamma}^{\eta-1}\vec r(s)\}} \|\vec r(s)-\vec y\|_2,
\end{align}
with $\mathcal{P}_{\Gamma}^1 := \mathcal{P}_\Gamma$.
We can then capture the $\eta$-th nearest neighbor in our sampling.
\begin{definition}
  The RaySense sampling operator of the $\eta$-th nearest neighbor is denoted by:
  \begin{equation}
    \mathcal{S}[\Gamma, \eta] : \Real^d \times \mathbb{N} \to \Real^d,
    \qquad
    \mathcal{S}[\Gamma, \eta](\vec r(s)) = \mathcal{P}^{\eta}_{\Gamma}\vec r(s),
    \qquad
    \vec r \sim \mathcal{L}.
    \label{eq:eta_nn_feature_space}
  \end{equation}
\end{definition}
The distance and direction from $\vec r(s)$ to $\mathcal{P}_\Gamma \vec r(s)$ is
sometimes useful: we sample that information using the vector from
$\vec r(s)$ to $\mathcal{P}_\Gamma \vec r(s)$:
\begin{definition}
  The RaySense sampling operator of the vector between a point on the
  ray and the nearest neighbor in $\Gamma$ is
\begin{subequations}
\begin{equation}
  \mathcal{S}[\Gamma, \hat{1}] :  \Real^d \times \mathbb{N} \to \Real^{d},
  \qquad
  \mathcal{S}[\Gamma, \hat{1}] (\vec r(s)) = \mathcal{P}_\Gamma \vec r(s)  - \vec r(s),
  \qquad
  \vec r \sim \mathcal{L},
  \label{eq:featspace_cpvec}
\end{equation}
and more generally
\begin{equation}
  \mathcal{S}[\Gamma, \hat{\eta}] :  \Real^d \times \mathbb{N} \to \Real^{d},
  \qquad
  \mathcal{S}[\Gamma, \hat{\eta}] (\vec r(s)) = \mathcal{P}^{\eta}_\Gamma \vec r(s)  - \vec r(s),
  \qquad
  \vec r \sim \mathcal{L}.
  \end{equation}
\end{subequations}
\end{definition}

We can augment the feature space by combining these various operators,
concatenating the output into a vector $\mathcal{X} \in \Real^c$.
We indicate this ``stacking'' with a list notation in $\mathcal{S}$,
for example the first three closest points could be denoted:
\begin{equation}
  %\mathcal{S}\big[\Gamma, [1, 2, 3]\big] : \Real^d \to \Real^{3d},
  %\quad
  \mathcal{S}\big[\Gamma, [1, 2, 3]\big](\vec r(s))
  =
  \begin{bmatrix}
    \mathcal{P}^1_{\Gamma}\vec r(s) \\
    \mathcal{P}^2_{\Gamma}\vec r(s) \\
    \mathcal{P}^3_{\Gamma}\vec r(s)
  \end{bmatrix}\!,
  \quad
  \vec r \sim \mathcal{L},
\end{equation}
or the closest point, its vector from the ray, and the value of a %scalar
function $g$ could all be denoted:
\begin{equation}
  %\mathcal{S}\big[\Gamma, [1, \hat{1}, g]\big] : \Real^d \to \Real^{2d + 1},
  %\quad
  \mathcal{S}\big[\Gamma, [1, \hat{1}, g]\big](\vec r(s))
  =
  \begin{bmatrix}
    \mathcal{P}_{\Gamma}\vec r(s) \\
    \mathcal{P}_{\Gamma}\vec r(s) - \vec r(s) \\
    g\big(\mathcal{P}_{\Gamma}\vec r(s)\big)
  \end{bmatrix}\!,
  \quad
  \vec r \sim \mathcal{L}.
  \label{eq:augmented_featspace_cpvec}
\end{equation}
These stacked feature spaces are used in the line integral approximation \S~\ref{sec:integral} and in our neural network \S~\ref{sec:NN}.

In summary,
a RaySense sketch $\mathcal{S}[\Gamma, \ldots]$ depends on nearest neighbors in the data set $\Gamma$, and operates on a ray from the distribution $\mathcal{L}$.
It maps a ray $\vec r(\cdot)$ to a piecewise curve in the chosen feature space $\mathbb{X}$.
%\uline{, defined by the choice of function $g$}.

\subsection{Discretization}

We propose to work with a discretized version of the operator, which we shall call a RaySense sketch tensor.
First, we take $m$ i.i.d.\ samples from the distribution $\mathcal{L}$ and define $m$ rays correspondingly.
We consider $\ppr$ uniformly-spaced points along each ray, with corresponding spacing $\delta r$.
We then work with a finite segment of each line (for example, $0 \le s \le 1$).
Appendix~\ref{sec:method} shows some different distributions of lines, and details for choosing segments from them.
% By collecting all the recorded information from the $m$ rays together, we obtain the RaySense signature:
% \begin{equation}
%     S_m(\Gamma,f;\mathcal{L}),
% \end{equation}
% consisting of $m$ random samples of these {paths}.
% We will work with a discrete version of the signature defined in the following way.
With $\vec r_{i,j}$ denoting the \mbox{$j$-th} point on the $i$-th ray, we define the
\emph{discrete RaySense sketch} of $\Gamma$ in the closest point feature space as
\begin{subequations}\label{eq:discrete_signature}
\begin{equation}
  \mathcal S_{m,\delta r}[\Gamma; \mathcal{L}]_{i,j} :=
  \mathcal{P}_{\Gamma}\vec r_{i,j},
  \label{eq:discrete_sig_simple}
\end{equation}
or the discrete RaySense sketch %using the $\nu$-th nearest neighbor and
of a function $g$:
\begin{equation}
  \mathcal S_{m,\delta r}[\Gamma, g; \mathcal{L}]_{i,j} :=
  g\big(\mathcal{P}_{\Gamma} \vec r_{i,j}\big),
  %\hat{f}_{\eta}\!\left(\mathcal{P}_{\Gamma}^1\vec r_{i,j}, \ldots, \mathcal{P}_{\Gamma}^{\eta}\vec r_{i,j}, \vec r_{i,j}\right),
  \label{eq:discrete_sig_multi}
\end{equation}
\end{subequations}
(and similarly for the various more-general feature spaces mentioned earlier).

Thus, $\mathcal S_{m,\delta r}$ is an array with $m$ entries,
where each entry is an array of $\ppr$ vectors in $R^c$; an $m \times \ppr \times c$ tensor.
%For simplicity, for $\eta=1$,
We will also denote these tensors as ``$\mathcal S(\Gamma)$'' and ``$\mathcal S(\Gamma, g)$''
when $m$ and $\delta r$ are not the focus of the discussion.
In any case, we regard $\mathcal S(\Gamma)$ as a ``sketch'' of the point cloud
$\Gamma$ in a chosen feature space $\mathbb{X}$.

\subsection{Operations on RaySense sketches}

In this paper, we will analyze the outcomes of the following
operations performed on the RaySense sketches.
After discretization, each operation can be expressed as one or more summations
performed on the sketches.

\subsubsection{Histograms}

One simple operation is to aggregate the sampled values for the feature space and count their corresponding sampling frequencies, which results in a histogram demonstrating the discrete distribution of the corresponding features.
For example, when considering the closest point feature space $\mathcal S[\Gamma]$, the value of a bin in the histogram, denoted by $H_k$, represents the number of times $\vec x_k\in\Gamma$ being sampled by RaySense:
\begin{equation}
  \label{eqn:histogram}
  H_k := H \big(\vec x_k; \mathcal S_{m,\delta r}[\Gamma]\big)
  = \sum_{i=1}^m \sum_{j=1}^{\ppr} \chi_{\vec x_k}\big({\mathcal{S}[\Gamma]_{i,j}}\big),
\end{equation}
where $\chi_{\vec x_k}$ denotes the indicator function for the coordinates of $\vec x_k$.
The histograms discard locality information in the sketch, treating it essentially as a weighted subsampling of $\Gamma$, as the aggregation process involves combining the values without considering the specific rays from which they originated.
\S~\ref{sec:convergence-of-histograms} further discusses properties of the histogram.

\subsubsection{Line integrals} \label{sec:method_lin_int}
In many applications, it is useful to maintain some ``locality''.
One such operation is the line integral along each ray $\int_{0}^{1} g(\vec r(s)) \,\textrm{d}s$.
However, for point cloud data, exact information of $g$ along the ray $\vec r(s)$ is not accessible, we instead consider the integral along the associated path in feature space,
$\int_0^1 \mathcal{S}[\Gamma, g]\left(\vec r(s)\right) \,\textrm{d}s$,
which we call \emph{a RaySense integral}.
We investigate the relationship between the two in \S~\ref{sec:integral}.

In the discrete setting, we can use a simple Riemann sum quadrature
scheme to approximate the RaySense integral along the $i$-th ray:
\begin{subequations}
  \label{eq:discrete_raysense_integral}
\begin{equation}
  \label{eq:riemann_sum}
  %\int f\left(\mathcal{P}^\eta_\Gamma \vec r_i(s)\right) \,\textrm{d}s
  \int_0^1 \mathcal{S}[\Gamma, g]\left(\vec r_i(s)\right) \,\textrm{d}s
  \approx
  \sum_{j=1}^{\ppr} \mathcal S[\Gamma, g]_{i,j} \,\delta r.
\end{equation}
Or in most of our examples, we approximate using the Trapezoidal Rule:
\begin{equation}
  \label{eq:trapezoidal_rule}
  %\int_0^1 g(\vec r(s)) \,\textrm{d}s
  \int_0^1 \mathcal{S}[\Gamma, g]\left(\vec r_i(s)\right) \,\textrm{d}s
  \approx
  %\frac{1}{2}(g(s_1)\delta r) + (g(s_1)\delta r) + \ldots + (g(s_{k-1})\delta r) + \frac{1}{2}(g(s_{k})\delta r)
  \sum_{j=1}^{\ppr} w_j \, \mathcal S[\Gamma, g]_{i,j} \, \delta r,
  \quad \text{with weights $w = \left\langle \tfrac{1}{2}, 1, \ldots, 1, \tfrac{1}{2}\right\rangle$,}
\end{equation}
and thus quadrature errors, typically $\mathcal{O}(\delta r^2)$, are incurred \cite{TrefethenWeideman:trapezoidalrule}.
Unpacking the notation, we can rewrite this as
\begin{equation}
  \sum_{j=1}^{\ppr} w_j g(\mathcal{P}_{\Gamma} \vec r_i(s_j)) \, \delta r.
\end{equation}
\end{subequations}

% TODO We remark that the two operations so far are invariant to the choice of ray direction.

\subsubsection{Convolutions}

Similar to line integrals, we will compute convolutions along the rays
\[
  \big(K * \mathcal{S}[\Gamma, \ldots](\vec r)\big)(t)
  =
  %\int_{-\infty}^{\infty}K(s) g(\mathcal{P}_\Gamma r(t-s) ) \,\textrm{d}s,
  \int_{-\infty}^{\infty}K(s) \mathcal{S}[\Gamma, \ldots](\vec r(t-s)) \,\textrm{d}s,
\]
where $K$ is some compactly-supported kernel function and $\mathcal{S}[\Gamma, \ldots]$ is one of the RaySense sampling operators.
In the discrete setting, this can be written as weighted sums of
$\mathcal S[\Gamma, \ldots]$, or as band-limited matrix multiplication.
In Sec.~\ref{sec:NN}, the discrete weights associated with discrete convolutions are the parameters of a neural network model.

Note unlike the previous cases, for non-symmetric kernels the orientation of the rays matters.

\subsection{Comparing data sets}

Since the sketch for any point set is a fixed-size tensor storing useful feature information, one might compare two point sets (of potentially different cardinalities) via comparing the RaySense sketches.

A natural idea is to choose a suitable metric to define the distances between the RaySense sketch tensors.
% \del{Then a test data set is labeled ``A'' if it is closest (by the metric)
% to a point set with label ``A''.
% We offer several choices of metric.}

\myparastartsentence{The Frobenius norm of the sketch tensor}
 is suitable if the sketches contain the distance and the closest point coordinates.
For data sampled from smooth objects, such 
information along each ray is piecewise continuous.
Thus, if the sketches are generated using the same set of sampling rays, one may compare the RaySense sketches of different data sets using the Frobenius norm.

\myparastartsentence{Wasserstein distances} are more appropriate for comparison of histograms of the RaySense data, especially when the sketches are generated by different sets of random rays.
The normalized histograms can be regarded as probability distributions.
In particular, notice (Fig.~\ref{fig:hist}) that RaySense histograms tend to have ``spikes'' that correspond to the salient points in the data set; $\ell^2$ distances are not adequate for comparing distributions with such features.

Here we briefly describe the Wasserstein-1 distance, or Earth mover's distance, that we used in this paper. Let $(X,\mu_1)$ and $(\tilde X,\mu_2)$ be two probability spaces and $F$ and $G$ be the cumulative distribution functions of $\mu_1$ and $\mu_2$, respectively. The Wasserstein-1 distance is defined as
\begin{equation*}%\label{def:W1}
    W_1(\mu_1,\mu_2) := \int_{\mathbb{R}} \lvert F(t)-G(t) \rvert \,\mathrm{d}t.
\end{equation*}

%One can also consider using Kullback-Leibler divergence for comparing the histograms.

\mypara{Neural networks:}
One can consider using a properly designed and trained \emph{neural network}.
In \S~\ref{sec:NN}, we present a neural network model, RayNN, for comparing point clouds in three dimensions
based on RaySense sketches.

%\section{Application 1: Point Clouds}

% $\Gamma$ is sampled from some known geometric object.

%   \begin{itemize}
%   \item in $\mathbb{R}^3$: The surface of objects, such as in  ModelNet, ShapeNet
%   \item in $\mathbb{R}^N$: Low dimension shapes embedded in high dimensions, such as helix, etc.
%   \item $S(\Gamma)$ is used to distinguish between different objects, which are themselves sampled by some point set.
%   \item Show that points sampled by RaySense are better than randomly-sampled points.
%   \end{itemize}

%\begin{figure}[tbp]
%\centering
%%\includegraphics[width=0.97\linewidth]{figures/voronoi.pdf}
%\includegraphics[width=0.97\linewidth]{figures/voronoi_two_rays.pdf}
%\caption{\label{fig:voronoi} A simple 2D point set (gray). Singular points, such as the tip of the tail, have larger Voronoi cells (dashed lines) and are more likely to be sampled by RaySense (blue). \Richard{This figure should be moved forward. I would prefer the two lines forming a cross, showing that form the two lines, one can already get a outline of the creature.}}
%\end{figure}

\section{Properties of RaySense}\label{sec:analysis}
In this section, we introduce several notable properties associated with our proposed method. Specifically, we begin by analyzing the characteristics of the method when utilizing a single ray, \emph{i.e.}, $m = 1$. We subsequently proceed to discuss how the RaySense sketch, resulting from using multiple rays, effectively utilizes and inherits the identified traits. Through these analyses, we aim to provide a more rigorous and comprehensive understanding of the properties and potential applications of our proposed method.

\myparastartsentence{Assumptions and notations} for this section:
\begin{itemize}
\item[A1]
The point set $\Gamma$, is a realization of a collection of $N$ i.i.d. random vectors $\{\vec X_i\}_{i=1}^N$, with each $\vec X_i \in \Real^d$.
\item[A2] The probability space induced by the random vector $\vec X_i$ is $(\mathbb{R}^d,\mathcal{F}, \mu)$, 
  where $\mathcal{F}$ is the Borel $\sigma$-algebra on $\mathbb{R}^d$ and $\mu$ is a probability measure.
\item[A3] The induced probability measure $\mu$ is also known as the distribution of $\vec X_i$, with compactly supported Lipschitz density~$\rho$.
\item[A4] The RaySense sketch uses the closest points feature space \eqref{eq:closest_point_feature_space}.
  %\del{(Of course the signature can contain addition information but in this section that would be discarded.)}
\item[A5] In the case that $\vec r(s)$, for some $s$, has more than one nearest neighbor, we will randomly assign one. 
\item[A6] A ray, denoted by $\vec r(s)$, $0 \le s \le 1$, generated by the method introduced in \S~\ref{sec:method}, is given in the embedding space $\mathbb{R}^d$ of $\Gamma$, and the support of $\rho$ is centered.
\item[A7] We assume $\supp(\rho)$ can be covered by a finite union of hypercubes $\{\Omega_j\}_{j}$ in $\mathbb{R}^d$, each of non-zero probability measure, i.e., $\supp(\rho)\subset \bigcup_{j} \Omega_j$,
with $P_{\Omega_j} = \int_{\vec x\in \Omega_j} \rho(\vec x) \mathrm{d}\vec x > 0$, and $\{\Omega_j\}_{j}$ overlap with each other only on sets of measure zero.
\end{itemize}
Note that by \blue{A2-A3}, we may regard $\rho$ as representing the density
of a solid body in $\mathbb{R}^d$.
In other words, $\Gamma$ does not consist of samples from a lower
dimensional set.
(RaySense can sample much more general sets,
but the line integral analysis (\S~\ref{sec:integral}) and the argument of sampling convex hull (\S~\ref{sec:salient}) of this section may not hold.)

Much of our analysis is based on the \emph{Voronoi cells} associated with each point in the point cloud.
%Let $\Gamma$ denote the point set of cardinality $N$.
\begin{definition}\label{def:voronoi}
The Voronoi cell of $\vec x\in\Gamma$ is defined as
\begin{equation}
    V(\vec x) := \{\vec y \in \Real^d : \mathcal{P}_\Gamma \vec y = \vec x\}.
\end{equation}
%\begin{remark}
  In practice, we \blue{often sample rays $\vec r\sim\mathcal L$ of finite length} as in Fig.~\ref{fig:voronoi}, \emph{i.e.} $\{\vec r_i\}_{i=1}^m\subset B_R(0)$ for some $R$. \blue{Correspondingly, we define the \emph{truncated Voronoi cell}}:   
  \begin{equation} \label{eqn:truncated_voronoi}
      V^R(\vec x) := \{\vec y \in B_R(0) : \mathcal{P}_\Gamma \vec y = \vec x\}.
  \end{equation}
%\end{remark}
\end{definition}
%\Richard{Shall we mention that the interior of $V(\vec x)$ is a uniquely defined open set?}

\subsection{Properties of sampling with a single ray}\label{sec:prop_oneray}
\subsubsection{Sampling points with larger Voronoi cells}\label{sec:vornoi}
\begin{figure}[htbp]
  \centering
  \includegraphics[width=0.8\linewidth]{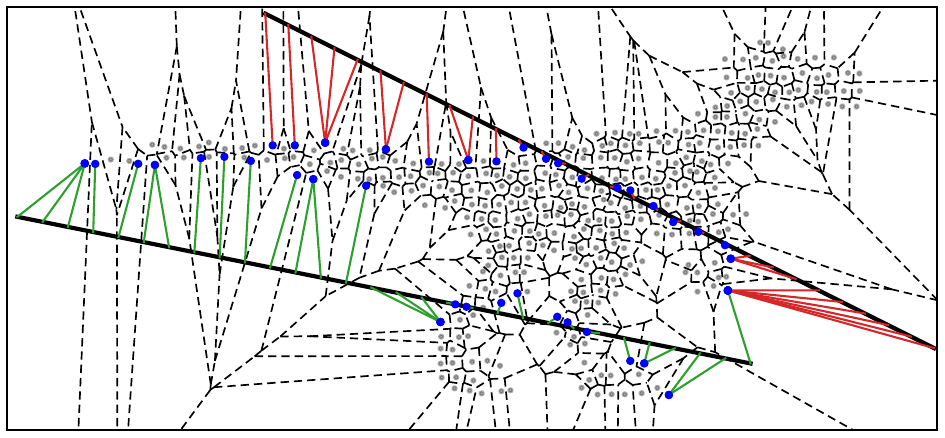}
  \caption{A simple 2D point set (gray).
  Two rays (black) sense nearest neighbors of the point set (blue).
  Singular points, such as the tip of the tail, have larger Voronoi cells (dashed lines) and are more likely to be sampled.
  Closest point pairings are shown in green and red.
  }
  \label{fig:voronoi}
\end{figure}

\blue{For discrete point sets, the likelihood that a ray senses a particular point is closely linked to the size of the Voronoi cell of the point. This relationship arises from the practice of utilizing closest point sampling, which governs the selection of points by a given ray. In this regard, the Voronoi cell of a point in the point clouds is a fundamental geometric construct that plays a key role in determining the probability of detection. This observation is visually depicted in  Fig.~\ref{fig:voronoi}, which demonstrates that points having larger Voronoi cells are more likely to be detected by a given ray.}

We refer to points with relatively large Voronoi cells as the ``salient points'' of $\Gamma$.
\blue{When the probability density for $\Gamma$ is compactly supported,
the salient points of $\Gamma$ tend to be situated in close proximity to geometric singularities present on the boundary of the support of the density; see
Fig.~\ref{fig:voronoi} for a demonstration.}
This saliency characteristic will be further elaborated on in \S~\ref{sec:salient}, where we will demonstrate how it manifests as the biased subsampling property of the RaySense method.

\blue{Furthermore, connections between the probability of a point $\in\Gamma$ being sampled and the size of its Voronoi cell can be made explicit in the following derivation.}
Let $B_R(0)\in \mathbb{R}^d$ be a solid dimension-$d$ ball of radius $R$ containing all the rays,
and $\Gamma \subsetneq B_R(0)$ the finite point set
containing $N$ distinct points. Let $V_k:=V(\vec x_k)$ denote the Voronoi cell for the $k$-th point, $\vec x_k$ in $\Gamma$ as in Def.~\ref{def:voronoi}.
Let $\ell_k(\vec r)$ denote the length of the segment of a ray, $\vec r$, that lies in the truncated Voronoi cell $V^R_k$.
If $\vec r$ does not intersect $V_k$, $\ell_k(\vec r) := 0$. 
Thus, $\ell_k(\vec r)$ is a random variable \blue{indicating how much a ray $\vec r$ senses $\vec x_k\in\Gamma$ }, and we denote its expectation by $\Exp[\ell_k]$; in other words, 
\begin{equation} \label{eqn:intersec_length}
  \Exp[\ell_k] := \int \ell_k(\omega) \,\mathrm{d}\mu_{\mathcal{L}}(\omega) = \blue{ \mathbb E_{\vec r\sim\mathcal L}\Big[\int_r \chi_{V_k} \big(\vec r(s)\big) \mathrm{d}s \Big] }, 
\end{equation}
where $\mu_{\mathcal{L}}$ is the probability measure corresponding to the distribution of lines $\mathcal{L}$ introduced in \S~\ref{sec:signature}, and $\chi_{V_k}$ is the indicator function of the Voronoi cell $V_k$. Consequently, the frequency of a ray $\vec r$ sampling a point $\vec x_k\in\Gamma$ is proportional to the \blue{$d$-dimensional Lebesgue} measure of its Voronoi cell $V_k=V(\vec x_k)$. 

\subsubsection{Sampling consistency}

The Voronoi cell perspective provides a framework to analyze certain properties of RaySense.
We will show in Theorem~\ref{thm:consistency} that
the sampling from a specific ray is consistent
when the number of points $N$ in the point clouds is large enough.
We begin with some lemmas, with proofs appearing in
Appendix~\ref{sec:proof_details_voronoi}.

Our first lemma tells us how large $N$ must be in order to have at least one sample in any region achieving a certain probability measure.

\begin{lemma}\label{thm:nonuniformsampling}
Suppose that $\supp(\rho)$ satisfies assumption A$7$.
% \del{can be covered by a non-overlapping union of finitely many closed hypercubes, \blue{each of non-zero probability measure}; i.e., $\supp(\rho)\subset \bigcup_{j=1}^M \Omega_j$,
% with $P_{\Omega_j} \blue{= \int_{\vec x\in \Omega_j} \rho(\vec x) \mathrm{d}x} > 0$,
% for some $M \in \mathbb{Z}$.}
Let $\Gamma$ be a set of $N$ i.i.d. random samples drawn from $\mu$, and $p_0 \in (0,1)$.
If the number of sample points $N>\nu\big(P\big)$ where $\nu:(0,1]\mapsto\mathbb{R}^+$ is defined by 
% \Richard{Should we write $\nu:(0,1]\mapsto\mathbb{R}^+$?}
\begin{equation}
 \nu\big(P\big) := \frac{\sqrt{\ln{(\frac{2}{1-p_0})}\big(\ln{(\frac{2}{1-p_0})}+2P\big)}+P + \ln{(\frac{2}{1-p_0})}}{P^2},
\end{equation}
then, with probability greater than $p_0$, at least one of the samples lies in every $\Omega_j$ with $P_{\Omega_j}\geq P$.
Additionally, we note bounds for $\nu(P)$:  %\Colin{pulled this into the lemma}
\begin{equation}
  \frac{2\ln{(\frac{2}{1-p_0})}+P}{P^2} < \nu\big(P\big) <\frac{2\ln{(\frac{2}{1-p_0})}+3P}{P^2}.
  \label{eq:nonuniformsampling_htm_bounds}
\end{equation}
\end{lemma}

We notice that \blue{for any fixed $p_0>0$}, $\nu\left(P\right) \sim\mathcal{O}\left(1/P^2\right)$ as $P \to 0$ indicating $\nu\left(P\right)$ is inversely proportional to $P^2$ asymptotically.
This matches with the intuition that
%"fewer i.i.d. samples are needed for sensing larger regions".}\Richard{Is there a better way to say this?} \Lewis{how about:}
more points are needed to ensure sampling in regions with smaller probability measure.
%, which draws a connection from the density function to the size of the Voronoi cell when a set of random samples are given.

The next two lemmas reveal that the volume of the Voronoi cell for a
sample point amongst the others in the point cloud $\Gamma$ decreases to zero with high
probability as the number of sampled points tends to $\infty$.

% \Richard{I propose that in Lemma 2, we change $\vec x_k$ to $x$. The subscript does not seem necessary.}
\begin{lemma}\label{thm:nonuniform_voronoi}
Suppose $\rho$ is $L$-Lipschitz and $\supp(\rho)$ satisfies assumption A7.
% \del{can be covered by a non-overlapping union of finitely many closed hypercubes.}
Given $\Gamma$ a set of $N$ i.i.d. random samples drawn from $\mu$, for $N$ large enough, the size of the Voronoi cell of a sample point $\vec x\in\Gamma$ in the interior of $\supp(\rho)$ is inversely proportional to its local density $\rho(\vec x)$, and with probability at least $p_0$ its diameter has the following upper bound:
\begin{equation*}
  \diam(V(\vec x)) \leq 3\sqrt{d} \left(\frac{21+7\big(9+8N\ln{(\frac{2}{1-p_0})}\big)^{1/2}}{6\rho(\vec x)N}\right)^{\!\!\frac{1}{d}}\!\!.
\end{equation*}
\end{lemma}

% \Richard{Same comment as above about $\vec x_k$.}
When the underlying distribution $\mu$ is uniform,
%\del{$P_{\Omega_j}$ is the same for all $\Omega_j$ of the same \del{size} \blue{Lebesgue measure} in the interior,}
$\rho(\vec x)$ is the same everywhere inside $\supp(\rho)$,
therefore the Voronoi diameter for every $\vec x$ in the interior should
shrink uniformly. However, a better bound can be obtained for
this case, shown in the following lemma.

\begin{lemma}\label{thm:uniform_diameter}
If $\mu$ is a uniform distribution, then given $\Gamma$ with $N$ large enough,
with probability at least $p_0$, the diameter of the Voronoi cell of any sample point in the interior has the bound
\begin{equation*}
  \diam(V) \leq 3\sqrt{d}\left(\left\lfloor\frac{1}{c(N)N}\ln{\frac{N}{1-p_0}}\right\rfloor\right)^{\!\frac{1}{d}}\!\!,
\end{equation*}
with some $c(N)$ such that $c(N) \to 1$ as $N \to \infty$.
\end{lemma}

\begin{theorem}[Consistency of sampling]\label{thm:consistency}
% TODO: fix these hard-coded \ref
Under A$1$-A$7$, suppose $\Gamma_1$ and $\Gamma_2$ are two
point clouds sampled from the same distribution,
with $N_1$ and $N_2$ points respectively, where in general $N_1 \neq N_2$.
Assume further that $\supp(\rho)$ is convex.
For a ray $\vec r(s)$ using $\ppr$ uniformly-spaced discrete
points to sample, for $N = \min(N_1, N_2)$ sufficiently large, the
RaySense sketches in the closest point feature space
$S[\Gamma_1]$ and $S[\Gamma_2]$
%collecting the corresponding sensed points
%$S[\Gamma_1](i,:) = x_{1[i]} $ and $S[\Gamma_2](i,:) = x_{2[i]}$, for
%$i=1,2,\dots,\ppr$
satisfy
\begin{equation}
  \normbig{S[\Gamma_1] - S[\Gamma_2]}_F
  \le \varepsilon\big(N\big),
\end{equation}
where $\varepsilon>0$ and 
% \del{\blue{for any $n_r>0, d>0$ and $supp(\rho$ HOW?}
$ \varepsilon\to 0$ as $N\to\infty$. 
% \Richard{I am a bit nervous about putting the dependence on $\supp(\rho$ here. How does $\epsilon$ depend on it exactly? Is it OK to not include it in (18)?} \Lewis{it's just when dealing with points on boundary we need to do an intersection with the support, and the area of that intersection relates to the threshold probability in the previous lemma, but it is not a crucial dependency, guess we can remove it if we have $N$ there, you can take a look around line 1905~1920}
% \Richard{I think that we can remove it. Particularly since $\rho$ is fixed for this theorem.}
% \Colin{I think I agree: it can go.  I'd say it depends on all sorts of stuff in A1-A7 implicitly, no need to mention that, so same for $supp(\rho)$}
\end{theorem}
\begin{remark}
The assumption that  $\supp(\rho)$ is convex is stronger than needed in many cases;
it excludes the situation where some point $\vec r_{i,j}$ in the discretized ray set is equidistant to two or more points on the non-convex $\supp(\rho)$ that are widely separated, which could lead to an unstable assignment of nearest neighbors. However, in practical scenarios where a ray is chosen randomly,
this situation is unlikely to occur.
Further details can be found at the end of \Cref{sec:proof_details_voronoi}.
\end{remark}

The consistency of sampling ensures the RaySense data on a specific ray would be close when sampling the same object, therefore one can expect a similar property for the RaySense sketch tensor where multiple rays are used, which will be discussed in \S~\ref{sec:salient}.

\subsubsection{Approximate line integrals}\label{sec:integral}
We demonstrate that the RaySense approach enables the computation of line integrals of functions defined on the points in a point cloud. Suppose we have a point cloud $\Gamma$ representing an object in $\mathbb{R}^d$.
%Suppose the point clouds are ``dense'' in the sense that points are drawn according to a density function $\rho(\vec x)$ which is non-zero everywhere inside the object and zero outside.
%  For example, perhaps we have a solid object in 3-d (with e.g., $\rho$ as a scaled indicator function), but not, for example, a thin shell such as a surface in 3-d.
As $N$ increases, the point cloud becomes denser, and for any $\vec r(s)$ lies inside $\supp(\rho)$,
$\mathcal{P}_\Gamma \vec r(s) \approx \vec r(s)$
as the Voronoi cells shrink around each point in $\Gamma$.
If we have a smooth function $g : \mathbb{R}^d \mapsto \mathbb{R}$ evaluated on the point cloud,
%\Colin{would be nice to redo all this with abstract feature space $\mathcal{X}$ here instead of $\mathbb{R}^d$: not today!}
%\Richard{Shall we replace $\mathbb{R}^c$ in line 218 by $\mathbb{R}^d$?}\Colin{No, $c$ is used in the RayNN bits too, where it really is important that $c \neq d$.}
then $g(\mathcal{P}_\Gamma \vec r(s)) \approx g(\vec r(s))$ and we expect that integrals of $g$ along lines can be approximated by integrals of the RaySense sketch
$\mathcal{S}[\Gamma, g]$ introduced in \S~\ref{sec:method_lin_int}
(and quadrature of the discrete RaySense sketch $\mathcal S_{m,\delta r}[\Gamma, g]$).

The following shows that the RaySense integral is
an approximation to the line integral along $\vec r(s)$ provided the point cloud is dense enough.

\begin{theorem} \label{thm:approx_voronoi}
  Suppose that $g\in C(\mathbb{R}^d;\mathbb{R})$ is $J$-Lipschitz,
  ray $\vec r(s) \in \supp(\rho)$ for $0\le s\le 1$,
  and $\Gamma$ is a set of $N$ i.i.d. random samples drawn from $\mu$, with
  corresponding RaySense sketch $\mathcal{S}[\Gamma, g]$,
  then
  the difference between the RaySense integral of $g$  and the line integral of $g$ has the following bound:
  % \begin{equation} \label{eqn:line_error_OLD_DELETE_ME}
  %  \absbig{\int_0^1 g\big(\vec r(s)\big) \mathrm{d}s - \int_0^1  g\big(\vec x_{k(s)}\big)\mathrm{d}s} \le J\sum_{i=1}^M \int_0^1 \chi\big(\{\vec r(s)\in V_{k_i}\}\big)\norm{\vec r(s) -  \vec x_{k_i}} \mathrm{d}s,~~[\del{old}]
  %  \end{equation}
  \begin{equation} \label{eqn:line_error-1}
    \absbig{\int_0^1 g\big(\vec r(s)\big) \mathrm{d}s - \int_0^1  g\big(\mathcal{P}_\Gamma \vec r(s)\big)\mathrm{d}s} \le J\sum_{\vec x\in\Gamma} \int_0^1 \chi_{V(\vec x)}\big(\vec r(s)\big)\norm{\vec r(s) -  \mathcal{P}_\Gamma \vec r(s)} \mathrm{d}s,
  \end{equation}
  where $\chi_{V(\vec x)}$ is the indicator function of the Voronoi cell $V(\vec x)$ for $\vec x\in\Gamma$.
\begin{proof}
  For a fixed number of sampling points $N$, the approximation error is given by
\begin{align*}
    \absbig{\int_0^1 g\big(\vec r(s)\big) \mathrm{d}s - \int_0^1  g\big(\mathcal{P}_\Gamma \vec r(s)\big)\mathrm{d}s} &\leq \int_0^1 \absbig{g\big(\vec r(s)\big) -  g\big(\mathcal{P}_\Gamma \vec r(s)\big)}\mathrm{d}s\\
    &= \sum_{\vec x\in\Gamma}\int_0^1 \chi_{V(\vec x)}\big(\vec r(s)\big)\absbig{g\big(\vec r(s)\big) -  g\big(\mathcal{P}_\Gamma \vec r(s)\big)} \mathrm{d}s\\
    &\leq  J\sum_{\vec x\in\Gamma}\int_0^1 \chi_{V(\vec x)}\big(\vec r(s)\big)\norm{\vec r(s) -  \mathcal{P}_\Gamma \vec r(s)} \mathrm{d}s.
\end{align*}
\end{proof}
\end{theorem}

In scattered data interpolation, the nearest neighbor interpolation
would have an error of $\mathcal{O}(h)$ where $h = \max_k \diam(V_k)$.
Integration of the interpolated data would result in an error of
$\mathcal{O}(h^2)$, consistent with Theorem~\ref{thm:approx_voronoi}.

Intuitively, we expect the RaySense line integral to converge in the limit of $N\to\infty$.
Here we show the corresponding convergence result for the case of uniform density
from the perspective of a Poisson point process \cite{daley2003introduction}.
Details of the proof are given in Appendix~\ref{sec:proof_details_poisson}.

% \Colin{why doesn't the theorem words include ``Poisson point process''?  In our meeting I seem to recall you said its very general... but is it?  Should we say something like ``assume $\Gamma$ is chosen from uniform $\mu$ by a Poisson point process''?}
% \Lewis{not sure if I understand the question, in thm6 we have the uniform distribution and $N$ is fixed; but when using poisson process, $N$ is a random variable; so there is one more step to map results from poisson process back to the fixed $N$. If we say $\Gamma$ is from poisson process then the result is not directly related to the practical setting where the $N$ is usually fixed.}
% \Colin{My question is, does the theorem need to mention ``Poisson point process''?}

\begin{theorem}\label{thm:line_error_conv}
Suppose $g\in C(\mathbb{R}^d; \mathbb{R})$ is $J$-Lipschitz,
  $\rho$ is uniform and satisfies assumption A$7$, and the ray $\vec r(s) \in \supp(\rho)$ for $0\le s\le 1$,
  then given $\Gamma$ a set of $N$ i.i.d. random samples
  \blue{as a realization of a Poisson point process}
%   \del{Assume the ray $\vec r$ intersects a total of $M$ distinct Voronoi cells.}
   with the corresponding RaySense sketch $\mathcal S[\Gamma, g]$ for the ray, 
   the probability that the following holds tends to 1 as $N\rightarrow \infty:$
%   \sout{the line integral  converges in probability to}
%   \Colin{``in probability''?  with high probability?  What is this supposed to say?} \Lewis{it means the event that the inequality does not hold has probability going to $0$ in the limit, it's a weaker form of convergence.}
  \[
\absbig{\int_0^1 g\big(\vec r(s)\big) \mathrm{d}s - \int_0^1   g\big(\mathcal{P}_\Gamma \vec r(s)\big)\mathrm{d}s}
\leq c(d,J) N^{-\frac{1}{d}+\varepsilon(d+1)}\!\!,
\]
for any small $\varepsilon>0$. \\ When $\varepsilon<\frac{1}{(d+1)^2}$, the line integral error converges to $0$ as $N\to\infty$ in a rate $\mathcal O(N^{-\frac{1}{d}})$.
\end{theorem}

\begin{figure}[htbp]
  \centerline{%
  \includegraphics[height=23ex]{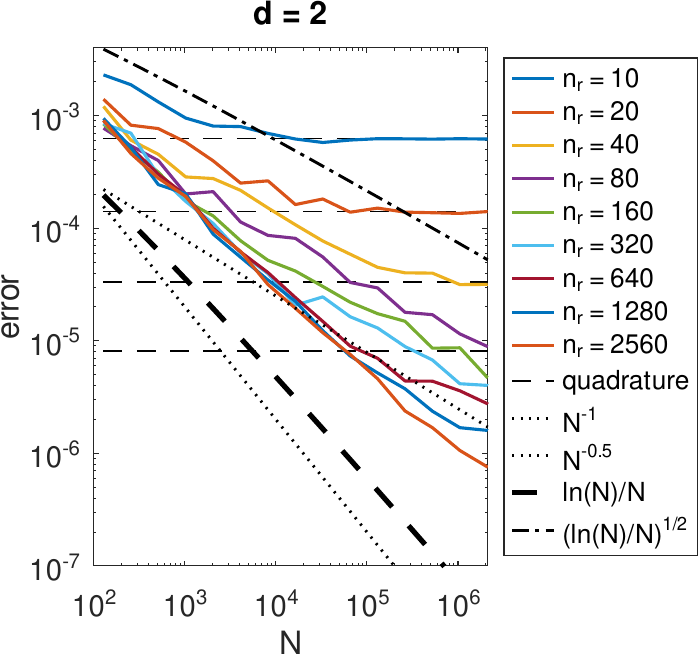}\hfill
  \includegraphics[height=23ex]{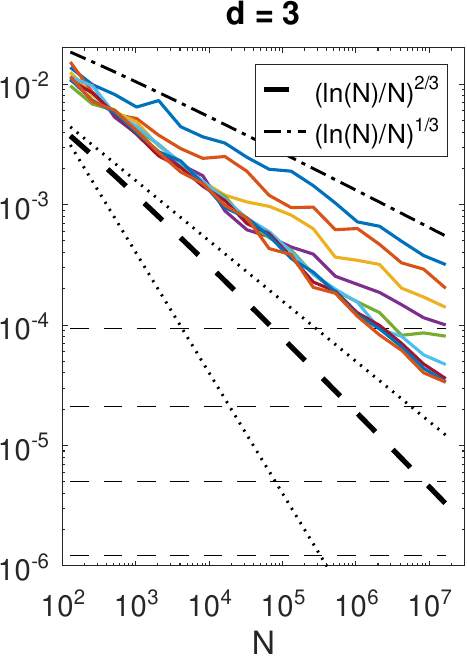}\hfill
  \includegraphics[height=23ex]{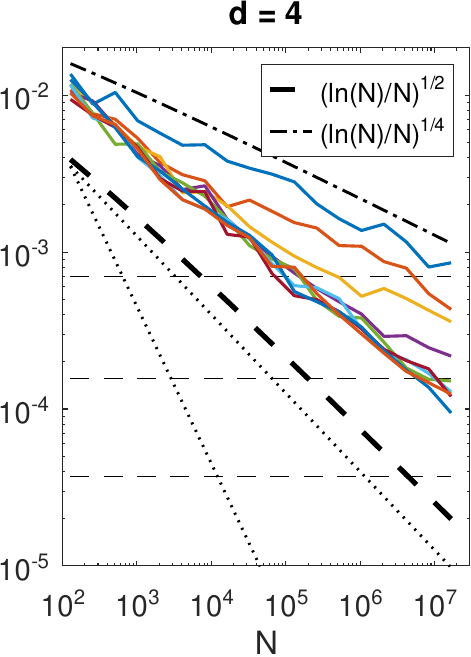}\hfill
  \includegraphics[height=23ex]{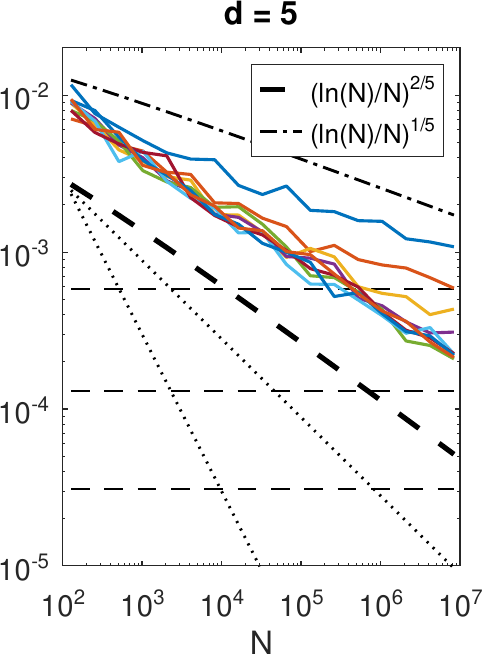}%
  }
  \caption{
    % Colin: Not convinced we want to define "RaySense integral" yet.
    Convergence studies for line integrals approximating from the \mbox{RaySense} sketch on point clouds sampled from a uniform density, in dimensions $d=2,3,4,5$.
    Horizontal dashed lines indicate error inherent to the trapezoidal rule quadrature schemes.
    Diagonal dashed lines indicate different convergence rates.
    }
  \label{fig:integral_convergence}
\end{figure}

We first confirm this rate with numerical experiments.
We return to explore applications of integral transforms in \S~\ref{sec:xray}.
Fig.~\ref{fig:integral_convergence} shows convergence studies of the RaySense integrals for the uniform density case from Theorem~\ref{thm:line_error_conv}.
The 5-d example uses an integrand of $g(x_1,x_2,x_3,x_4,x_5) = \cos(x_1x_2) - x_4x_5\sin(x_3)$
and a line
$\vec r(s) =
\frac{\vec{v}}{\|\vec{v}\|}s
+ \frac{\left\langle 1,1,1,1,1 \right\rangle}{10}$
with $\vec{v} = \left\langle 2,3,4,5,6 \right\rangle$.
    %%     $\left\langle
    %%     \frac{\sqrt{6} t}{9} + \frac{1}{10},
    %%     \frac{\sqrt{6} t}{6} + \frac{1}{10},
    %%     \frac{2 \sqrt{6} t}{9} + \frac{1}{10},
    %%     \frac{5 \sqrt{6} t}{18} + \frac{1}{10},
    %%     \right\rangle$.
In 4-d, 3-d and 2-d, we use $g(x_1,x_2,x_3,x_4,1)$, $g(x,y,z,1,1)$ and $g(x,y,0,0,0)$ respectively, and
drop unneeded components from the line.
The exact line integrals were computed with the Octave Symbolic package~\cite{OctaveSymbolicPkg}
which uses SymPy~\cite{sympy} and mpmath~\cite{mpmath}.
% https://packages.octave.org/packages/symbolic/
Each experiment is averaged over 50 runs.

In Fig.~\ref{fig:integral_convergence}, we see that for each fixed number of sample points $\ppr$ along the ray, the error decreases at the rate discussed above.
From the factor of two in the distance between the results of each fixed $\ppr$, infer a first-order decrease in $\ppr$,
and taken together an overall faster rate of convergence if both $N$ and $\ppr$ are increased.
In all cases in Fig.~\ref{fig:integral_convergence},
the experimental convergence rate is \blue{bounded by  $\mathcal{O}\left(\frac{1}{N}\ln N\right)^{1/d}$ which is asymptotically close to the predicted rate given in \Cref{thm:line_error_conv} since $\varepsilon$ can be taken arbitrarily small.}
However, when $\ppr$ is large, we appear to achieve a faster rate of $\mathcal{O}\left(\frac{1}{N}\ln N\right)^{2/d}$.
This suggests a tighter analysis may be possible in the future.

\subsubsection{Ray not fully contained in $\supp(\rho)$}

For any portion of $\vec r(s)$ that lie outside of $\supp(\rho)$ such portion should take no values when computing the line integral if $g$ is only defined on $\supp(\rho)$. Thus, the post-processing procedure involves eliminating sampling points on $\vec r(s)$ that are outside of $\supp(\rho)$.
When $N$ is large,
this can be accomplished by augmenting the RaySense feature space with vectors to the closest point as in \eqref{eq:augmented_featspace_cpvec} rather than using A4,
and redefining $g(\mathcal{P}_\Gamma \vec r(s)) =0$ if the distance to the closest point of $\vec r(s)$ is beyond some small threshold.

% \Lewis{this is a potential repetition to the previous paragraph? Also is there anything else to say in this section?}

% \blue{
% Therefore, for the remaining discussion we proceed with the assumption that the ray is in the interior of $\supp(\rho)$.}
% \Lewis{will need to modify accordingly once the method section is done}
% \Colin{I don't think this is so easy in practice: its only easy if you're taking a large $N$ limit, otherwise the choice of threshold is probably tricky.  OTOH, what is probably easy is some asymptotics of that threshold, like how it relates to $\delta r$.  TODO: a slight rewording rather than getting into this here!} \Lewis{yes, I was thinking about asymptotic limit here, would the added red text work?}

% \del{The RaySense integral along a ray $\vec r(s)$, $0 \le s \le 1$ is defined as:}
% \begin{equation}
%   \del{\int_0^1 g\big(\mathcal{P}_{\Gamma}\vec r(s)\big) \mathrm{d}s = \int_0^1 g\big(\vec x_{k(s)}\big) \mathrm{d}s,~~~\vec x_{k(s)}:=\mathcal{P}_{\Gamma} \vec r(s)\in\Gamma,}
%   \label{eq:raysense_integral}
% \end{equation}
% \del{where $k(s) \in \{1,2,\cdots,N\}$ indexes the nearest neighbor of $\vec r(s)$ in $\Gamma$.}

\subsection{Properties of sampling with multiple rays}\label{sec:prop_manyray}
When applying RaySense to the point set $\Gamma\in\mathbb R^d$ with $m$ rays and
$\ppr$ points on each ray, the RaySense sketch $S(\Gamma)$ is an $m\times \ppr \times d$ tensor defined
in \eqref{eq:discrete_sig_simple}.
In the following sections, we investigate properties of the RaySense sketch as an extension of the properties of RaySense using a single ray previously discussed.

\subsubsection{Biased samplings toward salient points}
\label{sec:salient}

\begin{figure}[htbp]
  \centering
  \includegraphics[width=0.7\linewidth]{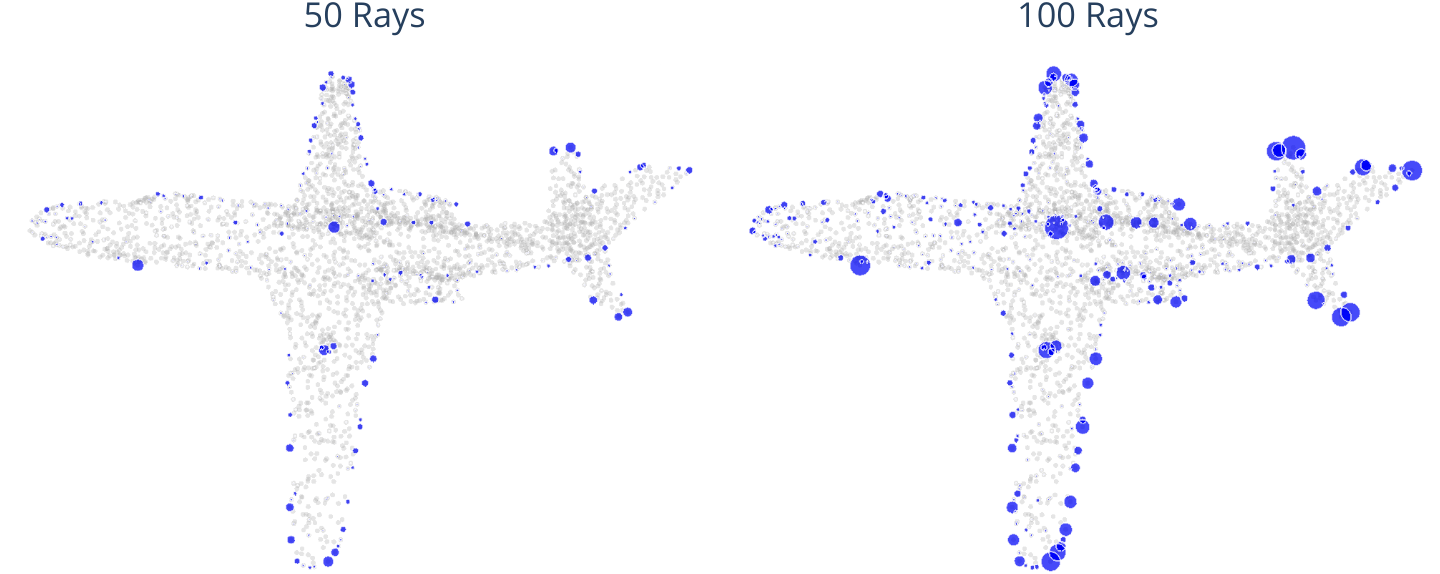}
  \caption{Points sampled by RaySense using different number of rays are indicated as blue dots. The larger blue dots correspond to points that are more frequently sampled. The effect of sampling saliency becomes more apparent as the number of rays increases. Each ray contains 30 sample points.
    \label{fig:airplane}}
\end{figure}

A direct consequence from \S~\ref{sec:vornoi} is that the RaySense sketch $S(\Gamma)$
tends to repeatedly sample points with larger Voronoi cells,
indicating a bias toward these salient points.
  Fig.~\ref{fig:airplane} demonstrates this property by visually presenting
  the frequency of the sampled points by the size of the plotted blue dots.

Another notable feature from Fig.~\ref{fig:airplane} is that the biased subsample generated by RaySense also depicts the outline of the object.
This is because points with larger Voronoi cells usually situate near the boundary and regions with positive curvatures. The following proposition gives a sufficient condition to identify such points:
\begin{proposition}\label{prop:convex_hull_is_salient}
  Vertices of the convex hull of the point cloud $\Gamma$ will be frequently
  sampled by RaySense, when using sufficiently long rays.
    % Vertices of the convex hull of the point clouds $\Gamma$ are salient points of RaySense with a suitable ray-length.
    % \Lewis{Or this version: Vertices of the convex hull of the point clouds $\Gamma$ will be frequently sampled by RaySense with a suitable ray-length.} \Richard{"suitable length" changed to  sufficiently long rays segments?}
\begin{proof}
   Because $\supp(\rho)$ is compact,
    the Voronoi cell for vertices on the convex hull are unbounded. Therefore, one can make the volumes of their truncated Voronoi cell \eqref{eqn:truncated_voronoi}, as large as desired by using rays with suitable length. And recall from \S~\ref{sec:vornoi} that the frequency of being sampled is closely related to the measures of the Voronoi cell.
    % \Colin{Is it possible to label ``proof'' something which involves something we have no precisely defined?  Namely ``saliency''...} \Lewis{in \Cref{sec:vornoi} I describe what is a salient point, is that enough to make it a definition? I didn't do that because The size of the Voronoi cell depends on your "Truncation"}
\end{proof}
\end{proposition}
\begin{remark}
  The arguments in the proof also implies when the length of the line segment $\to\infty$, with high probability RaySense is sampling mostly the convex hull of the point sets when $\supp(\rho)$ is compact.
  This can be viewed as a different approach to approximate convex hulls using rays and curvatures from \cite{graham2017approximate}.
\end{remark}

\begin{figure}[htbp]
\centering
\includegraphics[width=0.8\linewidth]{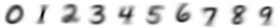}\\
\includegraphics[width=0.8\linewidth]{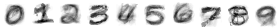}
\caption{Each digit averaged over the entire data set (top) versus those sampled by RaySense (bottom).  \label{fig:mnist}}
\end{figure}

Prop.~\ref{prop:convex_hull_is_salient} applies to abstract datasets as well; 
we consider the MNIST dataset \cite{lecun1998mnist}, treating each image as a point in $d=784$ dimensions.
Here $\Gamma$ is the point set consisting of all images of the same digit.
Fig.~\ref{fig:mnist} shows the average digits over the whole dataset, versus the average of those sampled by RaySense.

In the context of MNIST, salient points are digits that are drawn using less typical strokes (according to the data). 
These are the data points that may be harder to classify, since they appear less frequently in the data. 
RaySense may be used to determine the most \emph{useful} data points to label, as in active learning~\cite{settles2009active}.
RaySense also provides a special notion of importance sampling based on the notion of saliency described above.
An application of such a property is further discussed in \S~\ref{sec:salient_mnist}.

\subsubsection{Invariant histogram}
\label{sec:convergence-of-histograms}

The relationship between the  frequency of a point in $\Gamma$ being sampled and the measure of its Voronoi cell derived in \blue{\S~\ref{sec:vornoi} also provides a crucial guarantee. It ensures the RaySense histogram introduced in \eqref{eqn:histogram} possesses a well-defined limit as the number of rays $m\to\infty$. This convergence can be better elucidated using an alternative formulation of $H_k$ from \eqref{eqn:histogram}, which incorporates the concept of Voronoi cell:}

Draw $m$ rays, $\vec r_1, \vec r_2,\ldots, \vec r_m,$ from the distribution $\mathcal{L}$ \blue{that are contained in some $B_R(0)$ of suitable $R$}. Enumerate this set of points by $\vec r_{i,j}$, and the spacing between two adjacent points is given by $\delta r$. The closest point of $\vec r_{i,j}$ is $\vec x_{k}$ if $\vec r_{i,j}\in V_{k}:=V(\vec x_k)$. Therefore, the bin value of the RaySense histogram can also be given as:
\begin{equation*}
  H_k = H\big(\vec x_k; \mathcal S_{m,\delta r}(\Gamma)\big) := \sum_{i=1}^m \sum_{\vec r_{i,j}\in V_{k}} 1,
\end{equation*}
where $\mathcal S_{m,\delta r}(\Gamma)$ denotes the closest point sketch tensor using $m$ rays and $\delta r$ spacing.  The following Theorem shows that under proper normalization, $H_k$ can be thought as a hybrid Monte-Carlo approximation to $\mathbb E[\ell_k]$ defined in~\eqref{eqn:intersec_length}:
% \begin{equation*}
%   H_k = H\big(\vec x_k; S_{m,\delta r}(\Gamma)\big) := \frac{1}{m}\sum_{i=1}^m \sum_{\vec r_{i,j}\in V_{k}} \delta r.
% \end{equation*}

\begin{theorem}\label{thm:hist_conv}
The normalized bin values $\tilde{H}_k$ with the normalized constant $\delta r/m$, \emph{i.e.},
\begin{equation*}
  \tilde{H}_k := \frac{\delta r}m H\big(\vec x_k; \mathcal S_{m,\delta r}(\Gamma)\big) = \frac{1}{m}\sum_{i=1}^m \sum_{\vec r_{i,j}\in V_{k}} \delta r,
\end{equation*}
have the limit
  \begin{equation*}
    \lim_{\substack{%
      \rule{0pt}{1.3ex}m\rightarrow\infty \\
      \rule{0pt}{1.3ex}\delta r\rightarrow 0}%
    }
    \tilde{H}_k(\mathcal S_{m,\delta r}(\Gamma)) =  \Exp[\ell_k].
  \end{equation*}
  \begin{proof}
    \begin{align*}
     \lim_{\substack{%
      \rule{0pt}{1.3ex}m\rightarrow\infty \\
      \rule{0pt}{1.3ex}\delta r\rightarrow 0}%
    }
    \tilde{H}_k(\mathcal S_{m,\delta r}(\Gamma)) &= \lim_{\substack{%
      \rule{0pt}{1.3ex}m\rightarrow\infty \\
      \rule{0pt}{1.3ex}\delta r\rightarrow 0}%
    }\frac{1}{m}\sum_{i=1}^m \sum_{\vec r_{i,j}\in V_{k}} \delta r = \lim_{\delta r\to 0} \mathbb E_{\vec r_i\sim\mathcal L}\Big[ \sum_{\vec r_{i,j}\in V_{k}} \delta r \Big]\\
    &=  \mathbb E_{\vec r_i\sim\mathcal L}\Big[ \lim_{\delta r\to 0} \sum_{\vec r_{i,j}\in V_{k}} \delta r \Big] = \mathbb E_{\vec r_i\sim\mathcal L}\Big[ \int_{\vec r_i} \chi_{V_k} \big(\vec r(s)\big) \mathrm{d}s \Big] =  \Exp[\ell_k],
    \end{align*}
    where $\chi_{V_k}$ is the indicator function of $V_k$.
    Interchanging order of the limit and the expectation follows from the dominated convergence theorem since $\sum_{\vec r_{i,j}\in V_{k}} \delta r < \chi_{V_k} \big( \vec r(s)\big)+2\delta r$.
\end{proof}
\end{theorem}
Monte-Carlo approximations of integrals converge with a rate independent of the dimension \blue{\cite{caflisch1998monte}}.
Consequently, for sufficiently many randomly
selected rays, 
the histogram is essentially independent of the rays that are actually used.

\begin{figure}[htbp]
\centering
%\subfloat[Diamond]{
%\includegraphics[width=0.97\linewidth]{figures/hist_diamond.pdf}
%\includegraphics[width=0.97\linewidth]{figures/hist_airplane.pdf}
\includegraphics[trim=0 5pt 0 5pt,clip,width=0.49\linewidth]{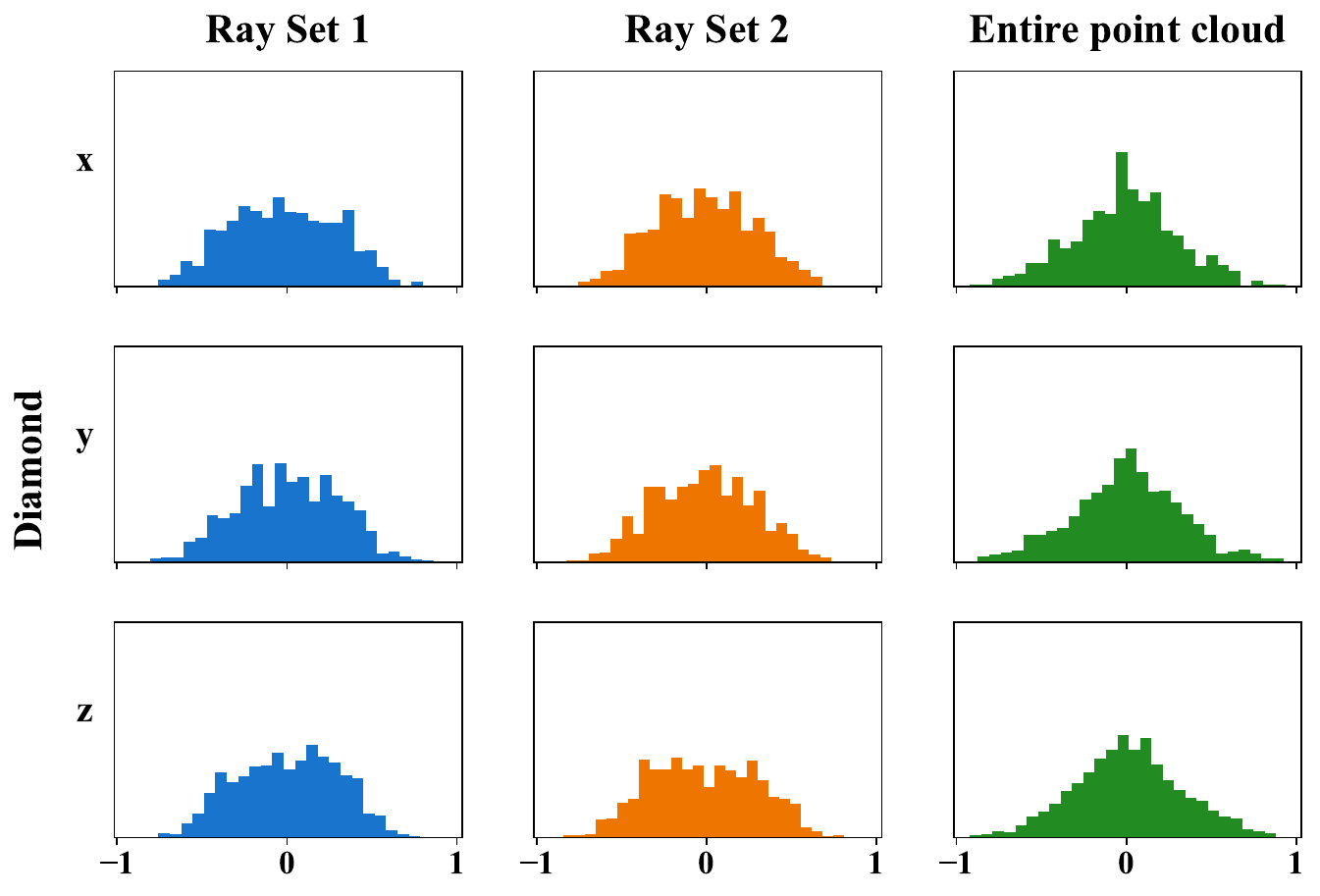}
\includegraphics[trim=0 5pt 0 5pt,clip,width=0.49\linewidth]{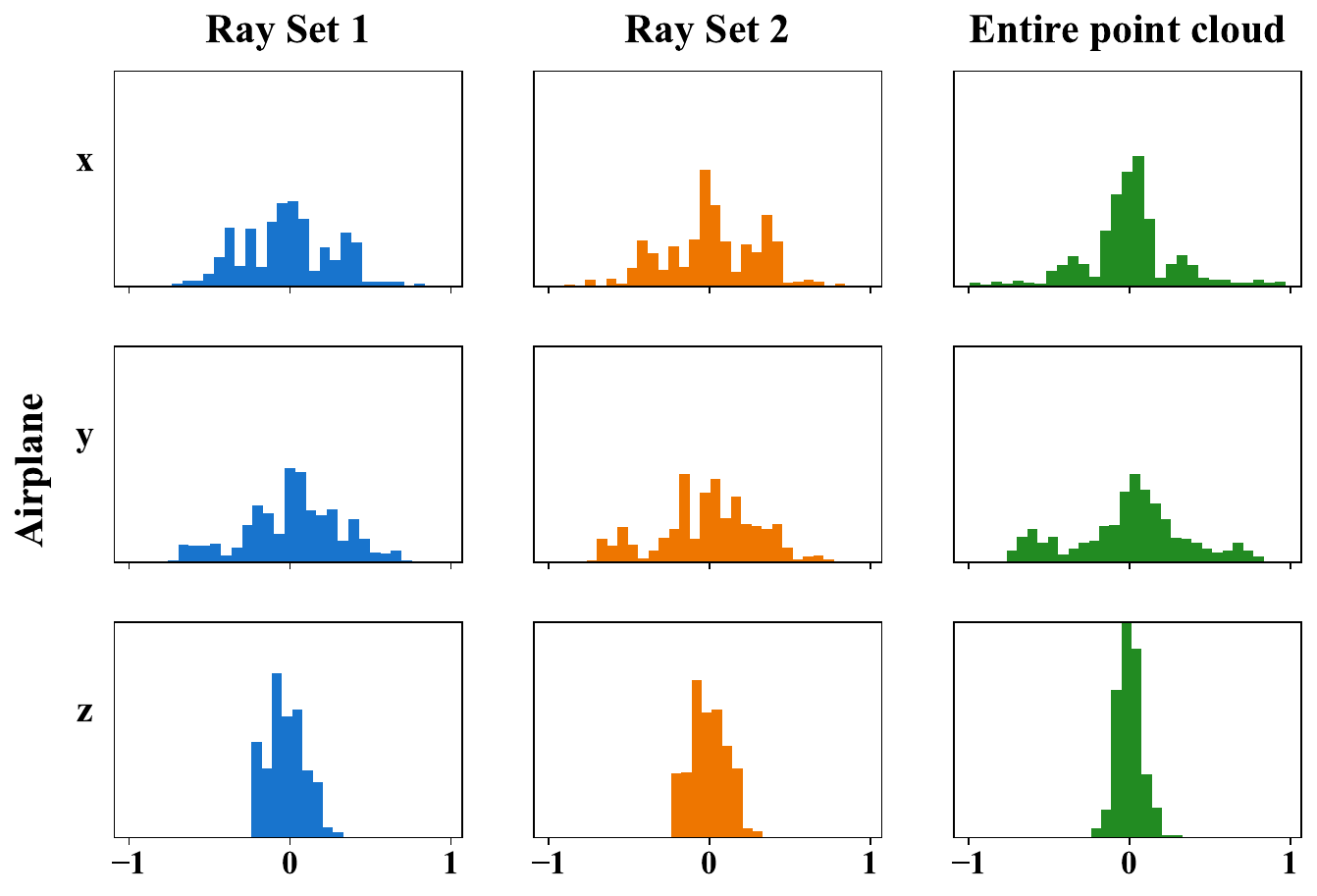}
%%\includegraphics[width=0.97\linewidth,height=16ex]{figures/diamond_hist.png}
%\includegraphics[width=0.97\linewidth]{figures/hist_diamond.pdf}
%}\\
%%\hfill
%\subfloat[Airplane]{
%%\includegraphics[width=0.97\linewidth,height=16ex]{figures/plane_hist.png}
%
%\includegraphics[width=0.97\linewidth]{figures/hist_airplane.pdf}
%}
  \caption{Histogram of coordinates from two point sets.
    Columns 1 and 2
    correspond to 2 different sets of rays, each containing $256$ rays and $64$ samples per ray.
    These histograms are similar for the same object and different for different objects.
    Column 3 corresponds to the entire point cloud; these differ from the RaySense histograms especially for the airplane which is not as regular as the diamond.
  }
\label{fig:hist}
\end{figure}

Similar arguments show that the sampling of
any function of the data set will be independent of the actual ray set, since the histograms are identical in the limit.
More precisely, suppose $g: \vec x\in\Gamma \mapsto \mathbb{R}$ is some finite function, then
\begin{equation*}
  \lim_{\substack{%
    \rule{0pt}{1.3ex}m\rightarrow\infty \\
    \rule{0pt}{1.3ex}\delta r\rightarrow 0}}
  \frac{1}{m}\sum_{i=1}^m \sum_{\vec r_{i,j}\in V_k} g(\vec x_k) \delta r
  = \Exp[g(\vec x_k)\ell_k].
\end{equation*}
In Fig.~\ref{fig:hist}, we show the histograms of the coordinates of the RaySensed points of $\Gamma$.

\begin{figure}[htbp]
\centering

\includegraphics[trim=0 5pt 0 5pt,clip,width=0.98\linewidth]{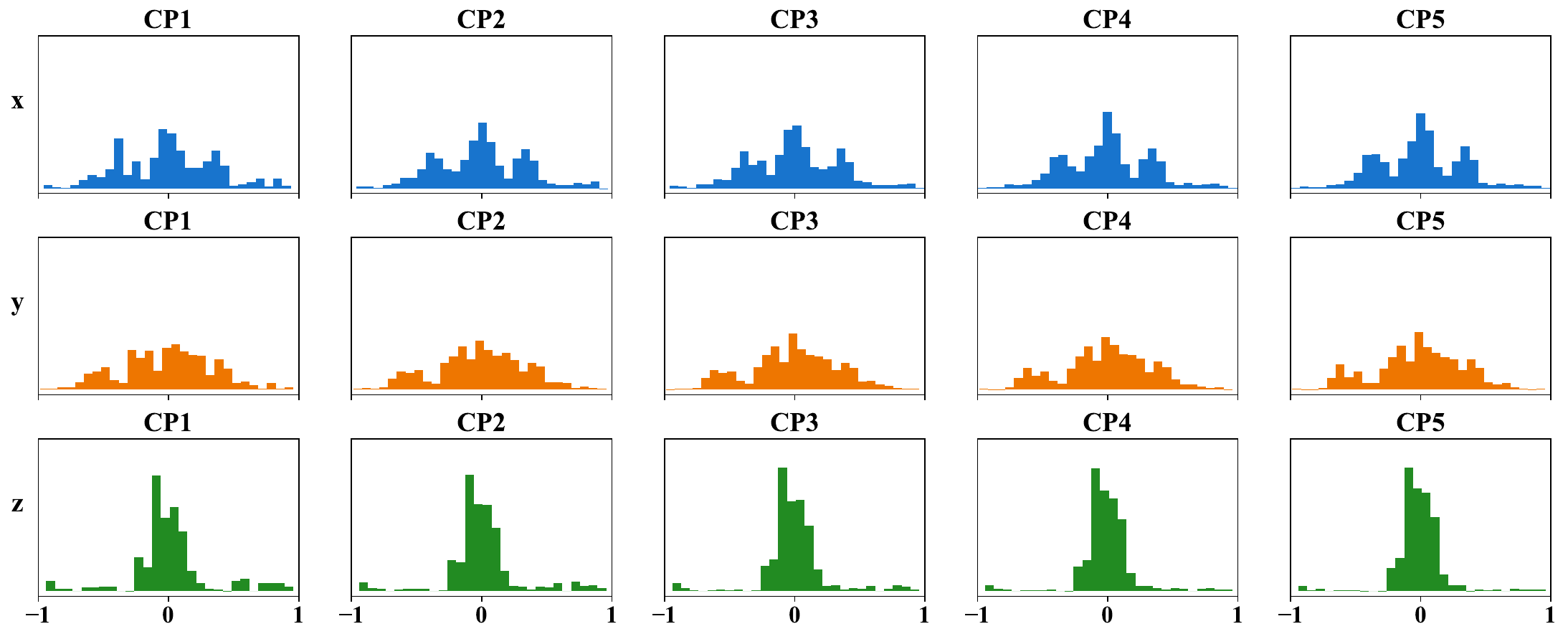}
  \caption{Histogram of the $\eta=5$ nearest neighbors sampled by RaySense, where the underlying point cloud is polluted by $50$ outliers uniformly sampled from the unit ball.
    Rows correspond to different coordinates while columns correspond to different closest points. Outliers introduce extreme values on coordinates of CP$1$ but these effects are significantly mitigated on CP$5$. 
  }
\label{fig:hist_cp5}
\end{figure}

%Since the Voronoi cell depends smoothly on $\Gamma$, 
The integral $\Exp[\ell_k]$ (or $\Exp[g(\vec x_k)\ell_k]$ for continuous $g$) depends smoothly on $\Gamma,$ and is therefore stable against perturbation to the coordinates of the points in $\Gamma$.
However, the effect of introducing new members to $\Gamma$, such as outliers, will be non-negligible. 
One possible way to overcome this is to use multiple nearest neighbors for points on the rays. \blue{In Fig.~\ref{fig:hist_cp5}, we show coordinates  of the $\eta=5$ nearest neighbors sampled by RaySense under the presence of outliers. It is observed that the feature of the $\eta$-th nearest neighbor for $\eta$ large is more robust against outliers while also maintaining the desired histogram information. In \S~\ref{sec:NN}, we will also demonstrate the effectiveness of this idea in dealing with outliers in practical tasks.}

\blue{By considering a similar argument as in the proof of Thm.~\ref{thm:consistency}, we can provide a simple and intuitive explanation for the robustness to outliers when using more nearest neighbors: when the underlying point cloud is dense enough, if an outlier disrupts the nearest neighbor search, excluding the outlier and finding the next few nearest neighbors would mitigate the impact caused by the outlier; if an outlier does not dominate the nearest neighbor search, then the next few nearest neighbors with high probability would also originate from a small neighborhood centered around the first nearest neighbor. Therefore, increasing the number of nearest neighbors  enhances the stability of RaySense.}

\section{Examples of Applications}\label{sec:application}

\subsection{Comparison of histograms of RaySense samples}

\begin{figure}[htbp]
  \centering
  \includegraphics[width=0.6\linewidth]{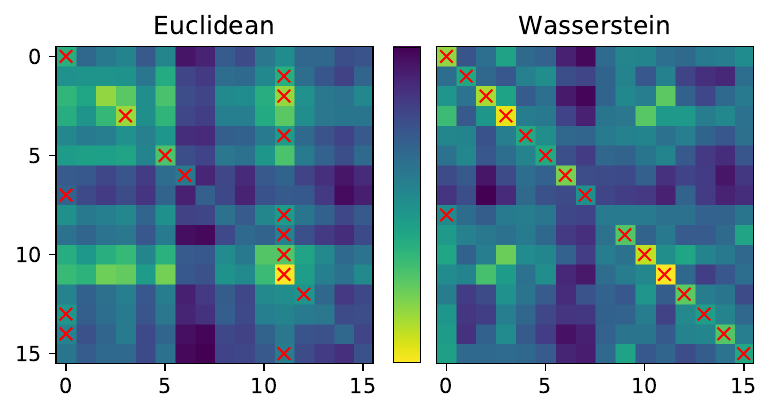}
  \caption{Comparison of histogram of the $x,y,z$ coordinates of points sampled by RaySense, using $\ell^2$ and the Wasserstein distance $W_1$. Rows and columns correspond to object labels. Red $\times$ indicate location of the argmin along each row.}
  \label{fig:hist_compare}
\end{figure}

We experiment by comparing $\Gamma$ drawn from \num{16384} objects of 16 categories from the ShapeNet dataset \cite{chang2015shapenet}.
Let $\beta^i$ be the label for object $\Gamma_i$.
We compute the histogram $h_x^i, h_y^i, h_z^i$ of the $x,y,z$ coordinates, respectively, for points sampled by 50 rays with $\ppr = 10$ samples per ray.
We compare the histograms against those corresponding to other objects in the dataset, using
$$D_{i,j} = d(h_x^i, h_x^j) +d(h_{y}^i, h_{y}^j) + d(h_z^i, h_z^j),$$
where $d(\cdot, \cdot)$ is either the $\ell_2$ or Wasserstein-1 distance.
We sum $D$ according to the respective labels
\begin{equation*}
  M_{a,b} \propto \sum_{i: \beta^i=a} \ \sum_{j: \beta^j=b} D_{i,j}, \qquad a,b = 1,\dots,16,
\end{equation*}
and normalize by the number of occurrences for each $a,b$ pair.
Fig.~\ref{fig:hist_compare} shows the matrix of pairwise distances $M$ between the 16 object categories.

Ideally, intra-object distances would be small, while inter-object distances would be large.
As expected, Was\-ser\-stein-1 is a better metric for comparing histograms.
Still, not all objects are correctly classified.
When comparing histograms in not sufficient,
we consider using neural networks to learn more complex mappings,
such as in \S~\ref{sec:NN}.
% \sout{one may consider using
% higher-order statistical information or neural networks to learn more
% complex mappings between the data and label} (see \S~\ref{sec:NN}). \Lewis{reviewer: speculative statement about H.O.S. and Neural Network}

\subsection{Salient points in the MNIST data}\label{sec:salient_mnist}

From previous discussion and simulation in \S~\ref{sec:vornoi} and \S~\ref{sec:salient}, we know RaySense has the ability to detect salient points or boundary points.
Here we provide further visualization of RaySense salient points on the MNIST dataset.

By vectorizing the MNIST image, each image is a vector in $\mathbb{R}^{784}$ with pixel value from $0$ to $255$.
We generate the random ray set in this ambient space using the method R1 in Appendix~\ref{sec:method},
where each ray has the fixed-length $1$,
with centers uniformly shifted in the half cube $[-\frac{1}{2},\frac{1}{2}]^{784}$.
Each ray set has $m=256$ random rays, with $\ppr = 64$ equi-spaced points on each ray.
To ensure a good coverage over the data manifold, we rescale the MNIST image by entry-wise dividing so that each data point is constrained in an $\ell^{\infty}$ ball of a certain radius as introduced below; we also shift the dataset to have mean $0$.

\begin{figure}[htbp]
  \centering
  \hfill
  \subfloat[normalized to $0.1$ cube]{\includegraphics[width=0.33\textwidth]{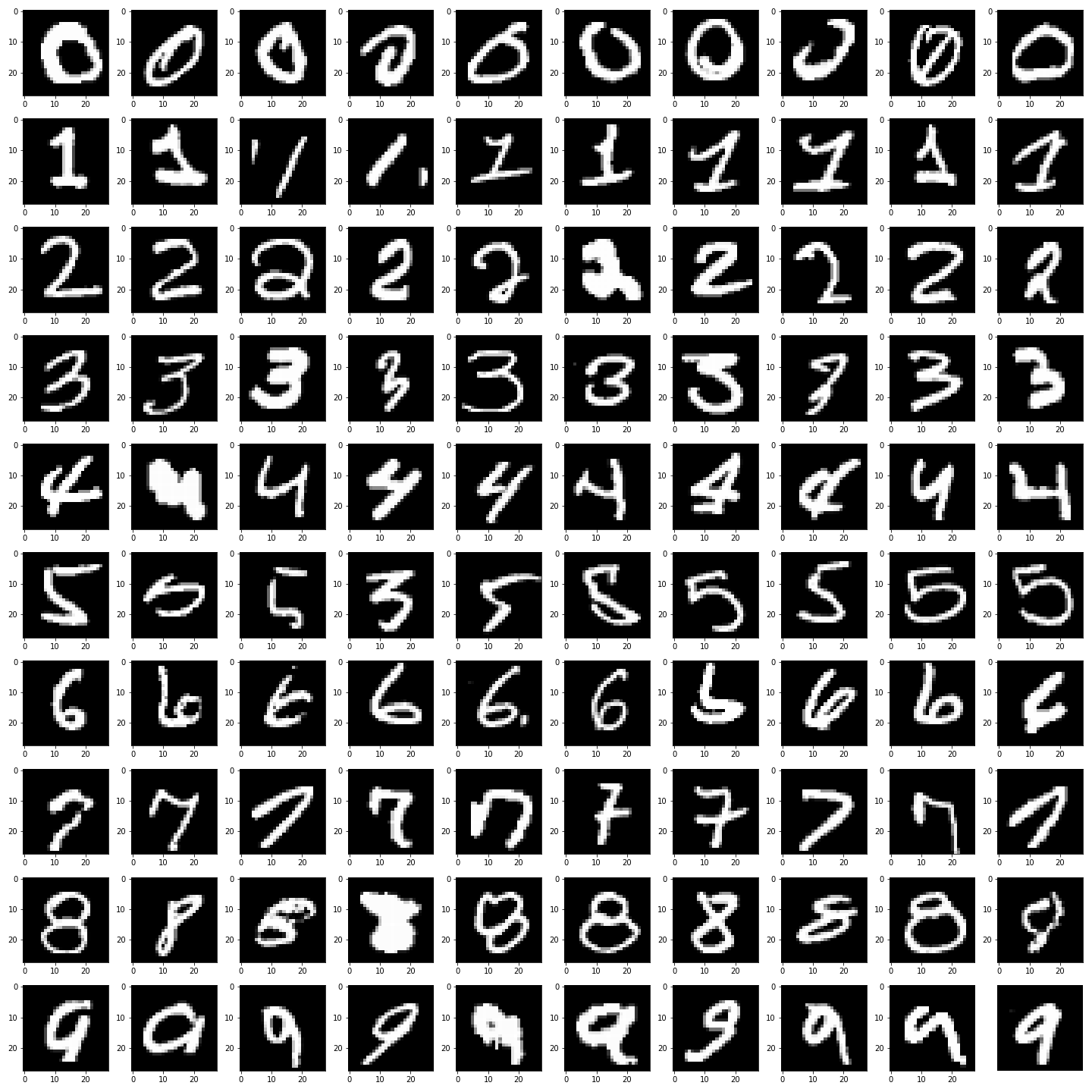}} \hfill
  \subfloat[normalized to $0.05$ cube]{\includegraphics[width=0.33\textwidth]{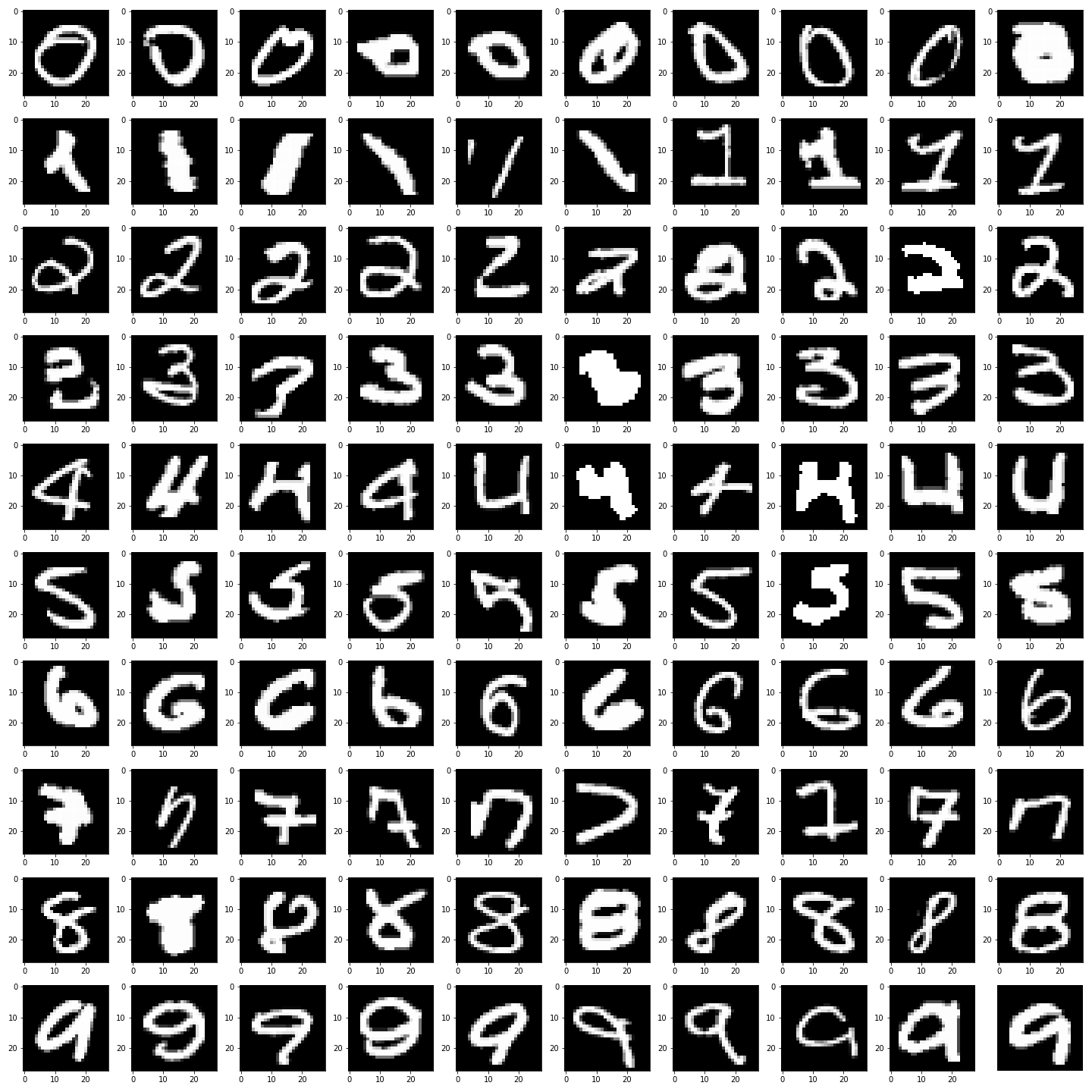}} \hfill\phantom{.}\\
  \hfill
  \subfloat[normalized to $0.01$ cube]{\includegraphics[width=0.33\textwidth]{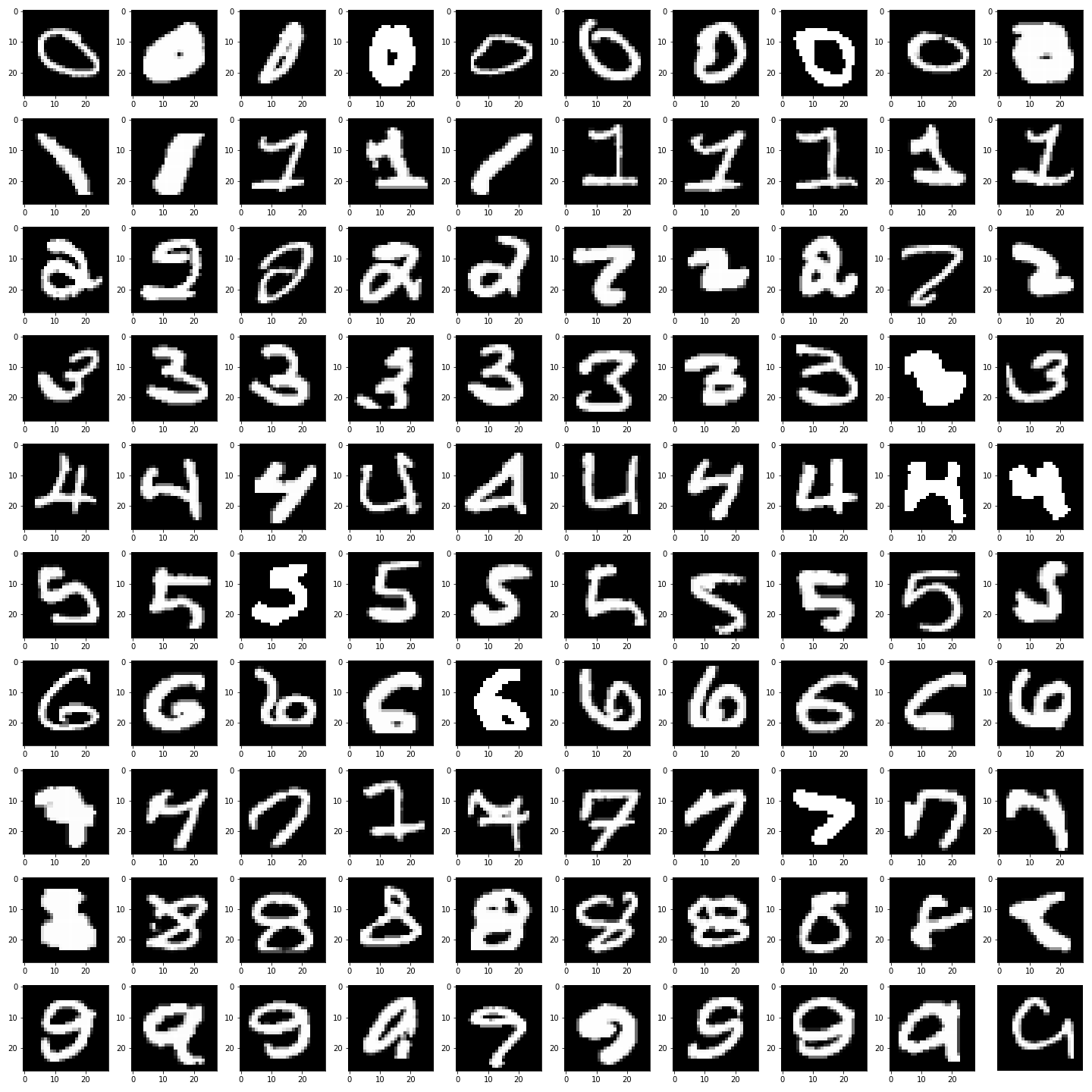}} \hfill
  \subfloat[uniformly random]{\includegraphics[width=0.33\textwidth]{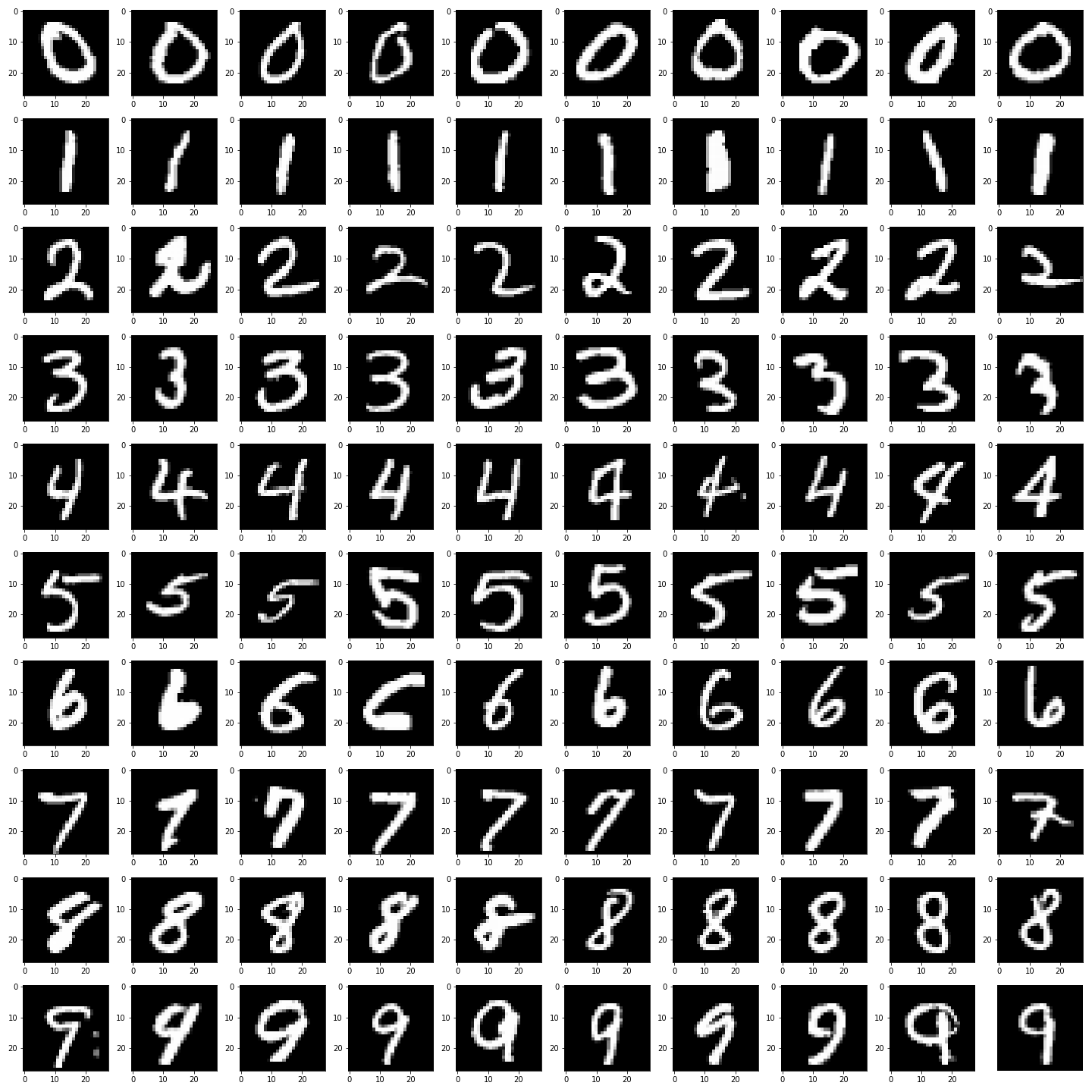}} \hfill\phantom{.}
  \caption{%
  MNIST digit images with highest RaySense sampling frequencies for each class.
  Three different normalizations are shown in (a), (b) and (c).
  Compared to a uniformly random subsample (d), we see a wider variety of hand-writing styles in the RaySense output.
  }
\label{fig:mnist_top10}
\end{figure}

As mentioned, the RaySense salient points are those in $\Gamma$ sampled most frequently by points from the rays.
We record the sampling frequency for each MNIST image, and in Fig.~\ref{fig:mnist_top10} we plot the top-$10$ images with highest frequency for each class.
From the figure, we see that the salient points often correspond to digits with untypical strokes and writing styles, similar to the conclusion obtained from Fig.~\ref{fig:mnist}.
Fig.~\ref{fig:mnist_top10} further shows that different normalizations of the data (by using scaling values $2550$, $5100$ and $25500$) also affects the sampling.
%\del{, suggesting some kind of multi-resolution analysis may be possible.}
\blue{This phenomenon can be better understood from the perspective of truncated Voronoi cell \eqref{eqn:truncated_voronoi}: when the scale of the point clouds shrinks while the length of rays remains constant, it has a similar effect as increasing the length of the rays while keeping the point clouds unchanged, causing the truncated Voronoi cells to grow. Specifically, the truncated Voronoi cells associated with salient points exhibit a larger growth rate, as their Voronoi cells are typically unbounded, \emph{e.g.} Prop.~\ref{prop:convex_hull_is_salient}, making the subsampling even more biased when normalized to a smaller cube.}

%\blue{maybe more interpretations can be added here}

\subsection{RaySense and integral transforms}\label{sec:xray}

A line $\vec r$ in $\mathbb{R}^d$ in the direction of $\vec{\theta}\in\mathbb{S}^{d-1}$ has parameterization $\vec r(s) = \vec b + s\vec{\theta}$, $s\in(-\infty,\,\infty)$, with $\vec b \in \mathbb{R}^d$ a reference point on the line.
Without loss of generality, let $\vec b$ be in $\vec{\theta}^{\perp}$, which is a hyperplane orthogonal to $\vec{\theta}$ passing through the origin.
The X-ray transform for a non-negative continuous function $g$ with compact support \cite{Natterer:computerized_tomography_siambook} is defined on the set of all lines in $\mathbb{R}^d$ by
\begin{equation}\label{eq:Xray-trans}
  %\del{\mathcal{X}[g](L) \equiv}
  \mathcal{X}[g](\vec b,\vec{\theta}) :=
  %\del{\int_{L} g :=}
  \int_{-\infty}^{\infty} g(\vec b + s\vec {\theta}) \mathrm{d}s.
\end{equation}
The spectrum of $g$ can be obtained via the Fourier slice theorem~\cite{solmonXRay1976}:
% In brief, for $\vec{\theta}$ fixed, let $X_{\theta}g$ denotes the X-ray picture taken in the direction $\vec{\theta}$, the Fourier transform of the X-ray picture in the direction of $\vec{\theta}$ equals to the $d-1$-dimensional slice of Fourier transform of $g$ under $\vec{\theta}^\perp$:
% L_{\theta}\rho(\vec x) = \int_{-\infty}^{\infty} \rho(x + t\theta) \mathrm{d}t,
% Two interesting results, see e.g.~\cite{solmonXray1976}), about X-ray transform: 1) If $\rho$ has compact support then $\rho$ is uniquely determined by an infinite set of rays in all direction.
% 2) If only finitely many rays are given, $\rho$ can have arbitrary behavior on a compact subset in the interior. However, this is due to the lack of resolution which occurs naturally in any applications. \cite{maass1987x} and \cite{louis1986incomplete} showed it is still possible to reconstruct part of the spectrum of the underlying density by analyzing the SVD of the X-ray transform operator. 
\begin{equation*}
    \mathcal{F}[\mathcal{X}g](\vec{\theta}, \vec{\xi}) = \mathcal{F}[g](\vec{\xi}), \quad \vec{\xi}\in\vec{\theta}^{\perp}.
\end{equation*}
When we restrict $\vec{\xi}$ to be only on a line in $\vec{\theta}^{\perp}$, we are effectively collecting information on a $2$-dimensional slice of $g$ parallel to $\vec{\theta}^{\perp}$.
%Therefore, we can obtain spectral information of slices of the density $g$ in arbitrary dimension by using just parallel rays along a fixed direction to perform X-ray transform. 

However, when the function $g$ only has a sampling representation, e.g., a point cloud, it is non-trivial to compute such integrals.
In Section~\ref{sec:integral}, we showed that if $\vec r$ is a member of the sampling ray set, one can compute an approximation of \eqref{eq:Xray-trans} from the RaySense sketch obtained from $\{\vec x_k, g(\vec x_k)\}_{k=1}^N$, where $\{\vec x_k\}$ are i.i.d.\ samples from a known probability density~$\rho$.
Thus, RaySense provides a convenient alternative in obtaining (or, in a sense, defining) the Fourier slices of the discrete data set $\{\vec x_k, g(\vec x_k)\}_{k=1}^N$.
%\del{The same idea works in higher dimensions which means that one can approximate the X-ray transform from a RaySense sketch using suitable ray sets,
%or, in the random case,}
%% if the ray distribution R1 (Appendix~\ref{sec:method}) is used,
Since \eqref{eq:Xray-trans} is defined for any dimension, one can approximate the X-ray transform from a RaySense sketch using suitable ray sets, or, in the random case, RaySense integrals can be regarded as randomized approximations of X-ray transforms.
%% Colin: I took this out as it sounded speculative..., in a way the reviewer did not like
%\blue{Last but not least, the $2$-d slice information obtained from high dimension would be useful when the underlying density is sparse}.

\begin{figure}[htbp]
  \centerline{%
  \includegraphics[width=0.24\linewidth]{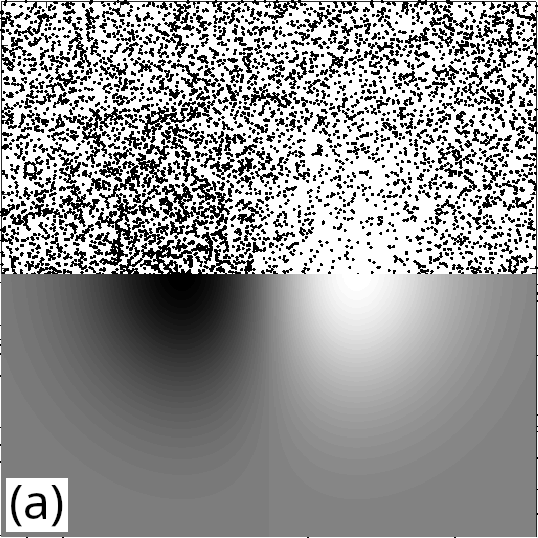}
  \includegraphics[width=0.24\linewidth]{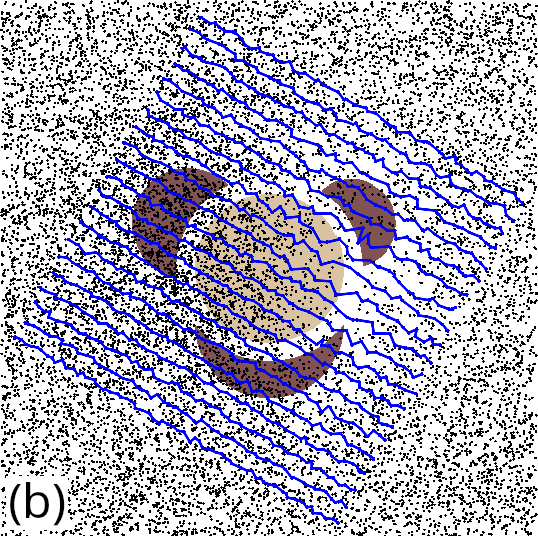}
  \includegraphics[width=0.24\linewidth]{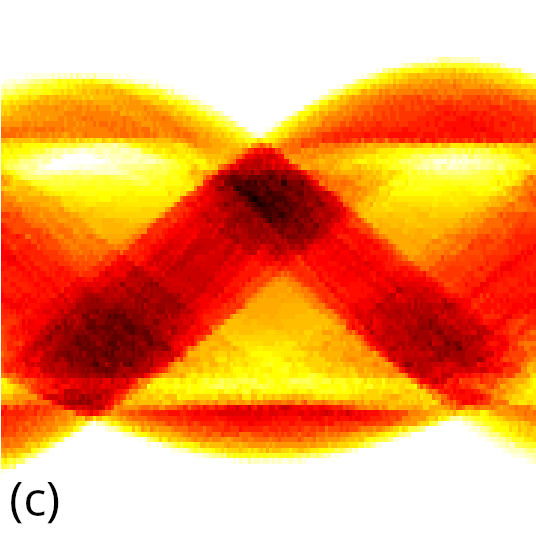}
  \includegraphics[width=0.24\linewidth]{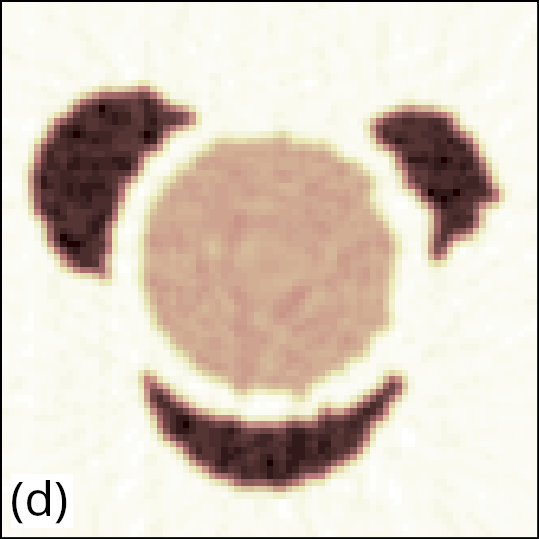}
  }
  \caption{%
    Approximate Radon transform computed with RaySense from point cloud data (a)--(c) and filtered back projection reconstruction (d).
    %\todo{label each subfig}.
  }
  \label{fig:radon}
\end{figure}

In Fig.~\ref{fig:radon} we show an example of using RaySense sampling with prescribed (rather than random) rays to approximate the Radon transform.
In this experiment, a point cloud $\Gamma$ (Fig.~\ref{fig:radon}(a) top) with $15010$ points, is sampled from density
    $\rho = \frac{1}{2} - 3x e^{-9x^2 - 9y^2}$ (Fig.~\ref{fig:radon}(a) bottom)---%
    note denser (darker) region on left and sparser (lighter) region on right.
    $\Gamma$ has data shown in Fig.~\ref{fig:radon}(b) evaluated from the piecewise constant function $g$, shown by the solid colours (for visualization only; $g$ is only known at the discrete points in $\Gamma$).
    Blue lines show the locations of the RaySense sketch for one particular angle (illustrated with 21 rays but the computation  uses 100).  We note increasingly jagged lines to the right where the point cloud is sparser.
    Fig.~\ref{fig:radon}(c) shows that approximate Radon transform computed over 180 degrees in steps of one degree by integrating the RaySense sketch using trapezoidal rule at $\ppr = 64$ points per ray.
    Fig.~\ref{fig:radon}(d) shows the filtered back projection computed by the Octave Image package \cite{OctaveImagePkg}.
    Note a more jagged reconstruction on the right where the point cloud is sparsest.
If we instead used random rays, we could generate samples at scattered points in the sinogram (Fig.~\ref{fig:radon}(c)) which could then be used for an approximate inverse transform.

\subsection{Point cloud registration}\label{sec:registration}

\begin{figure}[htbp]
% \captionsetup[subfigure]{justification=Centering}
\centering
\subfloat[original dense data]{\includegraphics[width=0.3\textwidth]{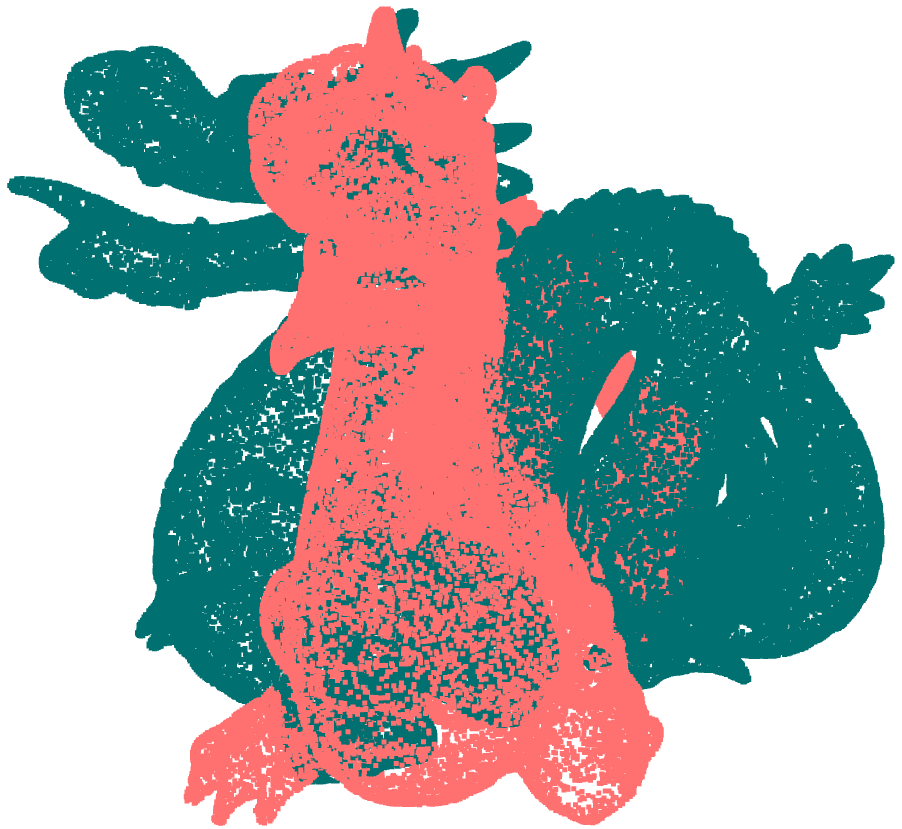}}
\subfloat[random sample]{\includegraphics[width=0.3\textwidth]{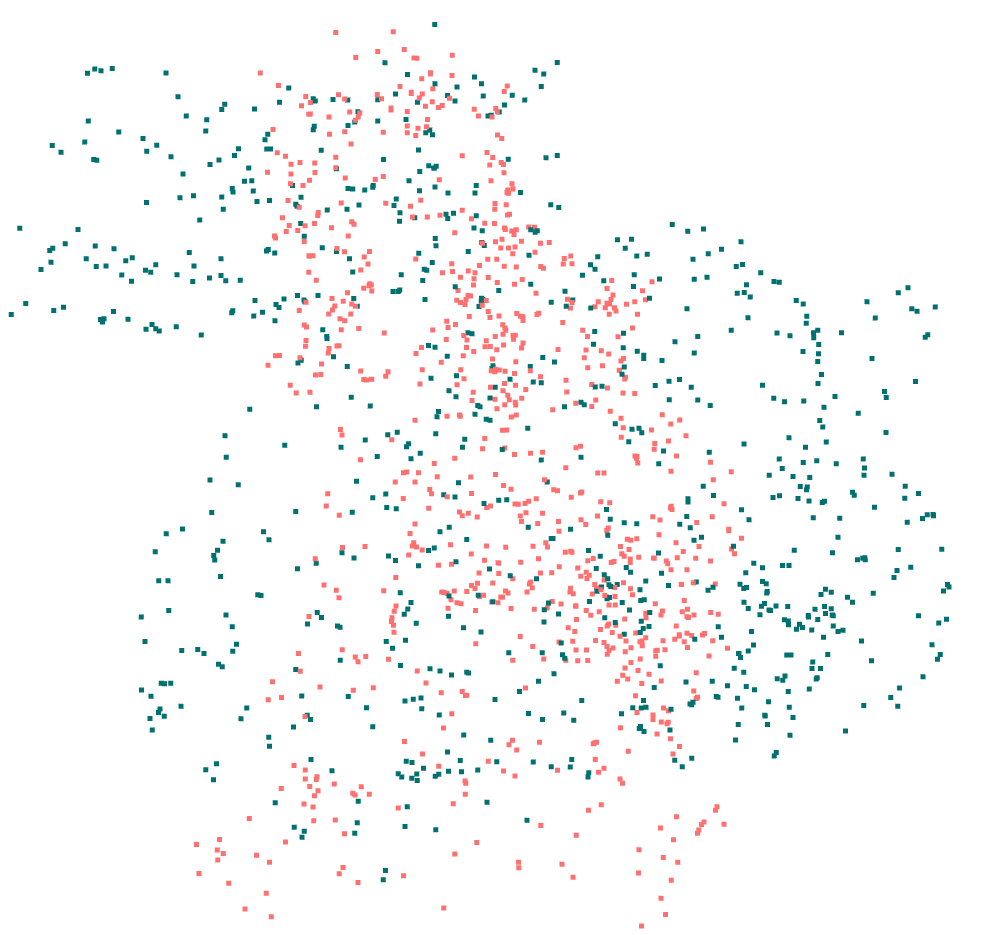}}
\subfloat[RaySense sample\label{fig:registration_salient}]{\includegraphics[width=0.3\textwidth]{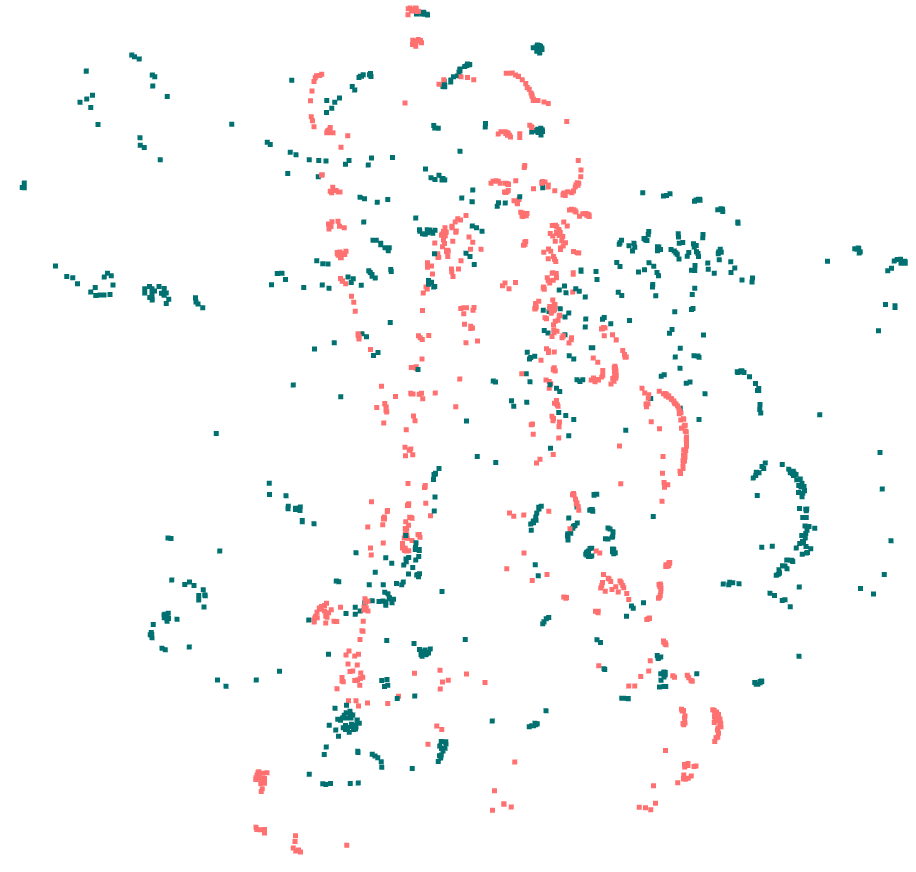}}\hskip1ex
\bigskip % more vertical separation
\subfloat[original dense data]{\includegraphics[width=0.3\textwidth]{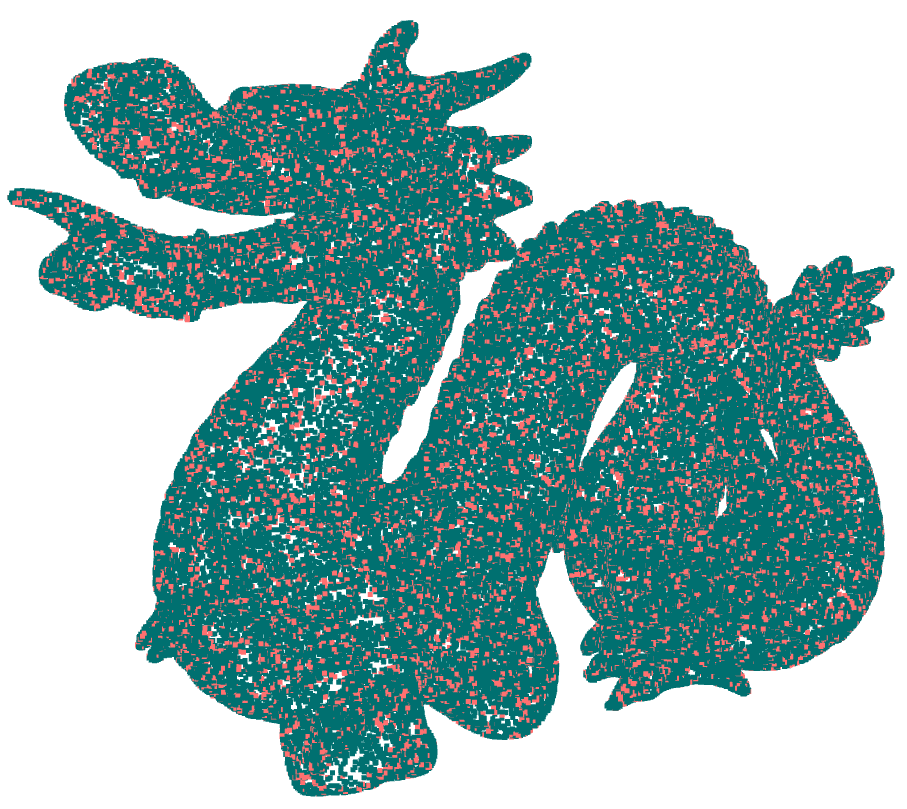}}
\subfloat[random sample]{\includegraphics[width=0.3\textwidth]{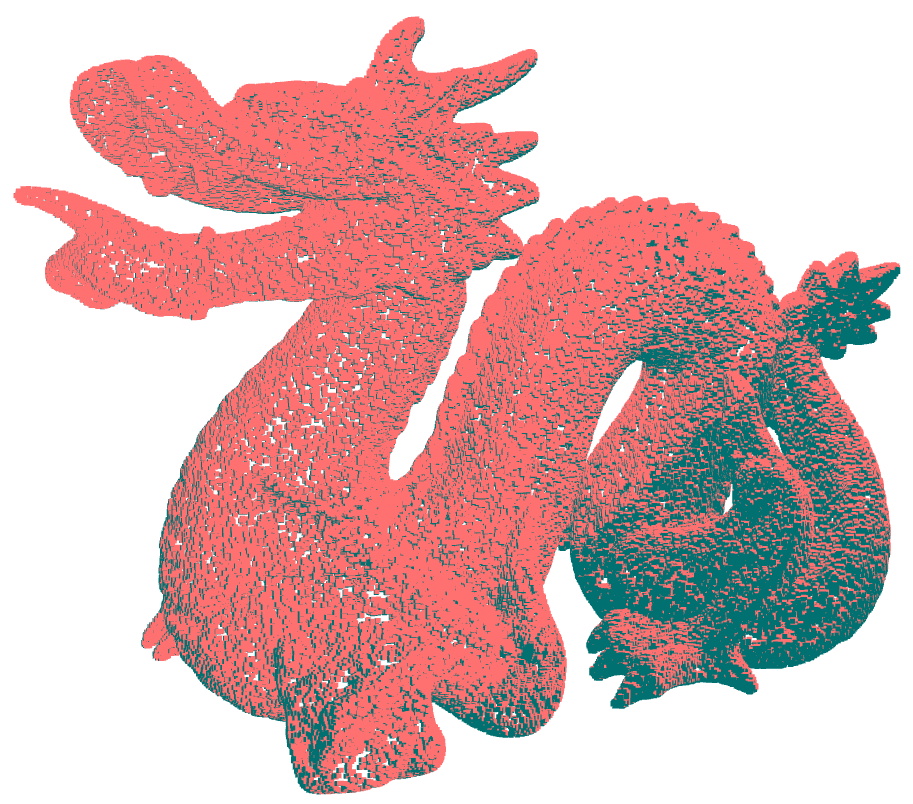}}
\subfloat[RaySense sample]{\includegraphics[width=0.3\textwidth]{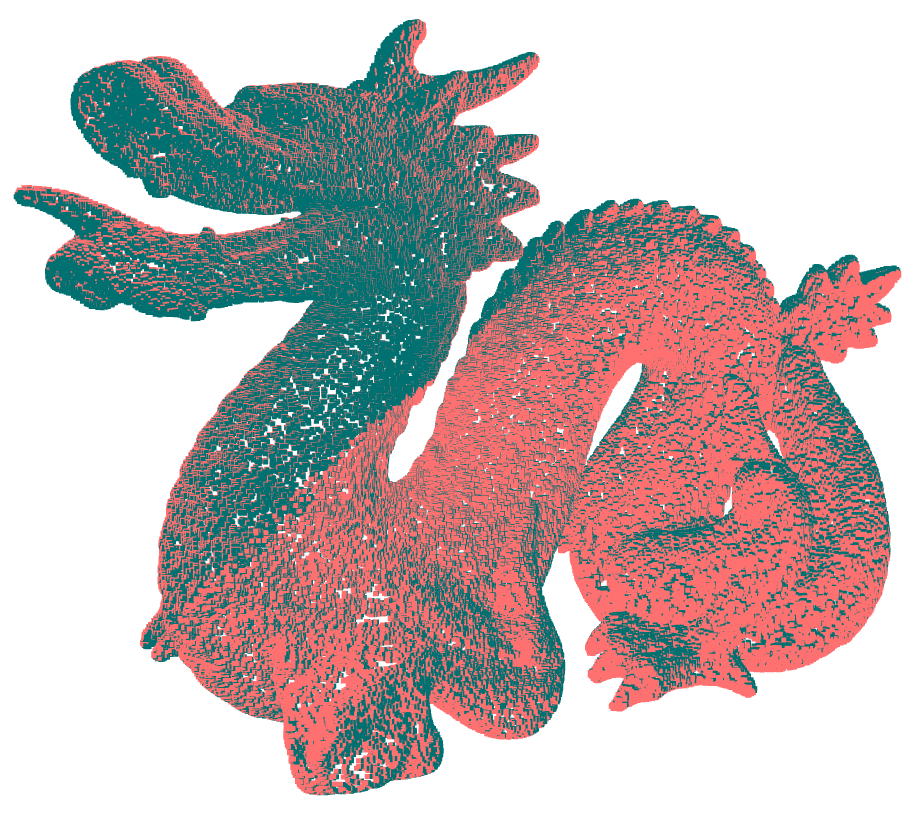}}
\caption{%
  Registration simulation on a rotated Stanford dragon.
  Top row: initial pose and sparse samples; Bottom row: registration results.
  Note that the visualizations (e) and (f) are obtained by applying the
  transformation computed from the sparse samples in (b) and (c)
  to the original dense point clouds.
}
\label{fig:registration}
\end{figure}

In this section, we explore the application of RaySense to the point cloud registration problem.
Given two point sets $\Gamma$ and $\tilde{\Gamma}$ in 3-d consisting of distinct points,
registration aims to find the 3-d rotation matrix $\vec U$ and translation vector $\vec b$
to minimize the Euclidean norm of points in correspondence.
When the correspondence is known, this is the orthogonal Procrustes problem
and the solution can be obtained explicitly via the singular value decomposition.
When the correspondence is unknown, one can formulate an optimization problem and
solve with various carefully-designed algorithms.
Here we choose to use the Iterative Closest Point (ICP)~\cite{besl1992method} due to its simplicity, which minimizes point-wise Euclidean distance iteratively
from the optimization problem
\begin{equation*}
  \min_{\vec U\in SO(3),\vec  b\in \mathbb{R}^3} \sum_{\vec x_j \in \tilde\Gamma} \min_{\vec y\in\Gamma} \|U(\vec x_j+\vec b)-\vec y\|^2_2.
\end{equation*}
We set up the problem using the Stanford Dragon \cite{StanfordDragon_Curless96} as a point cloud $\Gamma$ with $100\,000$ points.
We artificially generate the target point cloud to register by rotating by $\pi/3$ in one direction.
We compare the performance of ICP in three scenarios:
1) the original dense point clouds,
2) a uniformly random subsampling (in index) of the point clouds,
3) RaySense \blue{closest point samples using $\mathcal S[\Gamma]$} (without repetition of sampled points) of each point cloud.
Specifically, we use $m=512$ rays, each with $\ppr = 64$ sample points,
to subsample the original point cloud in RaySense,
which usually generates a set of around $800$ unique points.
We then sample the second point cloud with a different set of rays.
For fair comparison, we also subsample $800$ points in the case of uniformly random subsampling.
We use the root mean square error (RMSE) as our metric, and we also record the convergence time,
where the convergence criteria is a threshold of the relative RMSE.
We summarize the performance results in Tab.~\ref{tab:registration},
and we provide some visualization to compare the three different settings in Fig.~\ref{fig:registration}.

\begin{table}[htbp]
  \caption{Sample point cloud registration result.
    Performances are evaluated by registration accuracy (measured by root mean squared error (RMSE))
    and computation times.
    The statistics reported are averaged over $5$ runs.
    ``RMSE'' is evaluated over the subsampled points
    while ``RMSE(full)'' is evaluated over the original point cloud.
    }
    \label{tab:registration}
    \centering
    \begin{tabular}{l c c c c}
     \hline
       & number of points & RMSE & RMSE(full) & Convergence time (s) \\
      \hline
      Vanilla ICP     & 100\,000    & 4.544e-06 & 4.544e-06 & 6.319  \\
      ICP + random & 800    & 4.509e-02 & 3.053e-03 & 0.0192\\
      ICP + RaySense & 804.6 & 2.601e-02 & 1.077e-03 & 0.0116\\
      \hline
    \end{tabular}
\end{table}

From Tab.~\ref{tab:registration}, it is clear that both the sampling schemes accelerate the registration process drastically by considering only a portion of the original dense point cloud.
It also suggests that RaySense sample has slight advantage over the uniform random sample, in both accuracy and convergence time.
However, generating the RaySense samples on the fly needs around $0.65$s on average, while generating a random subsample requires only $0.01$s.
% \Colin{Do these times include the sampling time?}
% \Lewis{sadly no, the sampling time goes to ~$0.7s$ that I did not include}

Fig.~\ref{fig:registration_salient} again shows that RaySense is sampling salient features.
Thus a possible improvement is to use the repetition information from RaySense sampling
to perform a weighted registration, for example with the (autodetected) salient features
receiving higher weights.
This is left for future investigation.

\subsection{Point cloud classification using neural networks}\label{sec:NN}
% \Richard{Would it be possible to train RayNN for classifying MNIST images?}
% \Lewis{not really... RayNN is for 3D point cloud, each input is a point cloud, if do that for MNIST then we only have 10 point clouds...}

We use the RaySense sketch to classify objects from the ModelNet dataset~\cite{wu20153d},
using a neural network, which we call RayNN.
RayNN can use features from different sampling operators  $\mathcal S$ introduced in \S~\ref{sec:signature} as inputs.
When using multiple nearest neighbors: $\mathcal S\big[\Gamma, [1,2,\dots,\eta]\big]$, we denote our models by RayNN-cp$\eta$.
%\del{We use the RaySense sampling operator with different number of neighbors, denoted by RayNN-X, where X is related to the input features.}
For our implementation, while we might use different numbers of nearest neighbors, we always include the closest point coordinates and the vector to closest points in our feature space $\mathbb R^c$ ($c \ge 6$ fixed). Details of the implementation can be found in \Cref{sec:NN_detail}.

%\begin{figure}[tbp]
%\centering
%\includegraphics[width=0.9\linewidth]{figures/hist_compare.pdf}
%\caption{Comparison of histogram of the $x,y,z$ coordinates of points sampled by RaySense, using $\ell^2$ and the Wasserstein distance $W_1$. Rows and columns correspond to object labels. Red $\times$ indicate location of the argmin along each row.} \label{fig:hist_compare}
%\end{figure}

\begin{table}[htbp]
  \caption{ModelNet classification results.
  Here we report our best accuracy results over all experiments. For reference, the test scores for RayNN-cp5 $(m=32)$ has mean around $90.31\%$ and standard deviation around $0.25\%$ over $600$ tests. 
  %\del{Results reported here are the maximal accuracy we observed over all our experiments.}
    %\del{All tests use $N = 1024$ sample points.}
    % this is already mentioned in the Data section
  }
    \label{tab:classificationresults}
    \centering
    \begin{tabular}{l c c}
     \hline
       & ModelNet10 & ModelNet40\\
      \hline
      PointNet \cite{qi2017pointnet}     & --     & 89.2\\
      PointNet++ \cite{qi2017pointnet++} & --     & 90.7\\
      ECC \cite{simonovsky2017dynamic}   & 90.8   & 87.4\\
      kd-net \cite{klokov2017escape}     & 93.3   & 90.6\\
      PointCNN \cite{li2018pointcnn}     & --     & 92.5\\
      PCNN \cite{atzmon2018point}        & 94.9   & 92.3\\
      DGCNN \cite{wang2019dynamic}       & --     & \textbf{92.9}\\
      RayNN-cp1 ($m=16$)                 & 94.05  & 90.84 \\
      RayNN-cp5 ($m=32$)                 & \textbf{95.04} & 90.96 \\
      \hline
    \end{tabular}
\end{table}

%\subsubsection{RayNN Results}
We compare with some well-known methods for 3D point cloud classification tasks.
In addition to the results reported by \cite{qi2017pointnet},
we also compare against PointNet.pytorch, a
PyTorch re-implementation \cite{site:PointNetpytorch} of PointNet.
In all our experiments, we report overall accuracy.
Table~\ref{tab:classificationresults} shows RayNN is competitive. To investigate the robustness of our network, we perform several more experiments.

\begin{figure}[htbp]
  \centering
  % ran pdfcrop
  \includegraphics[width=0.6\linewidth]{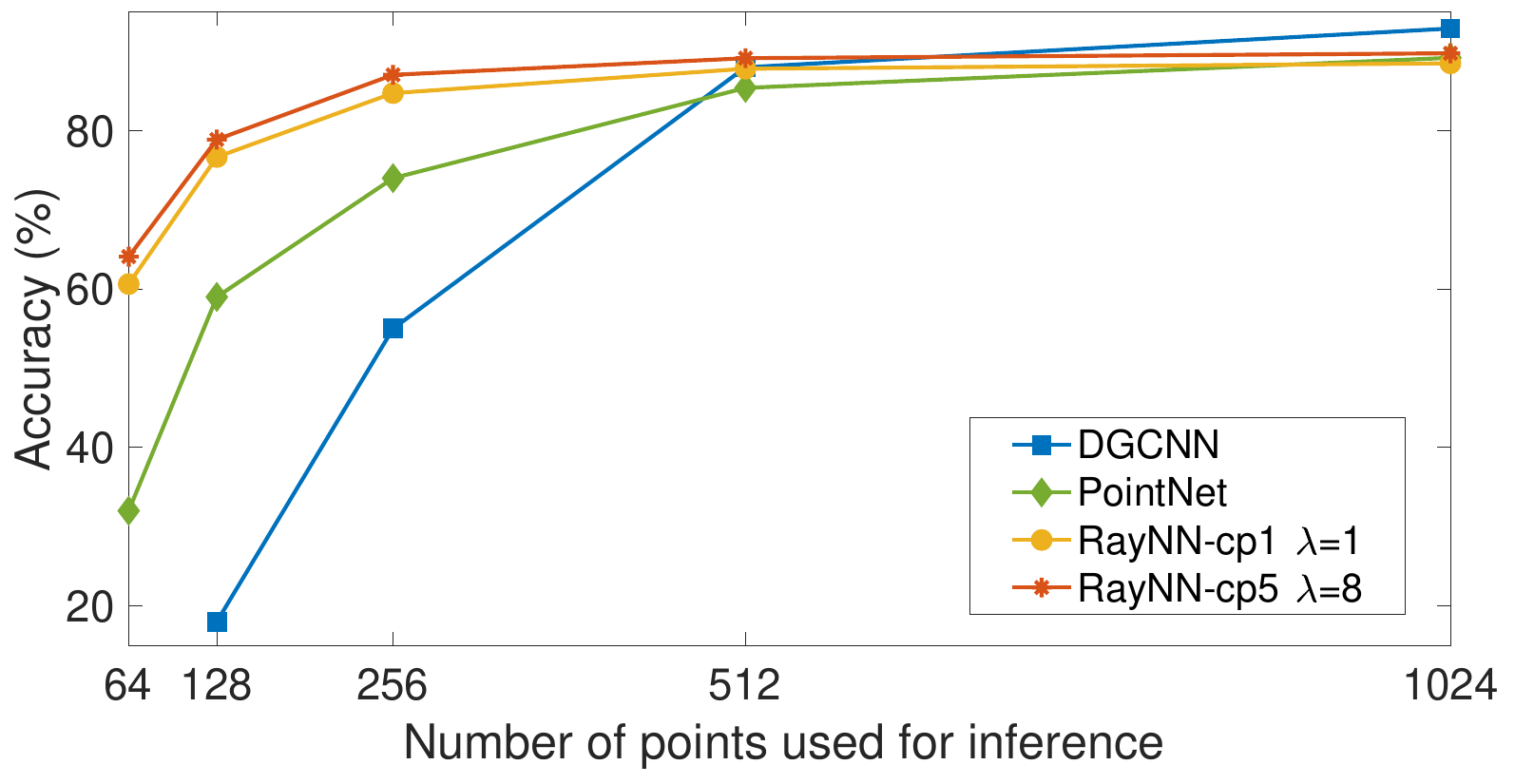}
  \caption{
  Testing DGCNN~\cite{wang2019dynamic}, PointNet~\cite{qi2017pointnet} and RayNN on ModelNet40 with missing data.
  %Comparison between DGCNN~\cite{wang2019dynamic}, PointNet~\cite{qi2017pointnet} and RayNN with random input dropping during testing on ModelNet40. 
  }
  \label{fig:missingdatacompare}
\end{figure}

\mypara{Robustness to sample size}
%\blue{Here, we present two studies reflecting different ways to vary sample size. }
We repeat the experiments in~\cite{qi2017pointnet, wang2019dynamic} whereby, after training,
data is randomly removed prior to testing on the remaining points.
The results in Fig.~\ref{fig:missingdatacompare} show that RayNN performs very well with significant missing data.

\mypara{Using fewer rays}
We experiment with training using a full set of $m=32$ rays but test using smaller number $\hat{m}$ of rays.
Table~\ref{tab:testwithfewer} shows that RayNN can achieve a reasonable score even if only $\hat{m}=4$ rays are used for inference. 
%\blue{In addition, Table~\ref{tab:sensitivity} shows that if we train with the same number of rays as used during inference, we can achieve better results, even with a few rays.}

\mypara{Robustness to outliers}
This experiment simulates situations where noise severely perturbs the original data during testing.
We compare the performance of RayNN-cp1, RayNN-cp5 and PointNet.pytorch in Table~\ref{tab:outliers}.
The comparison reveals RaySense's capability in handling unexpected outliers, especially when additional nearest neighbors are used.
Note the experiment here is different from that in \cite{qi2017pointnet} where the outliers are fixed and included in the training set.

\begin{table}[htpb]
  \caption{Accuracy when testing with a reduced ray set.
    RayNN-cp1 was trained using $m=32$ rays. Results averaged over 5 runs.}
  \label{tab:testwithfewer}
  \centering
    \begin{tabular}{ccccc} 
 %   \hline
 %   \multicolumn{2}{c}{RayNN-cp5 Ensemble 8 times}\\
 %   \hline
 %    Number of rays in test & Score\\
 %   \hline
 %    $m = 32$ & 93.85\%\\
 %    \hline
 %    $m = 16$ &93.37\%\\
 %    \hline
 %    $m = 8$ & 89.89\% \\
 %    \hline
 %    $m = 4$ & 79.14\% \\
 %    \hline
    % \hline
    %\multicolumn{5}{c}{RayNN-cp$1$}\\
    \hline
   \multicolumn{5}{c}{ModelNet40}\\
    \hline
              $\hat{m}$  & $32$ & $16$ & $8$ & $4$\\
    \hline
    $\lambda=1$ & 88.50\% &86.13\%& 74.64\% & 43.28\%\\
    $\lambda=8$ &  89.77\% & 88.94\% & 82.97\% & 55.24\%\\
     \hline
    \end{tabular}
\end{table}

\mypara{Comparison of model complexity}

Table~\ref{tab:complexity} shows that our network has an advantage in model size and feed-forward time even against the simple and efficient PointNet.
%\Colin{This next sentence no longer relevant right?!  \sout{``The longer training time per epoch is due to data preprocessing:''}}
In both training and testing,
there is some overhead in data preprocessing to build a kd-tree, generate rays,
and perform the nearest-neighbor queries to form the RaySense sketch.
For point clouds of around $N=1024$,
these costs are not too onerous in practice as shown in table~\ref{tab:complexity}.

%\del{The small model size in Table~\ref{tab:complexity} is because RayNN has fewer parameters.}
%\Richard{I delete this because it is kind of obvious.}
The convolution layers have $48c + \num{840016}$ parameters, where $c$ is the dimension of input feature space.
The fully-connected layers have $64K + \num{278528}$ parameters, where $K$ is the number of output classes.
In total, our network has $1.1\times10^6 + 48c + 64K \approx 1.1$M parameters.
In comparison, PointNet~\cite{qi2017pointnet} contains $3.5$M parameters.

\begin{table}[htbp]
  \caption{Outliers sampled uniformly from the unit sphere are introduced during testing.
  The networks are trained without any outliers. Results averaged over 5 runs.
  }
  % \del{RayNNs are trained with $m = 32$.}
  % \del{8 times ensemble in test is also used RayNN.}}
    % \todo{PointNet also averaged over 5 tests + ensemble?}\Lewis{Only average over 5 tests. Didn't implement ensemble in pointnet. Maybe we also report the non-ensemble results?}}
    \label{tab:outliers}
    \centering
    \begin{tabular}{lccc}
    \hline
     ModelNet10 & no outliers & 5 outliers & 10 outliers\\
    \hline
     RayNN-cp1 & 93.26\% & 79.76\% & 53.94\%\\
     RayNN-cp5 &  \textbf{93.85\%} & \textbf{92.66\%} & \textbf{90.90\%}\\
     PointNet.pytorch\!\! & 91.08\% & 48.57\% & 25.55\%\\
     \hline
     {ModelNet40} & & & \\
     \hline
     RayNN-cp1 & 89.77 \% & 54.66\% & 20.95\%\\
     RayNN-cp5 &  \textbf{90.38\%} & \textbf{88.49\%} & \textbf{78.06\%}\\
     PointNet.pytorch\!\! & 87.15\% & 34.05\% & 17.48\%\\
     \hline
    \end{tabular}
\end{table}

\begin{table}[htbp]
       \caption{Top: storage and timings for RayNN-cp1 and PointNet.pytorch on ModelNet40 using one Nvidia 1080-Ti GPU and batch size 32.
       The preprocessing and forward time are both measured per batch. 
       %Timing results averaged over $10$ epochs.
       %\Lewis{Yes, just checked CP5 its about 2-3 (s) slower in time/epoch}
       Bottom: data from \cite{wang2019dynamic} is included only for reference; no proper basis for direct comparison.
       % \blue{cannot directly compare between devices.}
       }
       \label{tab:complexity}
      \centering
         \begin{tabular}{lrrrr}
      \hline
        %&  Model    & Forward   \\
        %&  size\;\; & time\;\;  \\
        &  Model size & Forward time & Preprocessing time & Time per epoch \\
      \hline
      PointNet.pytorch\!\!  & \SI{14}{MB}  & \SI{12}{ms} &\textbf{3.6}\,\si{ms} & \textbf{14}\,\si{s}\\ 
      RayNN-cp1 & \textbf{4.5}\,\si{MB} & \textbf{2}\,\si{ms} &\SI{7.5}{ms} &\SI{22}{s}\\
      \hline
      PointNet \cite{qi2017pointnet} & \SI{40}{MB} & \SI{16.6}{ms} & - & - \\
      PCNN \cite{atzmon2018point}    & \SI{94}{MB} & \SI{117}{ms} & - & -  \\
      DGCNN \cite{wang2019dynamic}   & \SI{21}{MB} & \SI{27.2}{ms} & - & -  \\
      \hline
      \end{tabular}
     %\smallskip
     %\parbox[t]{0.75\linewidth}{\footnotesize
     % $^*$\,data from \cite{atzmon2018point} using Nvidia P100 GPUs.\\
     %}
  \end{table}

\section{Summary}\label{sec:summary}

RaySense is a sampling technique based on projecting random rays onto a data set. 
%\Lewis{maybe here we also want to say recording since Richard has changed the abstract}
This projection involves finding the nearest neighbors in the data for points on the rays.
These nearest neighbors collectively form the basic ``RaySense sketch'',
which can be employed for various data processing tasks. 

RaySense does not merely produce narrowly interpreted subsamples of given datasets. Instead, it prioritizes the sampling of salient features of the dataset, such as corners or edges, with a higher probability. Consequently, points near these salient features may be recorded in the sketch tensor multiple times. 

From the RaySense sketch, one can further extract snapshots of integral or local (differential) information about the data set.
Relevant operations are defined on the rays randomly sampled from a chosen distribution. Since rays are one-dimensional objects, the formal complexity of RaySense does not increase exponentially with the dimensions of the embedding space. 
We provide theoretical analysis showing that the statistics of a sampled point cloud depends solely on the distribution of the rays, and not on any particular ray set.
Additionally, we also demonstrated that by appropriately post-processing the RaySense sketch tensor obtained from a given point cloud, one can compute approximations of line integrals.
Thus, and by way of the Fourier Slice Theorem, we argue that RaySense provides spectral information about the sampled point cloud.

We showed that RaySense sketches could be used to register and classify point clouds of different cardinality.
For classification of point clouds in three dimensions, we presented a neural network classifier called ``RayNN'', which takes the RaySense sketches as input.
Nearest-neighbor information can be sensitive to outliers.
For finite point sets, we advocated augmentation of the sketch tensor by including
multiple nearest neighbors to enhance RaySense's capability to capture persistent features in the data set, thereby improving robustness.
We compared the performance of RayNN to several other prominent models, highlighting its lightweight, flexible, and efficient nature. Importantly, RayNN also differs from conventional models, as it allows for multiple tests with different ray sets on the same dataset.

% To the best of our knowledge RaySense is a new idea, so there are
% many avenues of possible study.

% On the theoretical side, one could study RaySense's invariant
% properties for more general geometric objects beyond point clouds.
% and
% its connections to topological structure.
% There are many practical applications to explore such as
% training with outliers or the simultaneous registration, classification and segmentation of point
% clouds.
% Finally, we expect to find applications to high-dimensional data sets.
% For example, RaySense could be used as intermediate step for the semi-guided identification of appropriate feature spaces.

%We look forward to exploring RaySense and RayNN further.

\medskip

\mypara{Acknowledgments}
Part of this research was performed while Macdonald and Tsai were visiting the Institute for Pure and Applied Mathematics (IPAM), which is supported by the National Science Foundation (Grant No. DMS-1440415).
This work was partially supported by a grant from the Simons Foundation, NSF Grants DMS-1720171 and DMS-2110895, and a Discovery Grant from Natural Sciences and Engineering Research Council of Canada.
The authors thank Texas Advanced Computing Center (TACC) and UBC Math Dept Cloud Computing for providing computing resources.

%Author **** is supported by grant ****.

\smallskip

\mypara{Statements and Declarations}
The authors have no competing interests to declare that are relevant to the content of this article.
% The authors have no relevant financial or non-financial interests to disclose.\\
%All authors certify that they have no affiliations with or involvement in any organization or entity with any financial interest or non-financial interest in the subject matter or materials discussed in this manuscript.\\
%The authors have no financial or proprietary interests in any material discussed in this article.

%\clearpage
%\bibliographystyle{ieee_fullname}
\bibliography{references.bib}

% not newpage: we want to force figures to appear before here.
\clearpage
\appendix
\section{Examples of ray distributions}\label{sec:method}
We assume all points are properly calibrated by a common preprocessing step.
This could also be learned.
In fact, one can use RaySense to train such a preprocessor to register the dataset,
for example, using \S~\ref{sec:registration} or similar.
However, for simplicity, in our experiments, we generally normalize
each point set to be in the unit $\ell^2$ ball, with center of mass at
the origin, unless otherwise indicated.

%\subsection{Generating random rays}
%How should the rays be generated? What does \emph{random} mean? 

We present two ways to generate random rays.
There is no \emph{right} way to generate rays, although it
is conceivable that one may find optimal ray distributions for specific applications.

\mypara{Method R1}
One simple approach is generating rays of fixed-length $L$, whose direction $\vec{v}$ is uniformly sampled from the unit sphere.
We add a shift $\vec{b}$ sampled uniformly from $[-\frac{1}{2}, \frac{1}{2}]^d$ to avoid a bias for the origin.
The $\ppr$ sample points are distributed evenly along the ray:
\begin{align*}
\vec r_i = \vec{b} + L \left( \frac{i}{\ppr-1}  - \frac{1}{2} \right) \vec{v} , \qquad i = 0, \dots, \ppr-1
\end{align*}
The spacing between adjacent points on each ray is denoted by $\delta r$, which is $L/(\ppr-1)$. We use $L=2$.

\mypara{Method R2}
Another natural way to generate random rays is by random endpoints selection:
choose two random points $\vec{p}$, $\vec{q}$ on a sphere and connect them to form a ray.
Then we evenly sample $\ppr$ points between $\vec{p}$, $\vec{q}$ on the ray.
To avoid overly short rays where information would be redundant, we use a minimum ray-length threshold $\tau$ to discard rays.
Note that the distance between $\ppr$ sample points are different on different rays:
\begin{equation*}
  \vec r_i = \vec{p} + \frac{i}{\ppr-1} (\vec{q}-\vec{p}), \qquad i=0,\dots,\ppr-1.
\end{equation*}
The spacing of points on each ray varies, depending on the length of the ray.

\begin{figure}[htbp]
\centering
\includegraphics[width=0.5\linewidth]{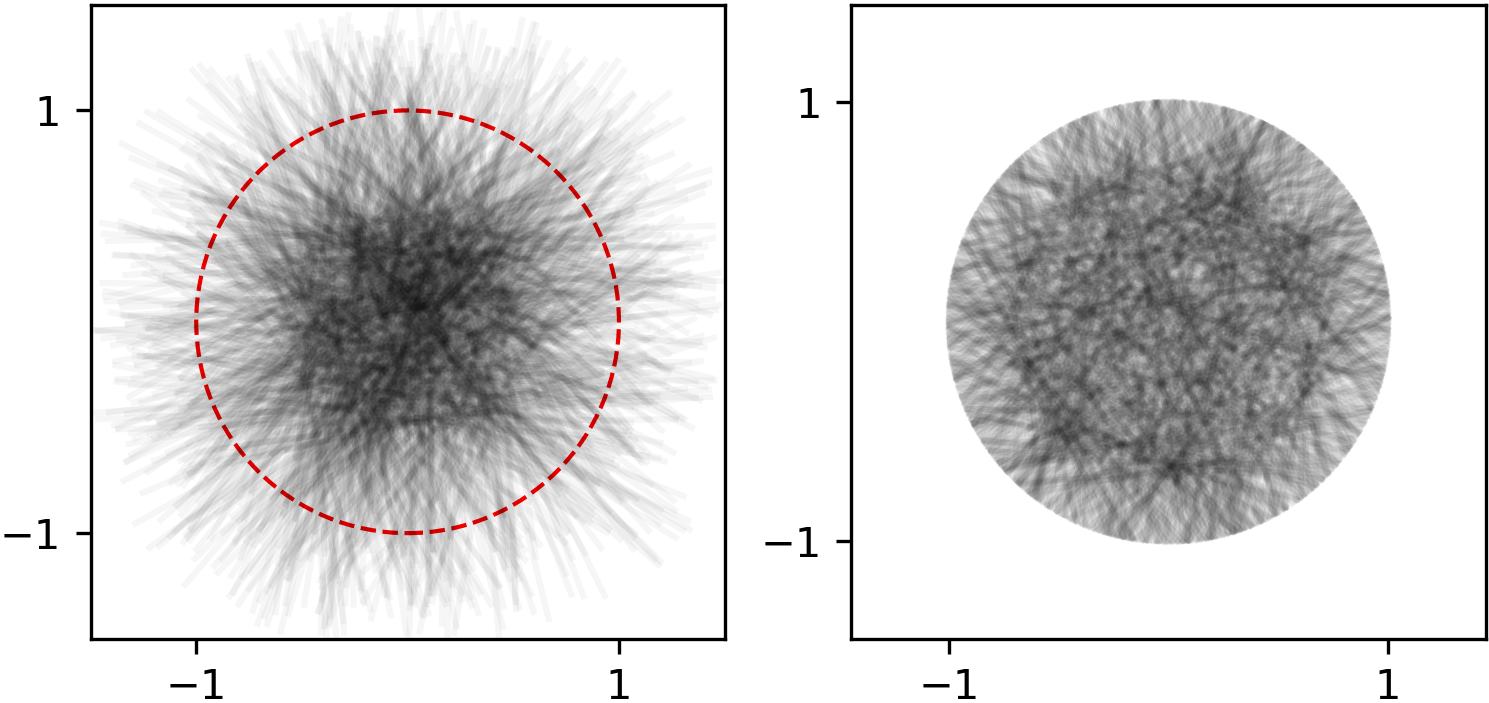}
\caption{\label{fig:rays} Density of rays from method R1 (left) and R2 (right). Red circle indicates the $\ell^2$ ball.}
\end{figure}

Fig.~\ref{fig:rays} shows the density of rays from the ray generation methods.
In this paper, we use Method R1;
%\del{although the point sets lie in the $l_2$ ball, having rays outside the sphere improves the sampling of salient points.}
a fixed $\delta r$ seems to help maintain spatial consistency along the rays, which increases RayNN's classification accuracy in~\S~\ref{sec:NN}.

% \Lewis{What should I do if I'm using Method R1... Long time ago I used a method similar to method $1$ in Bertrand Paradox Wikipedia, there we couldn't guarantee the equi-distance, and the performance was a bit off than method R1. Should I put this on?}

% \Lewis{ part of the reason is likely because R2 stays within the sphere. Maybe it helps to also have sample points outside the sphere. Also the equidistance issue. May be worth mentioning so we can justify using R1}

\section{Implementation details of RayNN}\label{sec:NN_detail}
Our implementation uses PyTorch~\cite{paszke2017automatic}.
%is available.\footnote{\url{https://github.com/****/****} [to be added]}
% \Colin{I commented out the link: put back if ready and not just stars.}

\mypara{Architecture}
RayNN takes the $m\times k \times c$ RaySense sketch tensor $\mathcal S(\Gamma)$ as input, and outputs a $K$-vector of probabilities, where $K$ is the number of object classes.

%\del{The signature $S(\Gamma)$ is of size $m\times k \times c$}
%\del{where $R_{m,k}$ is the ray set we use and $c$ is the dimension of feature space (\S~\ref{sec:signature}).}

\begin{figure*}[htbp]
  % ran pdfcrop
  \centering
  \includegraphics[width=0.99\linewidth]{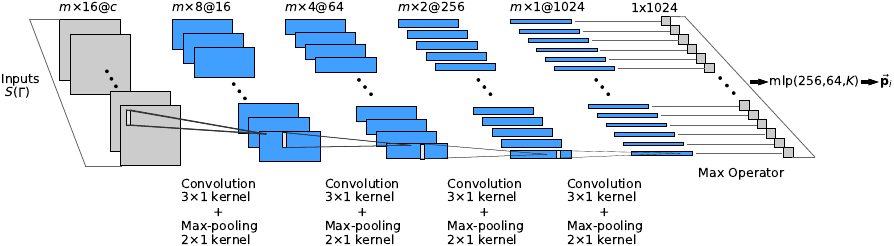}
  \caption{The RayNN architecture for $m$ rays and $\ppr$ samples per ray.
      The input is $c$ feature matrices from $\mathcal S(\Gamma)$ with suitable operations.
    %\Colin{prev ok?}  % basically I don't see how to go from $c$ matrices to 16 matrices
    With $\ppr = 16$, each matrix is downsized to an $m$-vector
   by 4 layers of 1-D convolution and max-pooling.
    % we choose $\ppr = 16$ specifically to make sure each feature matrix is downsized to a vector of size $1 \times m$ after $4$
    % convolution layers
    The max operator is then applied to each of the 1024 $m$-vectors.
    The length-$1024$ feature vector is fed into a multi-layer perceptron (mlp)
    which outputs a vector of probabilities, one for each of the $K$ classes in the classification task.
    % A dropout layer is implemented on the penultimate fully-connected layer and Batchnorm along with ReLU are used for every fully-connected and convolution layer.
    Note the number of intermediate layers (blue) can be increased based on $\ppr$ and $c$.
  }
  \label{fig:architecture}
\end{figure*}

The first few layers of the network are blocks of 1D convolution followed by max-pooling to encode the sketch into a single vector per ray.
Convolution and max-pooling are applied along the ray.
%; it does not make sense to compute convolutions \emph{between} rays.
After this downsizing, we implement a max operation across rays.
Fig.~\ref{fig:architecture} includes some details.
The output of the max pooling layer is fed into fully-connected layers with output sizes $256$, $64$, and $K$ to produce the desired vector of probabilities $\vec{\mathbf{p}}_i \in \Real^K$.
Batchnorm~\cite{ioffe2015batch} along with ReLU~\cite{nair2010rectified} are used for every fully-connected and convolution layer.

Note that our network uses convolution along rays to capture local information while the fully-connected layers aggregate global information.
Between the two, the max operation across rays ensures invariance to the ordering of the rays.
It also allows for an arbitrary number of rays to be used during inference.
These invariance properties are similar to PointNet's input-order invariance~\cite{qi2017pointnet}.

%   \begin{table*}[ht!]
%     \caption{Comparison between RayNN and the-state-of-the-art PointNet.  \TODO{ should we report number of epochs?  And time per epoch?/total time?} \Lewis{I think there is no need to include 2048/4096 case for ModelNet40 now. No significant difference plus pointnet paper doesn't have corresponding results}}
%       \centering
%      \begin{tabular}{ p{2.5cm}|p{3cm}p{3cm}p{3cm}p{3cm} }
%     %  \hline
%     %  \multicolumn{5}{c}{Network Performance Comparison} \\
%      \hline
%      Network  & ModelNet10 (2048) & ModelNet40 (2048) & ModelNet40 (1024)& ModelNet40 (4096)\\
%      \hline
%      RayNN   & 95.04\% ($mk$=256)   &90.03\% ($mk$ = 256?) &  90.60\% ($mk$ = 256?) & 90.44\% ($mk$ = 256?)\\
%      PointNet (paper) & - & - & 89.2\% & - \\
%      PointNet.pytorch &92.07\% (on 256 pts) & 85.3\% (on 1024 pts) & 87.52\% (all 1024) & 85.4\% (all 4096?) \\
%      \hline
%     \end{tabular}
%       \label{tab:Performance Compare}
%   \end{table*}
%   When the signature comes from different ray sets, 
%\todo{we add noise during training...    when?  must be before kd-tree?  We should describe precisely this error...} \Lewis{Yes here is an issue: if we add noises before kd-tree (ball-tree I used), then the noises are fixed for each object if we build the kd-tree beforehand. }

\mypara{Data}
We apply RayNN on the standard ModelNet10 and ModelNet40 benchmarks \cite{wu20153d} for 3D object classification.
ModelNet40 consists of \num{12311} orientation-aligned \cite{sedaghat2016orientation} meshed 3D CAD models,
divided into \num{9843} training and \num{2468} test objects.
ModelNet10 contains \num{3991} training and \num{908} test objects.
Following the experiment setup in \cite{qi2017pointnet},
we sample $N=1024$ points from each of these models and rescale them to be bounded by the unit sphere to form point sets.\footnote{%
  RaySense does not require point clouds for inputs:
  we could apply RaySense directly to surface meshes, implicit surfaces,
  or even---given an fast nearest neighbor calculator---the CAD models directly.}
%\Lewis{do we want this paragraph?} On ModelNet40, using $m=16$ rays, our best result is $90.84\%$.  If we instead sample the original objects using $N=2048$ or $N=4096$ points, the result is essentially unchanged (90.03\% and 90.44\% respectively).
Our results do not appear to be sensitive to $N$.
%\del{to $N$; we observe similar results with $N=2048$.} 

\mypara{Training}
During training, we use dropout with ratio $0.5$ on the penultimate fully-connected layer.
We also augment our training dataset on-the-fly by
% using random rotations about the vertical axis and 
adding $\mathcal{N}(0,0.0004)$ noise to the coordinates.
For the optimizer, we use Adam \cite{kingma2014adam} with momentum \num{0.9} and batch size \num{16}.
The learning rate starts at \num{0.002} and is halved every \num{100} epochs.

\mypara{Inference}
Our algorithm uses random rays, so it is natural to consider strategies to reduce the variance in the prediction.
We consider one simple approach during inference by making an ensemble of predictions from $\lambda$ different ray sets.
% \del{During inference, we increase the performance and stability of our network by ensemble predictions from multiple ray sets.}
% \del{For the same object, each random ray set yields a unique signature tensor, and for different signature tensors, we have different vectors of probabilities, denoted by $\Vec{v}_i$. }
The ensemble prediction is 
% \del{done by taking average}
based on the average
over the $\lambda$ different probability vectors $\Vec{\mathbf{p}}_i \in \Real^K$, i.e.,
\begin{equation*}
  \text{Prediction}(\lambda) = \frac{1}{\lambda} \sum_{i=1}^\lambda \Vec{\mathbf{p}}_i.
\end{equation*}
The assigned label then corresponds to the entry with the largest probability.
We denote the number of rays used during training by $m$, while the number of rays used for inference is $\hat{m}$.
Unless otherwise specified, we use $\lambda=8$, $m=32$ rays, and $\hat{m}=m$.

% This is where we should mention the difference between pointnet and pointnet.pytorch.

%\del{Alternatively, one can also ensemble by averaging the feature vector before the fully-connected layers, but its performance is slightly worse than ensemble over $\Vec{v}_i$.} 

% We compute final accuracy by taking weighted average of per-class accuracy over all the classes in test dataset.\Lewis{to do}

\section{Details of the proof of Theorem~\ref{thm:consistency}}
\label{sec:proof_details_voronoi}

This appendix contains the proofs of Lemmas~\ref{thm:nonuniformsampling}, \ref{thm:nonuniform_voronoi}, and~\ref{thm:uniform_diameter}, and Theorem~\ref{thm:consistency}.

\begin{proof}[Proof of Lemma~\ref{thm:nonuniformsampling}.]
The probability measure of $\Omega_j$ is
\begin{equation*}
  P_{\Omega_j}  = \int_{\vec x\in \Omega_j} \rho(\vec x) \mathrm{d}\vec x > 0,
\end{equation*}
which represents the probability of sampling $\Omega_j$ when drawing $i.i.d.$ random samples from $\mu$. For a fixed set of such hypercubes, any $\vec x\in \supp(\rho)$ will fall in one of the $\Omega_j$'s. Then one can define a mapping $h:\supp(\rho)\subset\mathbb{R}^d\to\mathbb{R}$ by: \[s = h(\vec x) =
j-1, \quad \text{where } \vec x\in \Omega_j, \; j = \,1,\,2,\,\dots,\,M.\]
By applying the mapping to the random vector $\vec X$, we obtain a new discrete random variable $S$ with the discrete probability distribution $\mu_M$ on $\mathbb{R}$ and the corresponding density $\rho_M$.  The random variable $S$ lives in a discrete space: $S \in \{0,\,1,\,\ldots,\,M-1\}$ and $\rho_M$ is given as a sum of delta spikes as  \[ \rho_M(s) = \sum_{j=1}^M P_{\Omega_j}\delta_j(s).\]
As a result, sampling from the distribution $\mu_M$ is equivalent to sampling the hypercubes according to the distribution $\mu$ in $\mathbb{R}^d$, but one cares only about the sample being in a specific hypercube $\Omega_j$, not the precise location of the sample. Let $F_M(s)$ denote the cumulative density function related to the density function $\rho_M(s)$. 

Now, given a set of $N$ independent samples of $\vec X:$ $\{\vec X_i\}_{i=1}^N\subset\mathbb{R}^d$, we have a corresponding set of $N$ independent sample points of $S:$ $\{s_i\}_{i=1}^N$ such that $\vec X_i\in \Omega_{s_i+1}$. From there, we can regard the histogram of $\{s_i\}_{i=1}^N$ as an empirical density of the true density $\rho_M$. Denote the empirical density by $\tilde{\rho}_M^N$ which is given by
\[ \tilde{\rho}_M^N = \frac{1}{N}\sum_{i=1}^{N} \delta_{s_i}.\]
 one can therefore also obtain an empirical cumulative density function $\tilde{F}^N_M(s)$ using the indicator function $\chi$:
\[ \tilde{F}^N_M(s) = \frac{1}{N} \sum_{i=1}^{N} \chi_{\{s_i\leq s\}}.\] By Dvoretzky--Kiefer--Wolfowitz inequality~\cite{dvoretzky1956asymptotic, massart1990tight} we have
\[ Prob\big(\sup_{s\in\mathbb{R}}\abs{F_M(s)-\tilde{F}^N_M(s)}>\varepsilon\big)\leq 2e^{-2N\varepsilon^2}, \quad \text{ for all }\varepsilon>0.\]
Therefore, for a desired fixed probability $p_0$, the above indicates the approximating error given by the empirical $\tilde{F}^N_M(s)$ is at most 
% $p_0 = \max(0, 1-2e^{-2N\varepsilon^2})$ 
\[ \sup_{s\in\mathbb{R}}\abs{F_M(s)-\tilde{F}^N_M(s)} \leq\varepsilon_N = \Big(-\frac{1}{2N}\ln{\big(\frac{1-p_0}2\big)}\Big)^\frac{1}{2},\] with probability at least $p_0$. Then note that the true probability measure $P_{\Omega_j}$ of $\Omega_j$ being sampled by random drawings from $\mu$ is equivalent to the true probability of $j-1$ being drawn from $\mu_M$, $i.e.$ 
\[ P_{\Omega_j} = P_M(j-1) := \rho_M(j-1),\]
therefore $P_{\Omega_j} = P_M(j-1)$ can be computed from $F_M$ by:
\begin{align*}
    P_{\Omega_j} &= F_M(j-1)-F_M(j-2) \\ &= F_M(j-1)-\tilde{F}^N_M(j-1)+\tilde{F}^N_M(j-1)-\tilde{F}^N_M(j-2)+\tilde{F}^N_M(j-2)-F_M(j-2).
\end{align*} Taking absolute value and using the triangle inequality, with the fixed $p_0$
\[ P_{\Omega_j} \leq 2\varepsilon_N + \tilde{P}^N_M(j-1),
\]
% \abs{F_M(j+1)-\tilde{F}^N_1(j+1)} + \abs{\tilde{F}^N_1(j+1)-\tilde{F}^N_M(j)} + \abs{\tilde{F}^N_M(j)-F_M(j)}
where $\tilde{P}^N_M(j-1)$ denotes the empirical probability at $j-1$. Apply the same argument to $\tilde{P}^N_M(j-1)$ one has 
\[ \abs{P_{\Omega_j} -  \tilde{P}^N_M(j-1)} \leq 2\varepsilon_N, \quad \text{for all }j=1,\,2,\,\dots,\,M.\]
For a set of $N$ sample points, $ \tilde{P}^N_M(j-1)$ is computed by $\displaystyle \frac{N_j}{N}$, where $N_j$ is the number of times $j-1$ got sampled by $\{s_i\}_{i=1}^N$, or equivalently $\Omega_j$ got sampled by $\{\vec X_i\}_{i=1}^N$, which indicates that in practice, with probability at least $p_0$, the number of sampling points $N_j$ in $\Omega_j$ satisfies the following bound:
\[ P_{\Omega_j} - 2\varepsilon_N\leq\frac{1}{N}N_j \leq P_{\Omega_j} + 2\varepsilon_N \implies P_{\Omega_j}N - 2\varepsilon_N N\leq N_j \leq P_{\Omega_j}N + 2\varepsilon_N N.\]
By taking $N$ large enough such that $P_{\Omega_j}N -2\varepsilon_N N = 1  \implies N_j \geq 1$:
\[\implies N = \frac{\sqrt{\ln{(\frac{1-p_0}2)}\big(\ln{(\frac{1-p_0}2)}-2P_{\Omega_j}\big)}+P_{\Omega_j} - \ln{(\frac{1-p_0}2)}}{P_{\Omega_j}^2}.\]
The above quantity is clearly a function with respect to the probability measure $P_{\Omega_j}$, and any $\Omega_i$ with $P_{\Omega_i}\geq P_{\Omega_j}$ would have $N_i\geq N_j\geq 1$.
Using $\nu$ to denote such a function and $0<P\leq 1$ as the threshold measure completes the first part of the proof:
\[\implies  \nu\big(P\big) = \frac{\sqrt{\ln{(\frac{2}{1-p_0})}\big(\ln{(\frac{2}{1-p_0})}+2P\big)}+P + \ln{(\frac{2}{1-p_0})}}{P^2}.\]
To establish the bounds on the expression,
%note that $0<P\leq 1$, we also have
we note
\begin{align*}
  \nu(P) &> \frac{\sqrt{\ln{(\frac{2}{1-p_0})} \ln{(\frac{2}{1-p_0})}} + P + \ln{(\frac{2}{1-p_0})}}{P^2}=\frac{2\ln{(\frac{2}{1-p_0})}+P}{P^2}, \\
  \nu(P) &< \frac{\sqrt{\big(\ln{(\frac{2}{1-p_0})}+2P\big)^2} + P + \ln{(\frac{2}{1-p_0})}}{P^2}=\frac{2\ln{(\frac{2}{1-p_0})}+3P}{P^2}.
\end{align*}
\end{proof}

%% \Colin{%
%%   \todo{probably I will comment this calculation out as Lewis has never verified my numbers (?) but mainly I'm just using it for my own understanding.}
%%   For example, consider the uniform density on the unit disc in
%% $\mathbb{R}^2$,
%% and suppose $\Omega_j$ is any $\frac{1}{2} \times \frac{1}{2}$
%% square inside the disc, then
%% $P_{\Omega_j} = \frac{1}{4\pi}$.
%% At $p_0 = 95\%$, Lemma~\ref{thm:nonuniformsampling} we have $\lceil \nu(P_{\Omega_j})\rceil = 1191$: this is an upper bound for number of samples we need to be confident at $p_0 = 95\%$ that every $\Omega_j$ contains at least one sample.
%% %Rising to 18742 for a $\frac{1}{4} \times \frac{1}{4}$ square.
%% The bound may not be very tight but it is sufficient for our purposes.
%% }

% \Colin{I moved some explanatory text that was here in between the lemmas and merged stuff about bounds of $\mu$ into Lemma~\ref{thm:nonuniformsampling}.}

\begin{proof}[Proof of Lemma~\ref{thm:nonuniform_voronoi}.]
  Consider a local hypercube centered at $\vec y$, $\Omega_{\vec y} := \{\vec x+\vec y\in\mathbb{R}^d: \|\vec x\|_\infty=l/2\}$ of length $l$ to be determined.
  We shall just say ``cube.''
  The probability of cube $\Omega_{\vec y}$ being sampled is given by $P_{\Omega_{\vec y}} = \int_{\Omega_{\vec y}} \rho(\vec x)\mathrm{d}\vec x$.
  Now for the set of standard basis vector $\{\vec e_i\}_{i=1}^d$, let $\vec v_d$ denotes the sum of all the basis: $\vec v_d :=\sum_{i=1}^d \vec e_i$.
  Without loss of generality, the probability of a diagonal cube, defined by $\Omega_{\vec y_d}:=\{\vec x+\vec y+\vec v_d \in\mathbb{R}^d: \|\vec x\|_\infty=l/2\}$,
  being sampled (unconditional to $\Omega_{\vec y}$ being sampled) has the following bound by Lipschitz continuity of $\rho$:
  \[
  \abs{P_{\Omega_{\vec y_d}} - P_{\Omega_{\vec y}}}\leq \int_{\Omega_{\vec y}} \abs{\rho(\vec x+l \vec v_d) - \rho(\vec x)}\mathrm{d}\vec x \leq L\sqrt{d}\,l\abs{\Omega_{\vec y}} \implies P_{\Omega_{\vec y_d}} \geq P_{\Omega_{\vec y}} - L\sqrt{d}\,l^{d+1}.
  \]
Furthermore, $P_{\Omega_{\vec y}}$ has the following lower bound also by Lipschitz continuity of $\rho$. For any $x \in \Omega_{\vec y}$, we have:
\begin{equation}
  \abs{\rho(\vec x) - \rho(\vec y)}
  \leq L \sqrt{d} \frac{l}{2}
  \implies
  \rho(\vec x)\geq \rho(\vec y)-L\frac{\sqrt{d}}{2}l
  \implies
  P_{\Omega_{\vec y}}\geq \big(\rho(\vec y)-L\frac{\sqrt{d}}{2}l\big)l^d.
\end{equation}
Combining with the previous bound for $P_{\Omega_{\vec y_d}}$, we further have:
\begin{equation*}
  P_{\Omega_{\vec y_d}}
  \geq \big(\rho(\vec y)-L\frac{\sqrt{d}}{2}l\big)l^d - L\sqrt{d}l^{d+1}
  = \rho(\vec y)l^d-\frac{3\sqrt{d}}2Ll^{d+1}.
\end{equation*}
By setting $ \rho(\vec y) > \frac{3\sqrt{d}}2Ll$ we can ensure $ P_{\Omega_{\vec y_d}} >0$,
but this extreme lower bound is based on on Lipschitz continuity.
To obtain a more useful bound, we will show below that by picking $l := l_N$ judiciously,
$\rho(\vec y) > 3\sqrt{d}Ll_N>0$, any surrounding cube has non-zero probability to be sampled.
Therefore, with $\rho(\vec y) > 3\sqrt{d}Ll_N$, for any diagonal cube $\Omega_{y_d}$:
\[ \rho(\vec y)l^d - \frac{1}{2}\rho(\vec y)l^d > \frac{3\sqrt{d}}2Ll^{d+1} \implies P_{\Omega_{\vec y_d}} > \frac{1}{2}\rho(\vec y)l_N^d.\]
Since the diagonal cube is the furthest to $\vec y$ among all the surrounding cubes, we have for every surrounding cube of $\Omega_{\vec y}$, their probability measure is at least $P_{\Omega_{\vec y_d}}$

% and one can always find some $\delta>0$ such that \[ P_{\Omega_{\vec y_d}} \geq  \rho(\vec y)\ell^d-\frac{3L\ell^{d+1}}{2} \geq \delta >  (\frac{\rho(\vec y)}{2})\ell^d\] For the lower bound above, $\ell \geq \big(\frac{2\delta}{\rho(\vec y)}\big)^{\frac{1}{d}}$. For the upper bound above:
%  \[ \rho(\vec y)\ell^d-\frac{3L\ell^{d+1}}{2} \geq \delta \implies \rho(\vec y) \geq \frac{3L\ell}{2} + \frac{\delta}{\ell^d} \]
% From here we can get a rough estimate for $\ell$ by letting
% \[\frac{3L\ell}{2} \leq \frac{\rho(\vec y)}{2} \; \&\; \frac{\delta}{\ell^d} \leq \frac{\rho(\vec y)}{2} \implies  \frac{\rho(\vec y)}{3L}\geq \ell \geq \big(\frac{2\delta}{\rho(\vec y)}\big)^d\] 

% The above two-sided bounds are valid since $\delta\leq (\rho(\vec y)+\frac{3L\ell}{2})\ell^{d}$ where supposedly $\ell\ll 1$. In fact, one can verify that, with $\rho(\vec y)>\frac{3L\ell}{2}$, we require only $\ell<(1/4)^{1/d}$, which is easily satisfied even when $d=2$. Therefore, the lower bound $\big(\frac{2\delta}{\rho(\vec y)}\big)^{d-1}\ll 1$. And we further assume $\rho(\vec y)>3L\ell$, a slightly stronger bound for $\rho(\vec y)$ than the previous, for the upper bound to hold, 

% which is an extreme lower bound for $\rho(\vec y)$ when $\ell$ is considered to be small. We can therefore try to restrict our analysis to $\vec y$ satisfying $\rho(\vec y)>3L\ell$ (). This also implies , again the lower bound is a small quantity for $\ell$ small. 
According to Lemma~\ref{thm:nonuniformsampling}, for $N$ sampling points, with probability at least $p_0$, if a region has probability measure $\geq P_N$, then there is at least one point sampled in that region, where $P_N$ is the threshold probability depending on $N$ obtained by solving the equation below:
\[ N = \frac{\sqrt{\ln{(\frac{2}{1-p_0})}\big(\ln{(\frac{2}{1-p_0})}+2P_N\big)}+P_N + \ln{(\frac{2}{1-p_0})}}{(P_N)^2}.\]
By the bounds for $N$ in \eqref{eq:nonuniformsampling_htm_bounds} of Lemma~\ref{thm:nonuniformsampling}, we know there is some constant $c\in(1,3)$ $s.t.$:
\[ N = \frac{2\ln{(\frac{2}{1-p_0})}+cP_N}{P_N^2} \implies NP_N^2 - cP_N -2\ln{(\frac{2}{1-p_0})} =0. \] 
Solving the above quadratic equation and realize that $P_N>0$, we have 
\[P_N = \frac{c+\big(c^2+8N\ln{(\frac{2}{1-p_0})}\big)^{1/2}}{2N}.\]
Therefore, for a fixed $N$, by requiring \[P_{\Omega_{\vec y_d}}>\frac{\rho(\vec y)}{2}l_N^{d} \geq P_N \implies l_N \geq \Big(\frac{c+\big(c^2+8N\ln{(\frac{2}{1-p_0})}\big)^{1/2}}{\rho(\vec y)N}\Big)^{\frac{1}d},\]
we have with probability $p_0$ that at every surrounding cube of $\Omega_{\vec y}$ of side length $l_N$,  there is at least one point. This lower bound for $l_N$ ensures the surrounding cube has enough probability measure to be sampled. Since $1<c<3$, we can just take $l_N$ to be:
\[l_N := \Big(\frac{3+\big(9+8N\ln{(\frac{2}{1-p_0})}\big)^{1/2}}{\rho(\vec y)N}\Big)^{\frac{1}d}>\Big(\frac{c+\big(c^2+8N\ln{(\frac{2}{1-p_0})}\big)^{1/2}}{\rho(\vec y)N}\Big)^{\frac{1}d}. \]
From above we see that for a fixed $\rho(\vec y)$, $l_N$ decreases as $N$ increases. Therefore, by choosing $N$ large enough, we can always satisfy the prescribed assumption $\rho(\vec y)\geq 3\sqrt{d}Ll_N$. 

Furthermore, when $N$ is so large such that   $\rho(\vec y)\geq 3\sqrt{d}Ll_N$ is always satisfied, we see that $l_N$ is a decreasing function of $\rho$,
meaning that with a higher local density $\rho(\vec y)$, the $l_N$ can be taken smaller while the sampling statement still holds, meaning the local region is more compact.

Finally, since there is a point in every surrounding cube of $\Omega_{\vec y}$, the diameter of the Voronoi cell of $\vec y$ has the following upper-bound with the desired probability $p_0$:
\[ \diam(V(\vec y)) \leq 3l\sqrt{d} = 3\sqrt{d} \Big(\frac{3+\big(9+8N\ln{(\frac{2}{1-p_0})}\big)^{1/2}}{\rho(\vec y)N}\Big)^{\frac{1}d}.\]
Now, for a sample point $x_0$ in the interior of $\supp(\rho)$, given a cover of cubes as in Lemma~\ref{thm:nonuniformsampling}, $x_0$ must belong to one of the cubes with center also denoted by $\vec y$ with a slight abuse of notation. Then note that the diameter of $V(\vec x_0)$ also has the same upper bound as shown above. To go from $\rho(\vec y)$ to $\rho(\vec x_0)$,
% note that if $\rho(\vec y)\geq \rho(\vec x_k)$, then it always holds that:
% \[\diam(V(\vec x_k)) \leq 3\sqrt{d} \Big(\frac{3+\big(9+8N\ln{(\frac{2}{1-p_0})}\big)^{1/2}}{\rho(\vec y)N}\Big)^{\frac{1}d} \leq 3\sqrt{d} \Big(\frac{3+\big(9+8N\ln{(\frac{2}{1-p_0})}\big)^{1/2}}{\rho(\vec x)N}\Big)^{\frac{1}d}\]
% If instead $\rho(\vec y)<\rho(\vec x_k)$, 
by Lipschitz continuity: $\rho(\vec y)\geq \rho(\vec x_0) - \frac{L\sqrt{d}}{2}l_N \implies \rho(\vec x_0) \leq \rho(\vec y)+\frac{L\sqrt{d}}{2}l_N$. Since we require $\rho(\vec y)\geq 3\sqrt{d}Ll_N$, we have $\rho(\vec x_0)\leq \frac{\rho(\vec y)}{6}+\rho(\vec y)=\frac{7}{6}\rho(\vec y)$. Therefore:
\[\rho(\vec y) \geq \frac{6}{7}\rho(\vec x_0) \implies \diam(V(\vec x_0)) \leq 3\sqrt{d} \Big(\frac{21+7\big(9+8N\ln{(\frac{2}{1-p_0})}\big)^{1/2}}{6\rho(\vec x)N}\Big)^{\frac{1}d}.\]
% \[\diam(V(\vec x_k)) \leq 3\sqrt{d} \Big(\frac{3+\big(9+8N\ln{(\frac{2}{1-p_0})}\big)^{1/2}}{\rho(\vec y)N}\Big)^{\frac{1}d} \leq 3\sqrt{d} \Big(\frac{3+\big(9+8N\ln{(\frac{2}{1-p_0})}\big)^{1/2}}{(\rho(\vec x_k)-\frac{L\sqrt{d}}{2}l)N}\Big)^{\frac{1}d} \]
% }
\end{proof}

\begin{proof}[Proof of Lemma~\ref{thm:uniform_diameter}.]
Without loss of generality, we assume that $\abs{\supp(\rho)}=1$, then $\rho=1$ everywhere within its support. We partition $\supp(\rho)$ into $M$ \emph{regions} such that each region has probability measure $\frac{1}{M}$.
This partition can be constructed in the following way: for most of the interior of $\supp(\rho)$, subdivide into hypercubes $\Omega_j$'s of the same size such that $P_{\Omega_j} = \frac{1}{M}$ and $\Omega_j$'s are contained completely inside $\supp(\rho)$. Then the length of the hypercube, $l$, is determined by $l^d/\abs{\supp(\rho)} = 1/M \implies l = (1/M)^{1/d}$. For the remaining uncovered regions of $\supp(\rho)$, cover with some small cubes of appropriate sizes and combine them together to obtain a region with measure $\frac{1}{M}$.

% \Lewis{please confirm if this is legit, or what's the potential issue. }
% \Richard{The above argument seems OK. But you need to be careful about the "regions" being non-overlapping. To be rigorous, if each region is an open set, then the cover will have to be the union of the completion of the regions. }

% Divide the cube to have $\frac{1}{\varepsilon}$ number of sub-cubes along each direction, meaning such sub-cube has side length $\varepsilon>0$. \blue{If one defines a uniform distribution on the minimal cube, which must have the new density $\tilde{\rho}(x)\leq\rho(\vec x)$ point-wise, and sample according to the new uniform distribution on the whole minimal cube,} 

Then, following a similar idea from Lemma~\ref{thm:nonuniform_voronoi}, one has a discrete sampling problem with equal probability for each candidate, which resembles the coupon collector problem. The probability $p(N,d,M)$ that each of the $M$ region contains at least one sample point has a well-known lower bound \cite{doerr2020probabilistic}: \[p(N,d,M) \geq 1-Me^{-\frac{N}{M}}.\]
With the probability $p(N,d,M)$ given above, for an interior hypercube we again have there is at least one sample in each of its surrounding hypercube, since now there is at least one sample in each of the $M$ region. Then the Voronoi diameter for each point is at most $3l\sqrt{d}$. 
Fixing a desired probability $p_0$, we want to determine the number of regions $M$ to get a control on $l$. We need to have a bound as follows: 
\begin{equation}
p\geq  1-Me^{-\frac{N}{M}} \geq p_0 \implies 0< Me^{-\frac{N}{M}} \leq 1-p_0.
\label{eq:lengthbound}
\end{equation}
By rearranging, the above equality holds only when 
\[ \frac{N}Me^{\frac{N}{M}} = \frac{N}{1-p_0}.\] The above equation is solvable by using the Lambert $W$ function: 
\[ M = \frac N{W_0\big(\frac{N}{1-p_0})}.\] where $W_0$ is the principal branch of the Lambert $W$ function. Note that the Lambert $W$ function satisfies
\[ W_0(x) e^{W_0(x)} = x \implies \frac x{W_0(x)} = e^{W_0(x)}.\]
Plug in the above identity, one has 
\[ \frac M{1-p_0} =  \frac N{(1-p_0)W_0\big(\frac{N}{1-p_0})} \implies M = (1-p_0)e^{W_0(\frac{N}{1-p_0}\big)}.\] 
Also note that the function $Me^{-\frac{N}M}$ is monotonically increasing in $M$ (for $M>0$),
so for the bound in \eqref{eq:lengthbound} to hold we require:
\[M \leq (1-p_0)e^{W_0(\frac{N}{1-p_0})}.\]
By taking the largest possible integer $M$ satisfying the above inequality, we then have 
\[ l = (1/M)^{1/d} = \big(\lfloor (1-p_0)e^{W_0(\frac{N}{1-p_0})} \rfloor)^{-1/d},\] for every hypercube contained in $\supp(\rho)$. Then this yields a uniform bound for the Voronoi diameter of any point that is in an interior hypercube surrounded by other interior hypercubes:
\[\diam(V) \leq 3\sqrt{d} \big(\lfloor (1-p_0)e^{W_0(\frac{N}{1-p_0})} \rfloor)^{-\frac{1}{d}}. \]
In terms of the limiting behavior, 
for large $x$, the Lambert $W$ function is asymptotic to the following~\cite{corless1996lambertw, de1981asymptotic}:
\[ W_0(x) = \ln{x} - \ln{\ln{x}} + o(1) \implies e^{W_0(x)} = c(x) \frac{x}{\ln x},\]
with $c(x)\to 1$ as $x\to\infty$. Therefore, for sufficiently large large $N/(1-p_0)$, we have
\begin{equation*}
  \diam(V) \leq
  3\sqrt{d}\left(\left\lfloor (1-p_0)c\frac{N}{(1-p_0)\ln{\frac{N}{1-p_0}}}\right\rfloor \right)^{-\frac{1}{d}}
  = 3\sqrt{d}\left(\left\lfloor\frac{1}{cN}\ln{\frac{N}{1-p_0}}\right\rfloor\right)^{\frac{1}{d}}.
\end{equation*}
\end{proof}
\begin{proof}[Proof of Theorem~\ref{thm:consistency}.]
Note that when using one ray: $S[\Gamma_1](1, j) = \vec x_{1[j]} $ and $S[\Gamma_2](1, j) = \vec x_{2[j]}$, for
$j=1,2,\dots,\ppr$.
The main idea is to bound the difference between each pair of points using
the results introduced in the previous lemmas.
% From Lemma~\ref{thm:nonuniform_voronoi} and~\ref{thm:uniform_diameter},
% with probability $p_0$ we have a bound for the diameter of the Voronoi cell of $x \in\Gamma$,
% denote it by $D(x)$ where $\rho,\,p_0,\,N$ and $d$ are assumed to be fixed.
Consider a fixed sampling point $\vec r_{1,j}\in \vec r(s)$ whose corresponding closest points are $\vec x_{1[j]}$ and $\vec x_{2[j]}$
in $\Gamma_1$ and $\Gamma_2$ respectively.
We consider two cases: first when $\vec r_{1,j}$ is interior to $\supp(\rho)$, in which case from Lemma~\ref{thm:nonuniform_voronoi} and~\ref{thm:uniform_diameter},
with probability $p_0$ we have a bound for the diameter of the Voronoi cell of any interior $\vec x$,
denote it by $D(\vec x)$ where $\rho,\,p_0,\, N$ and $d$ are assumed to be fixed. Therefore, 
\[ \norm{\vec x_{1[j]}-\vec r_{1,j}}_2 \le D(\vec x_{1[j]});\quad  \norm{\vec x_{2[j]}-\vec r_{1,j}}_2 \le D(\vec x_{2[j]}).\]
Then by the triangle inequality:
$\norm{\vec x_{1[j]}-\vec x_{2[j]}}_2 \le D(\vec x_{1[j]})+ D(\vec x_{2[j]})$,
which applies for all sampling points $\vec r_{1,j}$'s in the interior of $\supp(\rho)$, and as $N\to\infty$ we have $D\to 0$ in a rate derived in Lemma~\ref{thm:nonuniform_voronoi}.

In the case where sampling point $\vec r_{1,j}\in \vec r(s)$ is outside of $\supp(\rho)$,
since $\supp(\rho)$ is convex, the closest point to $\vec r_{1,j}$ from $\supp(\rho)$ is always unique, denoted by $\vec x_{\rho}$.
Then, choose $R_1$ depending on $N_1,\,N_2$ such that
the probability measure $ P_1 = P\Big(B_{R_1}(\vec x_{\rho})\cap \supp(\rho) \Big)$
achieves the threshold introduced in Lemma~\ref{thm:nonuniformsampling} so that there is
at least $\vec x_{\rho,1}\in\Gamma_1$ and $\vec x_{\rho,2}\in\Gamma_2$
that lies in $B_{R_1}(\vec x_{\rho})\cap \supp(\rho)$. For sufficiently large $N$, $\vec x_{1[j]}$ and $\vec x_{2[j]}$ would be points inside $B_{R_1}(\vec x_{\rho})\cap \supp(\rho)$ since $\supp(\rho)$ is convex. Then we have
\[ \norm{\vec x_{1[j]} - \vec x_{2[j]}}_2\le 2R_1, \]
and we can pick $N_1, N_2$ large to make $R_1$ as small as desired. Therefore, we have:
\[ \norm{\vec x_{1[j]}-\vec x_{2[j]}}_2 \to 0 \text{ as } N_1, N_2\to\infty, \]
in probability for both interior sampling points and outer sampling points $\vec r_{1,j}$,
and the convergence starts when $N_1, N_2$ get sufficiently large.
Consequently, for the RaySense matrices $S[\Gamma_1]$ and $S[\Gamma_2]$,
we can always find $N$ sufficiently large such that
\[
\norm{S[\Gamma_1]-S[\Gamma_2]}_F = \sqrt{\sum^{\ppr}_{i=1} \norm{\vec x_{1[j]}-\vec x_{2[j]}}_2^2}
\le \varepsilon, %\big(\ppr, N, d, \supp(\rho)\big),
\]
for arbitrarily small $\varepsilon$ depending on $N$, $\ppr$, $d$, and the geometry of $\supp(\rho)$.
\end{proof}

\begin{remark}
  In case of non-convex $\supp(\rho)$ and the sampling point $\vec r_{1,j}\in \vec r(s)$ is outside of $\supp(\rho)$, if the ray $\vec r$ is drawn from some distribution $\mathcal L$, \emph{with probability one}, $\vec r_{1,j}$
  is not equidistant to two or more points on $\supp(\rho)$,
so the closest point is uniquely determined and we only need to worry about the case that
$\vec r_{1,j}$ find the closest point $\vec x_{2[j]}$ from $\Gamma_2$
 that would be far away from $\vec x_{1[j]}$.

Let $\vec x_{\rho}$ be the closest point of $\vec r_{1,j}$ from $\supp(\rho)$, similarly,
choose $R_1$ depending on $N_1,N_2$ such that for balls $B_{R_1}(\vec x_{\rho, 1})$,
the probability measure $ P_1 = P\Big(B_{R_1}(\vec x_{\rho})\cap \supp(\rho) \Big)$
achieves the threshold introduced in Lemma~\ref{thm:nonuniformsampling} so that there is
at least $\vec x_{\rho,1}\in\Gamma_1$ and $\vec x_{\rho,2}\in\Gamma_2$
that lies in $B_{R_1}(\vec x_{\rho})\cap \supp(\rho)$.
Now, consider the case where the closest point $\tilde{\vec x}_{\rho}$ of $\vec r_{1,j}$ from the partial support
$\supp(\rho)\setminus B_{R_1}(\vec x_{\rho})$ is far from $\vec x_{\rho}$ due to the non-convex geometry, and denote 
\[ \delta = \norm{\vec r_{1,j}-\tilde{\vec x}_{\rho}}_2 - \norm{\vec r_{1,j} - \vec x_{\rho}}_2 >0. \]
We pick $N$ so large that
\[
\norm{\vec r_{1,j} - \vec x_{\rho,1}} \le \norm{\vec r_{1,j} - \vec x_{\rho}} + R_1 \le \norm{\vec r_{1,j} - \vec x_{\rho,1}} + \delta \le  \norm{\vec r_{1,j} - \vec x_{\rho,2}},
\]
implying we can find $\vec x_{\rho,1}, \vec x_{\rho,2}$ from $\Gamma_1$ and $\Gamma_2$ closer than $\vec x_{\rho,2}$ from the continuum.
Therefore, for sufficiently large $N_1$ and $N_2$, we can find both closest points $\vec x_{1[j]}$, $\vec x_{2[j]}$ of $\vec r_{1,j}$ inside $B_{R_1}(\vec x_{\rho, 1})\cap \supp(\rho)$ from $\Gamma_1$ and $\Gamma_2$,
$\implies \norm{\vec x_{1[j]}-\vec x_{2[j]}}_2 \le 2R_1 $. The rest follows identically as in the previous proof.
\end{remark}

\section{Details of the proof of Theorem~\ref{thm:line_error_conv}}
\label{sec:proof_details_poisson}

Before deriving the result, we first take a detour to investigate the problem under the setting of Poisson point process, as a means of generating points in a uniform distribution.
% todo: or some more words like "in a way that is controllable" or something like that

\subsection{Poisson point process}\label{sec:poisson}
A Poisson point process \cite{daley2003introduction} on $\Omega$ is a collection of random points such that the
number of points $N_{\Omega'}$ in any bounded measurable subsets $\Omega_j$ with measure $\mu(\Omega_j)$ is a Poisson random variable with rate $\lambda\abs{\Omega_j}$ such that $N_j\sim Poi(\lambda\abs{\Omega_j})$. In other words, we take $N$, 
instead of being fixed, to be a random Poisson variable: $N\sim Poi(\lambda)$,  where the rate parameter $\lambda$ is a constant. Therefore, the underlying Poisson process is homogeneous and it also enjoys the complete independence property, \emph{i.e.}, the number of points in each disjoint and bounded subregion will be completely independent of all the others.

What follows naturally from these properties is that the spatial locations of points generated by the Poisson process is uniformly distributed. As a result, each realization of the homogeneous Poisson process is a uniform sampling of the underlying space with number of points $N\sim Poi(\lambda)$.

Below we state a series of useful statistical properties and concentration inequality for the Poisson random variable:
\begin{itemize}
    \item The Poisson random variable $N\sim Poi(\lambda)$ has mean and variance both $\lambda$: 
    \[\mathbb{E}(N) = Var(N) = \lambda.\]
    \item The corresponding probability density function is
    	\[ \mathbb P({N=k}) = \frac{e^{-\lambda}\lambda^k}{k!}.\]
     \item A useful concentration inequality \cite{pollard1984convergence} ($N$ scales linearly with $\lambda$):
\begin{equation}\label{eqn:poi_lin_scale}
	 \mathbb P(N\le\lambda-\varepsilon) \le  e^{-\frac{\varepsilon^2}{2(\lambda+\varepsilon)}}\; \text{ or } \;\mathbb P(\abs{N-\lambda}\ge \varepsilon) \le  2e^{-\frac{\varepsilon^2}{2(\lambda+\varepsilon)}}.
\end{equation}
\end{itemize}
Furthermore, one can also derive a Markov-type inequality for the event a Poisson random variable $N\sim Poi(\lambda)$ is larger than some $a>\lambda$ that is independent of $\lambda$, different from~\eqref{eqn:poi_lin_scale}:
\begin{proposition}
For Poisson random variable $N\sim Poi(\lambda)$, it satisfies the following bound for any constant $a>\lambda$:
\begin{equation}\label{eqn:P(N>a)}
	\bbP(N \ge a) \le \frac{e^{(a-\lambda)}\lambda^a}{a^a} \iff \bbP(N < a) \ge 1 - \frac{e^{(a-\lambda)}\lambda^a}{a^a}.
\end{equation}
\begin{proof}
    By Markov's inequality:
\begin{align*}
	\bbP(N \ge a) &= \bbP(e^{tN}\ge e^{ta})\le \inf_{t>0}\frac{\bbE(e^{tN})}{e^{ta}} = \inf_{t>0}\frac{\sum_{k=1}^{\infty} e^{tk}\frac{\lambda^ke^{-\lambda}}{k!}}{e^{ta}} \\
	&  = \inf_{t>0}\frac{e^{-\lambda}\sum_{k=1}^{\infty} \frac{(e^{t}\lambda)^k}{k!}}{e^{ta}} = \inf_{t>0}\frac{e^{-\lambda} e^{e^t\lambda}}{e^{ta}}
	= \inf_{t>0}\frac{e^{(e^t-1)\lambda}}{e^{ta}}.
\end{align*}
To get a tighter bound, we want to minimize the R.H.S.. Let $\zeta=e^t>1$, then we minimize the R.H.S. over $\zeta$:
\[ \min_{\zeta>1} \frac{e^{(\zeta-1)\lambda}}{\zeta^a} \iff \min_{\zeta>1} (\zeta-1)\lambda - a \log(\zeta).\]
A simple derivative test yields the global minimizer: $\zeta = \frac{a}{\lambda}>1$ since we require $a>\lambda$. Thus:
\begin{equation}
	\bbP(N \ge a) \le \frac{e^{(a-\lambda)}\lambda^a}{a^a} \iff \bbP(N < a) \ge 1 - \frac{e^{(a-\lambda)}\lambda^a}{a^a}.
\end{equation}
    \end{proof}
\end{proposition}
A direct consequence of \eqref{eqn:poi_lin_scale} is that one can identify $N$ with $\lambda$ with high probability when $\lambda$ is large, or equivalently the other way around:

\begin{lemma} \label{thm:lambdaN} (Identify $N$ with $\lambda$)\\
  A point set of cardinality $N^*$ drawn from a uniform distribution, with high probability,
  can be regarded
  as a realization of a Poisson point process with rate $\lambda$ such that
    \[ \mathbb P\big(\frac{2N^*}{3}\le \lambda \le 2N^*\big) \ge 1 - e^{-\frac{N^*}{6}} -e^{-\frac{N^*}{18}}. \]
\begin{proof}
If $N\sim Poi(\lambda)$, by taking $\varepsilon = \lambda/2$, from \eqref{eqn:poi_lin_scale} we have:
\[ \mathbb P\big(\abs{N-\lambda} <\frac{\lambda}{2}\big) \ge  1- 2e^{-\frac{\lambda}{12}} \iff \mathbb P\big( \frac{\lambda} 2 < N < \frac{3\lambda}2 \big) \ge  1- 2e^{-\frac{\lambda}{12}}.  \]
Let $\lambda_u = 2N^*$ as a potential upper bound for $\lambda$, while $\lambda_l = \frac{2N^*}3$ the potential lower bound. Then for $N_u\sim Poi(\lambda_u)$ and $N_l\sim Poi(\lambda_l)$:
\[ \mathbb P(N_u\le N^*) = \mathbb P(N_u\le \frac{\lambda} 2) \le e^{-\frac{\lambda_u}{12}}, \]
\[\mathbb P(N_l\ge N^*) = \mathbb P(N_l\ge \frac{3\lambda_l}2) \le e^{-\frac{\lambda_l}{12}}.\]
Therefore, if we have some other Poisson processes with rate $\lambda_1 >\lambda_u$, and $\lambda_2<\lambda_l$ the probabilities of the corresponding Poisson variables $N_1\sim Poi(\lambda_1), N_2\sim Poi(\lambda_2)$ to achieve at most (or at least) $N^*$ is bounded by:
\[ \mathbb P(N_1\le N^*) < \mathbb P(N_u\le N^*) \le e^{-\frac{\lambda_u}{12}} = e^{-\frac{N^*}{6}},\]
\[ \mathbb P(N_2 \ge N^*) < \mathbb P(N_l \ge N^*) \le e^{-\frac{\lambda_l}{12}} = e^{-\frac{N^*}{18}}.\]
Note that both of the events have probability decaying to $0$ as the observation $N^* \to \infty$, therefore we have a confidence interval of left margin $ e^{-\frac{N^*}{6}}$ and right margin $ e^{-\frac{N^*}{18}}$ to conclude that the Poisson parameter $\lambda$ behind the observation $N^*$ has the bound:
\[\frac{2N^*}{3}\le \lambda \le 2N^*.\]
Since the margins shrink to $0$ as $N^* \to \infty$,
we can identify $\lambda$ as $cN$ with some constant $c$ around $1$ with high probability.
\end{proof}
\end{lemma}
By Proposition~\ref{thm:lambdaN}, for the remaining, we will approach the proof to Theorem~\ref{thm:line_error_conv} from a Poisson process perspective and derive results with the Poisson parameter $\lambda$. 

\subsection{Main ideas of the proof of Theorem~\ref{thm:line_error_conv}}\label{sec:poi_scale}

Consider a Poisson process with parameter $\lambda$ in $\supp(\rho)$ and a corresponding point cloud $\Gamma$ with cardinality $N\sim Poi(\lambda)$. Based on previous discussion from \S~\ref{sec:integral}, we assume the ray $\vec r(s)$ is given entirely in the interior of $\supp(\rho)$. From Theorem~\ref{thm:approx_voronoi}, by denoting $1\le k_i\le N$ such that $\{\vec x_{k_i}\}_{i=1}^M\subset\Gamma$ are points in $\Gamma$ sensed by the ray and $V_{k_i} := V(\vec x_{k_i})$,
equivalently the line integral error is:
% \footnote{For clarity,  we can always apply a rotation to both of the point cloud $\Gamma$ and $\vec r(s)$ so that $\vec r(s)$ is aligned with the $x_1$-axis in $\mathbb R^d$ where the underlying problem and the homogeneous Poisson  process remains the same with rate $\lambda$.}
% \Colin{Did this footnote do anything?  We never used that idea that I can see: remove it?}
\begin{equation}
  \label{eqn:line_error_M_form}
    \absbig{\int_0^1 g\big(\vec r(s)\big) \mathrm{d}s - \int_0^1  g\big(\vec x_{k(s)}\big)\mathrm{d}s} \le J\sum_{i=1}^M \int_0^1 \chi\big(\{\vec r(s)\in V_{k_i}\}\big)\norm{\vec r(s) -  \vec x_{k_i}} \mathrm{d}s,
    \end{equation}
To bound the above quantity, one needs to bound $M$ the number of Voronoi cells a line goes through, the length of $\vec r(s)$ staying inside $V_{k_i}$ and the distance to the corresponding $\vec x_{k_i}$ for each $\vec r(s)$ altogether. Our \emph{key intuition} is stated as follows:

Divide the ray $\vec r(s)$ of length $1$ into segments each of length $h$, and consider a hypercylinder of height $h$ and radius $h$ centering around each segment.
If there is at least one point from $\Gamma$ in each of the hypercylinders,
then no point along $\vec r(s)$ will have its nearest neighbor further than distance $H = \sqrt{2}h$ away from $\vec r(s)$.
Therefore, we restrict our focus to $\Omega$, a tubular neighborhood of distance $H$ around $\vec r(s)$---that is, a ``baguette-like'' region with spherical end caps.
$N_{\Omega}$, the number of points of $\Gamma$ that are inside $\Omega$, will serve as an upper bound for $M$
(the total number of unique nearest neighbors of $\vec r(s)$ in $\Gamma$)
while the control of the other two quantities (intersecting length and distances to closest points) comes up naturally.

Undoubtedly $M$ depends on the size of $\Omega$, which is controlled by $h$.
The magnitude of $h$ therefore becomes the crucial factor we need to determine.
The following lemma motivates the choice of $h=\lambda^{(\varepsilon-1/d)} $ for some small $1\gg\varepsilon>0$.

\begin{lemma} \label{thm:poi_scale}
Under A$1$-A$7$, for a point cloud of cardinality $N\sim Poi(\lambda)$ generated from a Poisson point process, and a ray $\vec r(s)$ given entirely in $\supp(\rho)$, the number of points $N_\Omega$ in the tubular neighborhood of radius $H=\sqrt{2}h$ around $\vec r(s)$ will be bounded 
% \big (by $c(d)\lambda^{\frac{1}{d}+(d-1)\varepsilon}$\big ) 
when
\[h =\lambda^{\!-\frac{1}{d}+\varepsilon}\!\!, \]
for some small $1\gg\varepsilon>0$, with probability $\to 1$ as $\lambda \to \infty$.
\begin{proof}
Note that the baguette region $\Omega$ has outer radius $H$, and hypercylinders of radius $h$ are contained inside $\Omega$. For simplicity we prescribe $h$ such that $Q = 1/h$ is an integer, then the baguette region $\Omega$ consists of $Q$ number of hypercylinders, denoted by $\{\Omega_j\}_{j=1}^Q$ and the remaining region, denoted by $\Omega_r$ consisting of an annulus of outer radius $H$, inner radius $h$, and two half spheres of radius $H$ on each side. Since each region is disjoint, according to \S~\ref{sec:poisson} the Poisson process with rate $\lambda$ in $\supp(\rho)$ will have Poisson sub-process in each of the regions in a rate related to their Lesbegue measure, and all the sub-processes are independent.

Now, let $\bbP_Q$ denote the probability of having at least one point in each $\Omega_j$ in $\{\Omega_j\}_{j=1}^Q$ while the number of points in each $\Omega_j$ is also uniformly bounded by some constant $N_Q$.
Since each $\Omega_j$ has the same measure, their corresponding Poisson processes have the identical rate $\lambda_q = 
\abs{\Omega_1}\lambda$. Let $N_j$ denote the Poisson random variable for $\Omega_j$, then:
\[\bbP(N_j\ge 1) = 1 - \bbP(N_j = 0) = 1 - e^{-\lambda_q}.\]
Combined with~\eqref{eqn:P(N>a)} by requiring $N_Q>\lambda_q$, this implies
\[ \bbP(N_Q > N_j\ge 1) = \bbP(N_j\ge 1) - \bbP(N_j \ge N_Q) \ge 1 - e^{-\lambda_q} - \frac{e^{(N_Q-\lambda_q)}\lambda_q^{N_Q}}{N_Q^{N_Q}}  \,,\]
and hence
\begin{equation}\label{eqn:P_Q}
	 \bbP_Q \ge \bigg(1 - e^{-\lambda_q}-\frac{e^{(N_Q-\lambda_q)}\lambda_q^{N_Q}}{N_Q^{N_Q}} \bigg)^Q  \ge 1 - Q\bigg(e^{-\lambda_q} + \frac{e^{(N_Q-\lambda_q)}\lambda_q^{N_Q}}{N_Q^{N_Q}} \bigg).
\end{equation}
The measure of the remaining region $\Omega_r$ is $\abs{\Omega_r} = \omega_{d}H^{d} + \omega_{d-1}(H^{d-1}-h^{d-1})$, where $\omega_d$ is the volume of the unit $d$-sphere.
Therefore the Poisson process on $\Omega_r$ has rate $\lambda_r = \abs{\Omega_r}\lambda$.
Let $N_r$ denote the corresponding Poisson random variable, again by~\eqref{eqn:P(N>a)} with $N'>\lambda_r$:
\begin{equation}\label{eqn:P_r}
	\bbP(N_r < N') \ge 1 -\frac{e^{(N'-\lambda_r)}\lambda_r^{N'}}{N'^{N'}}.
\end{equation}
Since $\Omega_r$ and $\bigcup\{\Omega_j\}_{j=1}^Q$ are disjoint, by independence, the combined probability $p_{tot}$ that all these events happen:
\begin{enumerate}
    \item the number of points $N_j$ in each hypercylinder $\Omega_j$ is at least $1$ 
    \item $N_j$ is uniformly bounded above by some constant $N_Q$
    \item the number of points $N_r$ in the remaining regions $\Omega_r = \Omega-\cup_j \{\Omega_j\}$ is also bounded above by some constant $N'$
\end{enumerate}
would have the lower bound:
\begin{align*}
    p_{tot} &\ge \bigg( 1 -\frac{e^{(N'-\lambda_r)}\lambda_r^{N'}}{N'^{N'}}\bigg) \bigg( 1 - Q\bigg(e^{-\lambda_q} + \frac{e^{(N_Q-\lambda_q)}\lambda_q^{N_Q}}{N_Q^{N_Q}} \bigg) \bigg) \\ &\ge 1 - \frac{e^{(N'-\lambda_r)}\lambda_r^{N'}}{N'^{N'}} - Q\bigg(e^{-\lambda_q} + \frac{e^{(N_Q-\lambda_q)}\lambda_q^{N_Q}}{N_Q^{N_Q}} \bigg).
\end{align*}
Then with probability $p_{tot}$, we have an upper bound for $N_{\Omega}$, the total number of points in $\Omega$:
\begin{equation} \label{eqn:N_omega}
	N_{\Omega} \le N' + QN_Q.
\end{equation}
Apparently $N_{\Omega}$ and $p_{tot}$ are inter-dependent:
as we restrict the R.H.S. bound in~\eqref{eqn:N_omega} by choosing a smaller $N'$ or $N_Q$, the bound for $p_{tot}$ will be loosened.
From Lemma~\ref{thm:lambdaN}, we set $N' = \alpha\lambda_r$, $N_Q = \beta\lambda_q$ for some $\alpha,\beta>1$.
Therefore, the next step is to determine the parameter set $\{h,\,\alpha,\,\beta\}$ to give a more balanced bound
to the R.H.S. in~\eqref{eqn:N_omega} while still ensuring the probability of undesired events will have exponential decay.

For that purpose we need some \emph{optimization}. We know:
\[\lambda_r = \abs{\Omega_r}\lambda =(\omega_{d}H^{d} +\omega_{d-1}(H^{d-1}-h^{d-1}))\lambda = (\omega_{d}2^{\frac{d}{2}}h^{d} + \omega_{d-1}(2^{\frac{d-1}{2}}-1)h^{d-1})\lambda; \]
\begin{equation} \label{eqn:sublambda_size}
    \lambda_q = \abs{\Omega_1}\lambda = (\omega_{d-1}h^{d-1}h)\lambda = \omega_{d-1}
h^d\lambda.
\end{equation}
We need to investigate how $h$ should scale with $\lambda$, so we assume $h\sim\lambda^{-p}$ for some constant $p$ to be determined.
The following optimization procedure provides some motivations for choosing $p$.
On the one hand, for the constraints we need to ensure that the probability of each of three events above not occurring decays
to $0$ as $\lambda\to\infty$
\begin{align*}
  \frac{e^{(N'-\lambda_r)}\lambda_r^{N'}}{N'^{N'}} \to 0
  &\iff
  (N'-\lambda_r) + N'\log(\lambda_r) - N'\log(N') \to -\infty, \\
  Qe^{-\lambda_q}\to 0
  &\iff
  \log(Q) - \lambda_q \to -\infty, \\
  Q\frac{e^{(N_Q-\lambda_q)}\lambda_q^{N_Q}}{N_Q^{N_Q}}  \to 0
  &\iff
  \log(Q) + (N_Q-\lambda_q) + N_Q\log(\lambda_q) - N_Q \log(N_Q)  \to -\infty.
\end{align*}
%% \begin{equation*}
%% \begin{dcases}
%% 	\frac{e^{(N'-\lambda_r)}\lambda_r^{N'}}{N'^{N'}} \to 0\\
%% 	Qe^{-\lambda_q}\to 0\\
%% 	Q\frac{e^{(N_Q-\lambda_q)}\lambda_q^{N_Q}}{N_Q^{N_Q}}  \to 0\\
%% \end{dcases}
%% \hspace{-2em}\iff
%% \begin{dcases}
%% 	(N'-\lambda_r) + N'\log(\lambda_r) - N'\log(N') \to -\infty\\
%% 	\log(Q) - \lambda_q \to -\infty\ \\
%% 	\log(Q) + (N_Q-\lambda_q) + N_Q\log(\lambda_q) - N_Q \log(N_Q)  \to -\infty\\
%% \end{dcases}
%% \end{equation*}
and representing all the quantities in terms of $\lambda,\, p,\, \alpha,\,\beta$ and simplifying:
\begin{align*}
  (\alpha-1)\lambda_r - \alpha\lambda_r\log(\alpha) \to -\infty
  \quad & \implies
  \alpha\big(1-\log(\alpha)\big)<1
  & \implies
  \alpha > 1 \\
  p\log(\lambda) -  \omega_{d-1}\lambda^{-pd+1} \to -\infty
  \quad & \implies
  \beta\big(1-\log(\beta)\big)<1
  & \implies
   \beta > 1 \\
  \log(Q) + (\beta-1)\lambda_q - \beta\lambda_q\log(\beta) \to -\infty
  \quad & \implies
  -pd + 1 > 0
  & \implies
  p < \frac{1}{d}
\end{align*}
%% \[
%% \begin{dcases}
%% 	% (\alpha-1)\lambda_r + N'\log(\lambda_r) - \alpha\lambda_r\log(\alpha) -N'\log(\lambda_r) \to -\infty\\
%% 		(\alpha-1)\lambda_r - \alpha\lambda_r\log(\alpha) \to -\infty\\
%% 	p\log(\lambda) -  \omega_{d-1}\lambda^{-pd+1} \to -\infty\ \\
%% 	\log(Q) + 	(\beta-1)\lambda_q - \beta\lambda_q\log(\beta) \to -\infty\\
%% \end{dcases}
%% \implies
%% \begin{dcases}
%% 	\alpha\big(1-\log(\alpha)\big)<1\\
%% 		\beta\big(1-\log(\beta)\big)<1\\
%%  -pd + 1 > 0 \\
%% \end{dcases}
%% \implies \begin{dcases}
%% 	\alpha > 1 \\ \beta > 1 \\ p < \frac{1}{d}
%% \end{dcases}
%% \]
On the other hand, for the objective, note that
\begin{align*}
	N' + \frac{N_Q}{h} &\le 2\max (N',\, \frac{N_Q}{h}) \\ &= 2\max\bigg(\alpha\big((\omega_{d}2^{\frac{d}{2}}h^{d} + \omega_{d-1}(2^{\frac{d-1}{2}}-1)h^{d-1})\lambda\big), \;\frac{\beta}{h}\omega_{d-1}
	h^d\lambda\bigg)\\
	&\le 3\max\bigg(\alpha\omega_{d}2^{\frac{d}{2}}h^{d}\lambda,\; \alpha\omega_{d-1}(2^{\frac{d-1}{2}}-1)h^{d-1}\lambda,\; \frac{\beta}{h}\omega_{d-1}
	h^d\lambda   \bigg),
\end{align*}
since $\alpha,\,\beta$ are just some constants $>1$, fixing $\alpha$ and $\beta$ so that $h = \lambda^{-p}$ and we want to minimize $h$ to obtain an upper bound for the total number of points in $\Omega$:
\begin{align*}\label{eqn:obj}
	&\argmin_{h} \max\bigg(\alpha\omega_{d}2^{\frac{d}{2}}h^{d}\lambda,\, \alpha\omega_{d-1}(2^{\frac{d-1}{2}}-1)h^{d-1}\lambda,\, \frac{\beta}{h}\omega_{d-1}
	h^d\lambda \bigg)\\
	% \iff &\argmin_{h} \max \bigg(\log(\alpha\omega_{d}2^{\frac{d}{2}}) +(1-pd)\log(\lambda), \, \log(\alpha\omega_{d-1}(2^{\frac{d-1}{2}}-1))+(1-pd+p)\log(\lambda) , \, \log(\omega_{w-1})+(1-pd+p)\log(\lambda)\bigg)
		\iff &\argmin_{p} \max \bigg(c_2+(1-pd+p)\log(\lambda) , \, c_3+(1-pd+p)\log(\lambda)\bigg)\\
		\iff &\argmin_{p} \bigg((1-pd+p)\log(\lambda)\bigg).
\end{align*}
Combined with bounds derived from the constraints, to minimize $(1-(d-1)p)$, we need to maximize $p$,  therefore we take $p = \frac{1}{d}-\varepsilon$ for an infinitesimal $\varepsilon>0$.

%  and $\alpha = \beta = e>1$ for simplifying calculations to arrive at:
% \begin{align*}
% 	 N_{\Omega} \le& e \bigg(\omega_{d}2^{\frac{d}{2}}\lambda^{-1 +\varepsilon d} + \omega_{d-1}(2^{\frac{d-1}{2}}-1)\lambda^{-(d-1)/d+\varepsilon(d-1)}\bigg)\lambda  + e\lambda^{\frac{1}{d}-\varepsilon} \omega_{d-1}\lambda^{-1+\varepsilon d}\lambda\\
% 	 \le& e\bigg(\omega_{d}2^{\frac{d}{2}}\lambda^{\varepsilon d} + \omega_{d-1}(2^{\frac{d-1}{2}}-1)\lambda^{\frac{1}{d}+\varepsilon(d-1)}\bigg)  + e\omega_{d-1}\lambda^{\frac{1}d + (d-1)\varepsilon}\\
% 	 \le & e\omega_{d}2^{\frac{d}{2}}\lambda^{\varepsilon d} + e\omega_{d-1}2^{\frac{d-1}{2}}\lambda^{\frac{1}{d}+\varepsilon(d-1)} \le e\omega 2^d \lambda^{\frac{1}{d}+\varepsilon d}
% \end{align*}
% where $\omega = \max (\omega_{d-1}, \omega_d)$. 
% And with $\alpha=\beta = e\implies N' = e\lambda_r, \, N_Q = e\lambda_q$, the calculations for probability bound in~\eqref{eqn:P_Q}~\eqref{eqn:P_r} are much simplified because:
% \[ \frac{e^{(N'-\lambda_r)}\lambda_r^{N'}}{N'^{N'}} =\frac{e^{N'}e^{-\lambda_r}\lambda_r^{N'}}{e^{N'}\lambda_r^{N'}} = e^{-\lambda_r}; \quad \frac{e^{(N_Q-\lambda_q)}\lambda_q^{N_Q}}{N_Q^{N_Q}} = \frac{e^{N_Q}e^{-\lambda_q}\lambda_q^{N_Q}}{e^{N_Q}\lambda_q^{N_Q}} = e^{-\lambda_q}\]
\end{proof}
\end{lemma}

\subsection{Proof of Theorem~\ref{thm:line_error_conv}}
\begin{proof}[Proof of Theorem~\ref{thm:line_error_conv}.]

Consider a Poisson process with rate $\lambda$ on the $\supp(\rho)$.
As in \S~\ref{sec:poi_scale}, let $Q = 1/h$ be an integer for simplicity (or take ceiling if desired), and consider $Q$ hypercylinders of radius $h$ centered along $\vec r(s)$.
Again as in \S~\ref{sec:poi_scale}, let $\Omega$ be the tubular neighborhood of distance $H = \sqrt{2}h$ around $\vec r(s)$.
Motivated by \S~\ref{sec:poi_scale}, we set
% \[ h =\left(\frac{1}{a\lambda}\ln b\lambda\right)^{\!\frac{1}{d}}, \]
\[ h =\lambda^{\!-\frac{1}{d}+\varepsilon}, \]
for some small constants $1\gg\varepsilon>0$ to be determined.

Divide the tubular neighborhood $\Omega$ into two parts,
one consists of the set of hypercylinders $\bigcup_{j=1}^Q \Omega_j$  around $\vec r(s)$,
the other is the remainder $\Omega_r$.
From the setting of Lemma~\ref{thm:poi_scale}, let $N' = \alpha\lambda_r$ be the number of points in $\Omega_r$ while $N_Q = \beta\lambda_q$ is for $\Omega_j$, and we set $\alpha=\beta = e > 1$ (also satisfying the constraints in Lemma~\ref{thm:poi_scale}) to simplify the calculations so that we have $N_Q = e\lambda_q, N_r = e\lambda_r$, and Equation~\eqref{eqn:P_Q} becomes:
\[ \bbP(e\lambda_q> N_j\ge 1) = \bbP(N_j\ge 1) - \bbP(N_j \ge e\lambda_q) \ge 1 - 2e^{-\lambda_q} \,,\]
% \label{eqn:P_Q}
\begin{equation}
	 \implies \bbP_Q \ge \bigg(1 - 2e^{-\lambda_q}\bigg)^Q  \ge 1 - 2Qe^{-\lambda_q}.
\end{equation}
So the total number in $\bigcup_{j=1}^Q \Omega_j$ is bounded by $Qe\lambda_q$ while there is still at least one point in every $\Omega_j$ with the above probability. On the other hand for~\eqref{eqn:P_r}:
\[ \bbP(N_r < e\lambda_r) \ge 1 - e^{-\lambda_r}.\]
Again by the same independence argument, the total probability that all the events happen has the following lower bound:
\[p_{tot} \ge (1 - e^{-\lambda_r}) ( 1 - 2Qe^{-\lambda_q}) \ge 1 - 2Qe^{-\lambda_q} - e^{-\lambda_r}.\]
And the total number of points inside $\Omega$ is bounded by:
\[ N_{\Omega}\le e\lambda_r + Qe\lambda_q = e(\lambda_r + Q\lambda_q) = e\lambda_{\Omega}.\]
Finally, when there is at least one point in each of $\Omega_j$, the maximum distance from any point on $\vec r(s)$ to its nearest neighbor is given by $H=\sqrt{2}h$ as we have argued. Furthermore, under this setting, for any of the potential nearest neighbors, the maximum length that $\vec r(s)$ intersect its Voronoi cell has an upper bound of $3h$.
Therefore, the line integral error~\eqref{eqn:line_error_M_form} is bounded by:
\[
\absbig{\int_0^1 g\big(\vec r(s)\big) \mathrm{d}s - \int_0^1  g\big(\vec x_{k(s)}\big)\mathrm{d}s}
\le \frac{J}{2} \sum_{i=1}^M H\times 3h
\le \frac{J}{2} N_{\Omega}\times H\times 3h
= 3\sqrt{2}h^2e\lambda_{\Omega}
\]
\[
\le \frac{3e\sqrt{2}J}2 h^2 (\omega_d2^{\frac{d}2}h^d + \omega_{d-1} 2^{\frac{d-1}{2}}h^{d-1})\lambda
\le c(d,J) (h^{d+2}+h^{d+1})\lambda
\le c(d,J) \lambda^{-\frac{1}{d}+\varepsilon(d+1)}.
\]
Finally, for the total probability $p_{tot}$:
\[p_{tot} \ge 1 - 2Qe^{-\lambda_q} - e^{-\lambda_r}=1- \frac{2}{h}e^{-\omega_{d-1}h^d\lambda} - e^{-\abs{\Omega_r}\lambda},\]
and recall from~\eqref{eqn:sublambda_size}: $\lambda_q = \abs{\Omega_1}\lambda = \omega_{d-1}h^{d-1}h\lambda = \omega_{d-1}h^d\lambda = \omega_{d-1}\lambda^{\varepsilon d}$. Then:
\[2Qe^{-\lambda_q} = 2(\lambda)^{\frac{1}{d}-\varepsilon}e^{-\omega_{d-1}\lambda^{\varepsilon d}} \to 0 \;\text{ as } \lambda \to\infty.\]
The above convergence can be shown by taking the natural log:
\[ \ln\Big(2(\lambda)^{\frac{1}{d}-\varepsilon}e^{-\omega_{d-1}\lambda^{\varepsilon}}\Big) = \ln2 + (\frac{1}{d}-\varepsilon)\ln\lambda -\omega_{d-1}\lambda^{\varepsilon}\to -\infty\;\text{ as } \lambda \to\infty,\]since $\ln{\lambda} $ grows slower than $\lambda^{\varepsilon}$ for any $\varepsilon>0$. As for the last term $e^{-\abs{\Omega_r}\lambda}$:
\[ e^{-\abs{\Omega_r}\lambda} \le e^{-\omega_{d-1}(2^{\frac{d-1}{2}}-1)h^{d-1}\lambda}\le e^{-c\lambda^{\frac{1}{d}+\varepsilon(d-1)}} \le e^{-c(d)\lambda^{\frac{1}{d}} } \to 0 \;\text{ as } \lambda \to\infty.\]
Thus, the probability $p_{tot}\to 1$ as $\lambda \to\infty$, and we have our line integral error $\le c(d,J) \lambda^{-\frac{1}{d}+\varepsilon(d+1)} \to 0$ as long as $\varepsilon <\frac{1}{(d+1)^2}$.
To obtain the convergence in terms of the actual number of points $N$ in the point cloud,
we invoke Lemma~\ref{thm:lambdaN} and set $N=c\lambda$ to conclude the proof.
\end{proof}

\end{document}